\documentclass[11pt]{article}

\usepackage[final]{acl}

\usepackage{times}
\usepackage{latexsym}

\usepackage[T1]{fontenc}

\usepackage[utf8]{inputenc}

\usepackage{microtype}

\usepackage{inconsolata}

\usepackage{graphicx}
\usepackage{subcaption}
\usepackage{amsmath}
\usepackage{amsfonts}
\usepackage{url}
\usepackage{array}
\usepackage{makecell}
\usepackage{multirow}
\usepackage{booktabs}
\usepackage{todonotes}
\usepackage{placeins}
\usepackage{afterpage}
\usepackage{subcaption}
\usepackage{cuted}
\usepackage{caption}
\usepackage{needspace}
\usepackage{tabularx}
\usepackage[percent]{overpic}
\usepackage{xcolor}
\usepackage{xcolor}
\usepackage{tikz}
\usepackage{lineno}
\usepackage{capt-of}
\title{Beyond a Bag of Features: Set-Level Instability in Sparse Autoencoders}

\author{
	\textbf{Nikolai Bolik} \and
	\textbf{Lennart Stöpler} \and
	\textbf{Artur Andrzejak} \\
	Heidelberg University \\
	\texttt{\{nikolai.bolik,lennart.stoepler,artur.andrzejak\}@uni-heidelberg.de}
}

\begin{document}
\raggedbottom
\maketitle
\begin{abstract}
\citet{shani2025tokensthoughtsllmshumans} show that LLM representations broadly recover human category boundaries, while failing to reflect fine-grained typicality structure. Their analysis uses cosine similarity over dense model representations. We revisit their approach using overlap over active sparse autoencoder (SAE) latent sets as a more interpretable similarity measure. We first verify that this set-level measure is meaningful: SAE latent sets can recover union-like compositional structure in controlled toy models and induce semantically coherent neighborhoods in natural text. Extending the human-concepts analysis to SAE set similarities, we find that SAE activation sets do not recover human category boundaries or within-category typicality more faithfully than dense embeddings or residual-stream states, but instead track model-internal similarity structure. To probe this gap further, we study active latent sets under well-controlled semantic modifications, revealing a substantial mismatch between human judgements of conceptual change and change in the SAE active set. We interpret this as evidence that, outside idealised settings, SAE features do not compose via simple bag-of-features semantics.
\end{abstract}

\section{Introduction}
In a recent study \citet{shani2025tokensthoughtsllmshumans} probe how human concepts are encoded in Transformer models. Using cosine similarity of token embeddings as a measure of semantic closeness, they find that LLM-derived similarity clusters broadly align with human conceptual structure, while failing to reflect the fine-grained internal organisation of those clusters. A natural follow-up question is whether a more interpretable similarity measure changes this picture. Active feature sets from sparse autoencoders (SAEs) (\citet{cunningham2023sparseautoencodershighlyinterpretable}; \citet{scaling_templeton_2024}; \textit{inter alia}) offer such a candidate: They decompose Transformer activations into sparse features \citep{elhage2022toy, 10.5555/3692070.3693675, bricken2023towards}, with individual latents shown to admit meaningful semantic interpretations, scaling up to state-of-the-art model sizes \citep{bricken2023towards, cunningham2023sparseautoencodershighlyinterpretable, scaling_templeton_2024, lieberum-etal-2024-gemma}. We therefore revisit the experiments of \citet{shani2025tokensthoughtsllmshumans}, replacing cosine similarity of embeddings with overlap over active SAE latent sets\footnote{In this work, we use \emph{feature} to refer to a ground-truth concept the model operates on, and \emph{latent} to refer to a learned SAE direction that may encode such a feature.}.\\
While active SAE latent sets are increasingly used as objects of analysis \citep{park-etal-2025-decoding, Olsen2025Probing, wattenberg2024relational, olah2023distributedcomposition}, most SAE work focuses on the interpretability of individual latents \citep{gao2024scalingevaluatingsparseautoencoders, batchtopk_bussmann_2024, improving_rajamanoharan_2024, jumping_rajamanoharan_2024}. We therefore first test whether set-level SAE similarity is meaningful in our setting. In preliminary studies, we show that SAE latent-set overlap forms semantically coherent clusters on large text corpora, and that SAEs can compose in a bag-of-features manner in a controlled synthetic toy model. We then replicate the conceptual-similarity experiments of \citet{shani2025tokensthoughtsllmshumans} using SAE latent-set overlap. The qualitative picture is preserved: SAE-feature similarity broadly tracks human conceptual groupings. However, SAE-based similarity performs slightly \emph{worse} than raw cosine similarity, despite the interpretability of individual latents. To understand this gap, we study active latent sets under controlled semantic modifications of the input and find a substantial mismatch between human judgements of conceptual change and changes in SAE active latent sets. We interpret this as evidence that, outside idealised settings, SAE features do not compose via simple bag-of-features semantics.\\
We summarise our main contributions as follows:
\begin{enumerate}
    \item We provide preliminary evidence that active SAE latent sets are a meaningful unit of analysis, using clustering experiments on naturalistic text sequences and a toy-model study demonstrating bag-of-features compositionality on synthetic inputs.
    \item We replicate the conceptual-similarity experiments of \citet{shani2025tokensthoughtsllmshumans} with SAE latent-set overlap as the similarity measure, finding qualitatively similar but quantitatively weaker alignment with human conceptual structure than cosine similarity.
    \item We study active SAE latent sets under controlled semantic modifications, revealing a mismatch between human conceptual change and latent-set change, suggesting that bag-of-features compositionality does not hold in naturalistic settings.
\end{enumerate}

\section{Terms and Definitions}
\label{sec:terms_and_defs}
\paragraph{Linear Representation Hypothesis}
A central conjecture in mechanistic interpretability is the Linear Representation Hypothesis (LRH) \citep{elhage2022toy, 10.5555/3692070.3693675}. It states that the residual stream of a Transformer encodes many human-interpretable concepts as directions in activation space, and that model computations can be understood as non-linear operations over these linearly embedded concepts. In the form commonly assumed in the SAE literature, a representation vector \(x\) is decomposed as
\begin{align}
x &= b + \sum_i f_i d_i ,
\label{eq:decomposition}
\end{align}
with non-negative coefficients \(f_i\), directions \(d_i\in\mathbb{R}^d\), and bias \(b\in\mathbb{R}^d\) \citep{bricken2023towards}.\footnote{More complex formulations of the LRH exist, which we discuss in Section~\ref{sec:LRH_critique}.}
Since the number of encoded concepts is assumed to vastly exceed the dimensionality of the representation space, the directions \(d_i\) cannot all be orthogonal, and the model must be robust to the resulting interference.

\paragraph{Sparse Autoencoders}
If the LRH holds, recovering the directions \(d_i\) and coefficients \(f_i\) from observed activations is an instance of dictionary learning, i.e. finding an overcomplete dictionary together with sparse coordinates. This is a well-studied problem in compressed sensing \citep{10.1109/TIT.2006.871582, 10.1109/TIT.2005.862083}, where sparsity regularizes an otherwise underdetermined problem.

Sparse autoencoders operationalize this idea as one-hidden-layer MLPs trained to reconstruct their input while enforcing sparse hidden activations. Given \(x \in \mathbb{R}^d\), encoder weights \(W_e \in \mathbb{R}^{m \times d}\), decoder weights \(W_d \in \mathbb{R}^{d \times m}\), and biases \(b_e \in \mathbb{R}^m\), \(b_d \in \mathbb{R}^d\), an SAE computes\footnote{We adopt the pre-encoder bias parameterization used by Anthropic \citep{bricken2023towards}, which treats the decoder bias as a constant shift in the data and most closely matches Equation~\ref{eq:decomposition}. SAEs trained without a pre-encoder bias can be converted to this form after training.}
\begin{align}
F(x) &= \tau\!\left(W_e (x-b_d) + b_e\right),\label{eq:encode} \\
(G\circ F)(x) &= W_d F(x) + b_d ,
\label{eq:decode}
\end{align}
where \(\tau\) is a sparsity-inducing nonlinearity such as ReLU \citep{bricken2023towards, scaling_templeton_2024} or JumpReLU \citep{jumping_rajamanoharan_2024}, and \(m \gg d\) makes the representation overcomplete. The columns of \(W_d\) are the learned latents, conventionally constrained to unit norm \citep{olshausen1996emergence, bricken2023towards}; the hidden activations \(F(x)\) provide the sparse coordinate representation of \(x\).

\paragraph{Active Latents and Signatures.}
We call a latent active if \(F(x)_i>0\). The \emph{signature} of an activation is the set of active latent indices. For a collection of activations, such as all activations induced by an input, we define its signature as the union of the individual signatures.

\section{Preliminary Studies}
\label{sec:preliminary_studies}
Our analysis treats sets of active SAE latents, i.e. SAE signatures as the unit of analysis. While prior work motivates such set-level perspectives \citep{park-etal-2025-decoding, Olsen2025Probing, wattenberg2024relational, olah2023distributedcomposition}, it is not obvious that signatures are meaningful objects in their own right. We therefore first test whether they carry interpretable structure. This section summarizes the main observations; detailed results are provided in Appendix~\ref{app:preliminary_studies}.\\
\newcommand{\imgpanel}[5]{%
    \begin{tikzpicture}[baseline=(img.center)]
        \node[inner sep=0pt] (img) {\includegraphics[width=#5]{#1}};
        \node[
            anchor=north east,
            fill=white,
            fill opacity=0.85,
            text opacity=1,
            inner sep=0.4pt,
            font=\fontsize{8}{8}\selectfont\bfseries
        ] at ([xshift=#4,yshift=-#3]img.north west) {#2};
    \end{tikzpicture}%
}

\begin{figure*}[t]
    \centering
    \imgpanel{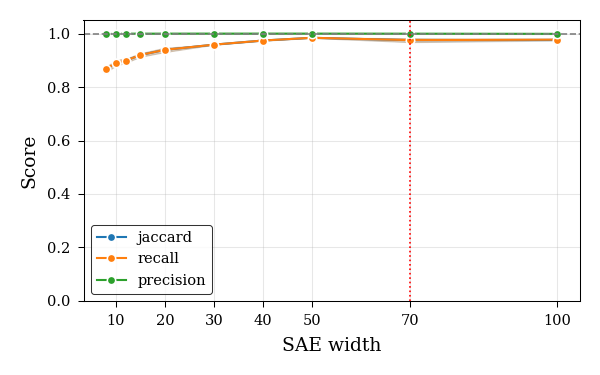}{a)}{2pt}{+8pt}{0.30\textwidth}%
    \hspace{0.02\textwidth}%
    \imgpanel{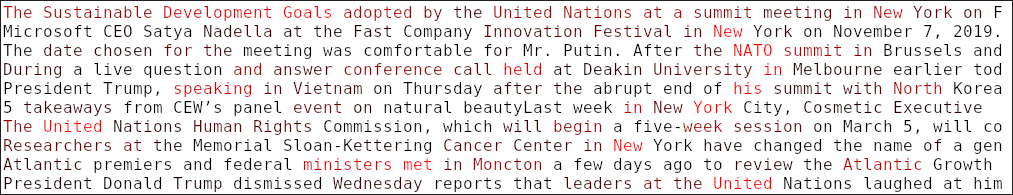}{b)}{2pt}{-2pt}{0.64\textwidth}%

    \caption{Excerpts from the preliminary studies in Appendix~\ref{app:preliminary_studies}. Panel~(a) shows that sufficiently wide SAEs recover set-consistent behaviour with respect to active ground-truth latents. Jaccard, recall, and precision compare the composed-input signature with the union of the individual-feature signatures; the red dotted line marks \(N=70\), the number of generating ground-truth features. Panel~(b) shows the first ten entries of an example cluster from \texttt{The Pile}~\citep{gao2020pile800gbdatasetdiverse}, extracted using SAE latent-set overlap. The cluster groups contexts related to \emph{political conferences}, with highest-overlap tokens often corresponding to the hosting city or organisation.}
    \label{fig:prelim_study}
\end{figure*}
In a toy-model study, we use a controlled substitute for feature composition. Synthetic residual-stream states are generated from known directions, representing a sparse set of ground-truth features. SAEs are trained to reconstruct these synthetic  states. We then compare the SAE signature for a composed input, in which two ground-truth features are active, with the union of the signatures obtained when each feature is presented alone. We call this \emph{union consistency}. Under a bag-of-features view, adding an independent feature should add corresponding latents without removing those already active. Figure~\ref{fig:prelim_study}(a) shows that sufficiently wide SAEs can satisfy this property, with high Jaccard, recall, and precision between the composed signature and the expected union. Details and further experiments are given in Appendix~\ref{app:toy_union_setup}.\\
We then test coherence of the set-level view on natural text. We sample 30,000 snippets from \texttt{The Pile}, compute pairwise distances using a SAE latent set overlap based metric, and construct local neighborhoods around seed snippets. The resulting neighborhoods are often semantically interpretable. An example is shown in Figure~\ref{fig:prelim_study}~(b), with details and further examples in Appendix~\ref{app:interactive_sae_clusters}.\\
These two checks suggest that SAE signatures can support compositional structure in controlled settings and meaningful semantic neighborhoods in real data.

\section{Experiment 1 - Typicality Judgements}
\label{sec:typicality_judgements}
\citet{shani2025tokensthoughtsllmshumans} show that LLM representations, extracted from static embeddings and hidden states, broadly recover human-like category boundaries, but only weakly reproduce the internal associative geometry of human concepts. In particular, model states reflect the grouping of items into categories similarly to humans, while failing to reliably capture graded typicality judgments within categories, such as the intuition that a \emph{robin} is a more typical \emph{bird} than a \emph{penguin}. We closely follow their experimental setup and evaluation procedure, but extend the analysis to sparse feature representations induced by SAEs. Since SAEs are commonly motivated as decomposing model activations into more monosemantic and human-interpretable latent features \citep{scaling_templeton_2024, cunningham2023sparseautoencodershighlyinterpretable, bricken2023towards}, one might expect their active feature sets to expose conceptual structure more directly than dense residual-stream states. We therefore ask two related questions; whether SAE-induced similarities recover human category boundaries, and whether they better capture fine-grained within-category typicality than dense representations. \paragraph{Design} For this analysis, we use the human-concepts dataset introduced by \citet{shani2025tokensthoughtsllmshumans}, based on classic psychology studies of category typicality~\citep{rosch1973internal, rosch1975cognitive, mccloskey1978natural}. The dataset consists of semantic categories, such as \emph{bird}, paired with category instances, such as \emph{hawk} or \emph{penguin}, together with human typicality scores \(s_\text{hum}\) indicating how representative each instance is of its category. To evaluate model-induced category structure, we first define similarity scores between arbitrary word pairs; instance--instance pairs for category-boundary clustering, and category--instance pairs for within-category typicality.\\
\begin{figure*}[t]
    \centering
    \includegraphics[width=\textwidth]{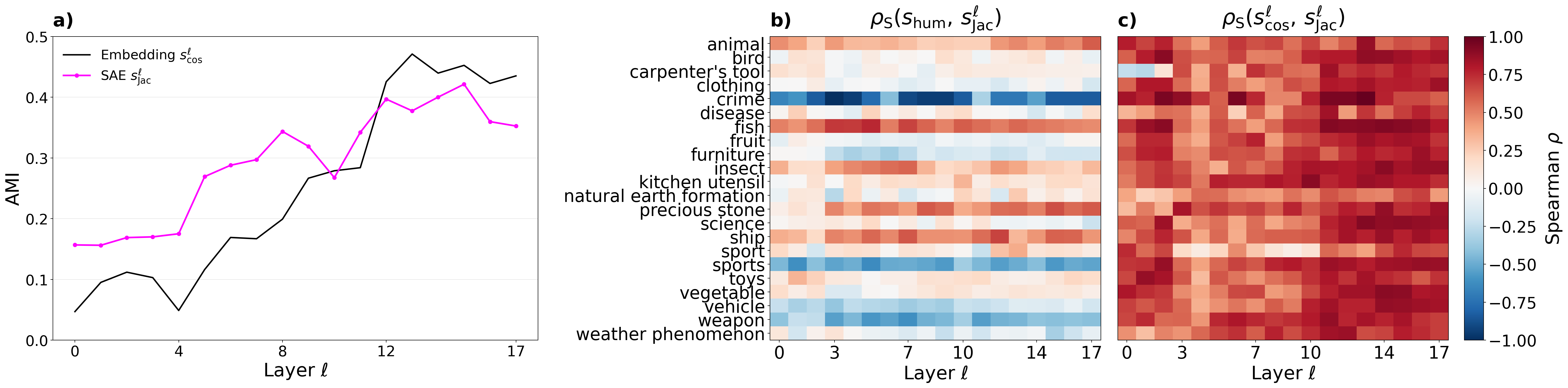}
    \caption{Category-boundary and typicality analyses. Panel~(a) reports AMI-based category-boundary recovery across layers, comparing dense residual-stream cosine similarity \(s_{\mathrm{cos}}^\ell\) from Equation~\eqref{eq:typicality_cosine} with SAE set similarity \(s_{\mathrm{Jac}}^\ell\) from Equation~\eqref{eq:typicality_jaccard}. Panel~(b) shows per-category Spearman correlations between human typicality rankings and SAE-Jaccard rankings. Panel~(c) compares residual-stream cosine rankings with SAE-Jaccard rankings.}
    \label{fig:1}
\end{figure*}
Since words may consist of multiple tokens, we first aggregate token-level representations into a single word-level representation. Following the setup from \citet{shani2025tokensthoughtsllmshumans}, we use static input embeddings for the raw word tokens. For contextual hidden states, we use the neutral prompt template "\texttt{This is a \{word\}.\_}" and evaluate only the token positions corresponding to the inserted word, thereby controlling the context without involving the surrounding prompt in the direct evaluation. Let $T(w)$ denote the token positions of the target word $w$. For contextual representations, $x_t^\ell(w)$ is the residual-stream state of token $t$ at layer $\ell$.\footnote{For static embeddings, $x_t(w)$ is taken directly from the embedding matrix and the layer index is omitted.} We compute the word-level representation by averaging over target-word tokens,
\begin{equation}
    \bar{x}^{\,\ell}(w)
    =
    \frac{1}{|T(w)|}
    \sum_{t \in T(w)} x_t^\ell(w),
    \label{eq:typicality_pooling}
\end{equation}
and use cosine similarity as the model-induced similarity score for two words \(w\) and \(w'\):
\begin{equation}
    s_{\mathrm{cos}}^\ell(w,w')
    =
    \operatorname{cossim}
    \left(
    \bar{x}^{\,\ell}(w),
    \bar{x}^{\,\ell}(w')
    \right).
    \label{eq:typicality_cosine}
\end{equation}
For the SAE-based comparison, we use the same prompted inputs and target-token positions. We encode the corresponding residual states with the SAE, binarize the latent activations, and aggregate all active latents across target-word tokens into a single word-level SAE signature.
\begin{align}
    S^\ell(w)
    &=
    \bigcup_{t \in T(w)}
    \{k \mid z_k^\ell(w,t) > 0\},\\
    z^\ell(w,t)
    &=
    F_\ell(x_t^\ell(w)),
    \label{eq:sae_signature}
\end{align}
where $F$ corresponds to the SAE encoder from Equation~\ref{eq:encode}.
Similarity for SAE signatures is then measured by Jaccard similarity between the corresponding word-level SAE signatures:
\begin{equation}
    s_{\mathrm{Jac}}^\ell(w,w')
    =
    \frac{|S^\ell(w) \cap S^\ell(w')|}
         {|S^\ell(w) \cup S^\ell(w')|}.
    \label{eq:typicality_jaccard}
\end{equation}
With these similarity scores in place, we evaluate category structure at two levels. First, following \citet{shani2025tokensthoughtsllmshumans}, we ask whether the induced geometry recovers human category boundaries. For each representation, we compute pairwise similarities between category instances, convert them to distances \(d=1-s\), cluster instances with \(K\)-medoids using \(K\) equal to the number of ground-truth categories, and compare the resulting partition to human category labels using Adjusted Mutual Information (AMI). AMI is chance-corrected: \(\operatorname{AMI}=1\) indicates perfect agreement with human labels, whereas \(\operatorname{AMI}\approx 0\) indicates chance-level agreement. Further information theoretical motivation on AMI is given by \citet{shani2025tokensthoughtsllmshumans} and a brief introduction is provided in Appendix~\ref{app:category_clustering_ami}.\\
Second, we evaluate within-category typicality. For each semantic category \(c\), the category--instance similarities \(s_{\mathrm{cos}}^\ell(c,i)\) or \(s_{\mathrm{Jac}}^\ell(c,i)\) induce a ranking over instances \(i\). We compare this ranking to the human typicality ranking using Spearman rank correlation. This tests whether representations recover not only coarse category boundaries, but also fine-grained human judgments about which instances are central category members. If SAE latents capture semantically meaningful and human-interpretable factors more directly than dense model states, one might expect \(s_{\mathrm{Jac}}^\ell\) to align more closely with human typicality than cosine similarity over embeddings or residual-stream states.\\
Unless stated otherwise, main-text results use Gemma~3 270M with the 16k \texttt{big} SAE; further models and SAE variants are in the appendix.
\paragraph{Discussion}
\citet{shani2025tokensthoughtsllmshumans} found that LLM representations can recover human category boundaries while remaining largely misaligned with human within-category typicality rankings. We recover the latter result for embedding state similarities and extend the analysis to SAE-induced set similarities. The category-level clustering AMI results show that SAE activation sets do not recover human category boundaries more reliably than dense residual-stream representations, as can be read from Figure~\ref{fig:1}~(a). Interestingly, the highest SAE-based AMI remains below the highest embedding-based AMI, a pattern observed in many investigated settings. Additional AMI results are provided in Appendix~\ref{app:category_clustering_ami}.\\
Similarly, the within-category typicality analysis shows that human typicality rankings and SAE set-based similarity scores are essentially uncorrelated. As shown in Figure~\ref{fig:1}~(b), the per-category Spearman rank coefficients fluctuate around zero for most categories, indicating that Jaccard overlap between SAE signatures does not systematically recover human typicality structure. Additional models and SAE variants are reported in Appendix~\ref{app:typicality_rankings}.\\
Taken together, SAE activation sets do not recover human category understanding more faithfully than dense embedding states, either for coarse category boundaries or for graded within-category typicality. This is surprising given their role as interpretability tools \citep{bricken2023towards, cunningham2023sparseautoencodershighlyinterpretable, scaling_templeton_2024} and the qualitative cluster structure observed in the preliminary studies Section~\ref{sec:preliminary_studies} and Appendix~\ref{app:interactive_sae_clusters}. Instead, SAE-based Jaccard similarities more closely track model-internal similarity structure, as can be seen in Figure~\ref{fig:1}~(c). Since these correlations are not one, however, SAE activations appear to capture a related but partially distinct structure rather than simply reproducing dense-state similarity.

\section{Experiment 2 - Set Consistency under Semantically Related Inputs}
\label{sec:motivation_set_consistency}
Having observed that SAE activation sets align more closely with model-internal similarity structure than with human conceptual intuition, we next examine their compositional consistency under near-identical prompts that differ only by the addition of a single adjective. Motivated by prior work on feature-based representations \citep{park-etal-2025-decoding, Olsen2025Probing, wattenberg2024relational, olah2023distributedcomposition} and by our LRH-aligned toy-model preliminary investigation in Appendix~\ref{app:toy_union_setup}, we ask whether SAE signatures behave compositionally under such refinements. If SAE latents faithfully decompose an activation into its underlying active concepts, adding an adjective should \emph{refine} rather than overwrite the representation. A \emph{shovel} should not become less of a shovel because it is \emph{yellow}; rather, \emph{yellow shovel} should preserve latents associated with \emph{shovel} while additionally activating latents associated with \emph{yellow}. This motivates a \emph{union-of-properties} hypothesis: Composing an object description with a compatible property should approximately expand the active latent signature, so that the composed signature is close to the union of the base-object and adjective signatures. Our preliminary results in Appendix~\ref{app:toy_union_setup} show that SAEs can satisfy such set-consistent behaviour in controlled settings.\footnote{This behavior is structurally unattainable for SAEs with fixed sparsity constraints, such as top-\(k\) SAEs \citep{gao2024scalingevaluatingsparseautoencoders}, where activating new latents necessarily forces others to deactivate in order to maintain constant support size.}
\paragraph{Compositional Consistency}
\label{sec:compositional_consistency}
We construct sequences of increasingly specific prompts by prefixing adjectives to a noun, yielding chains of the form 
\begin{equation}
\begin{aligned}
t_0 &= \text{\texttt{noun}},\\
t_1 &= \text{\texttt{adj}}_1\,\text{\texttt{noun}},\\
&\ \ \dots\\
t_k &= \text{\texttt{adj}}_1 \cdots \text{\texttt{adj}}_k\,\text{\texttt{noun}}.
\end{aligned}
\label{eq:prompt_chain}
\end{equation}
We then compare the SAE signatures induced by these related prompts. Similar to the typicality analysis above, we use the fixed prompt template "\texttt{This is a \{$t_i$\}.\_}", where $t_i$ denotes the phrase obtained at construction step $i$. This fixes the surrounding context while varying only the composed noun phrase. Analogously to Equation~\eqref{eq:sae_signature}, we compute SAE signatures over the token positions corresponding to the "$\texttt{noun}$" word. We construct 200 such sequences from a hand-curated dataset, typically involving adjectives such as color, condition, size, and other descriptive modifiers. For each noun, five adjectives are available, so that $k_{\text{max}}=5$.\\

\subsection{Overview}
\begin{figure}[t]
    \centering
    \includegraphics[width=\columnwidth]{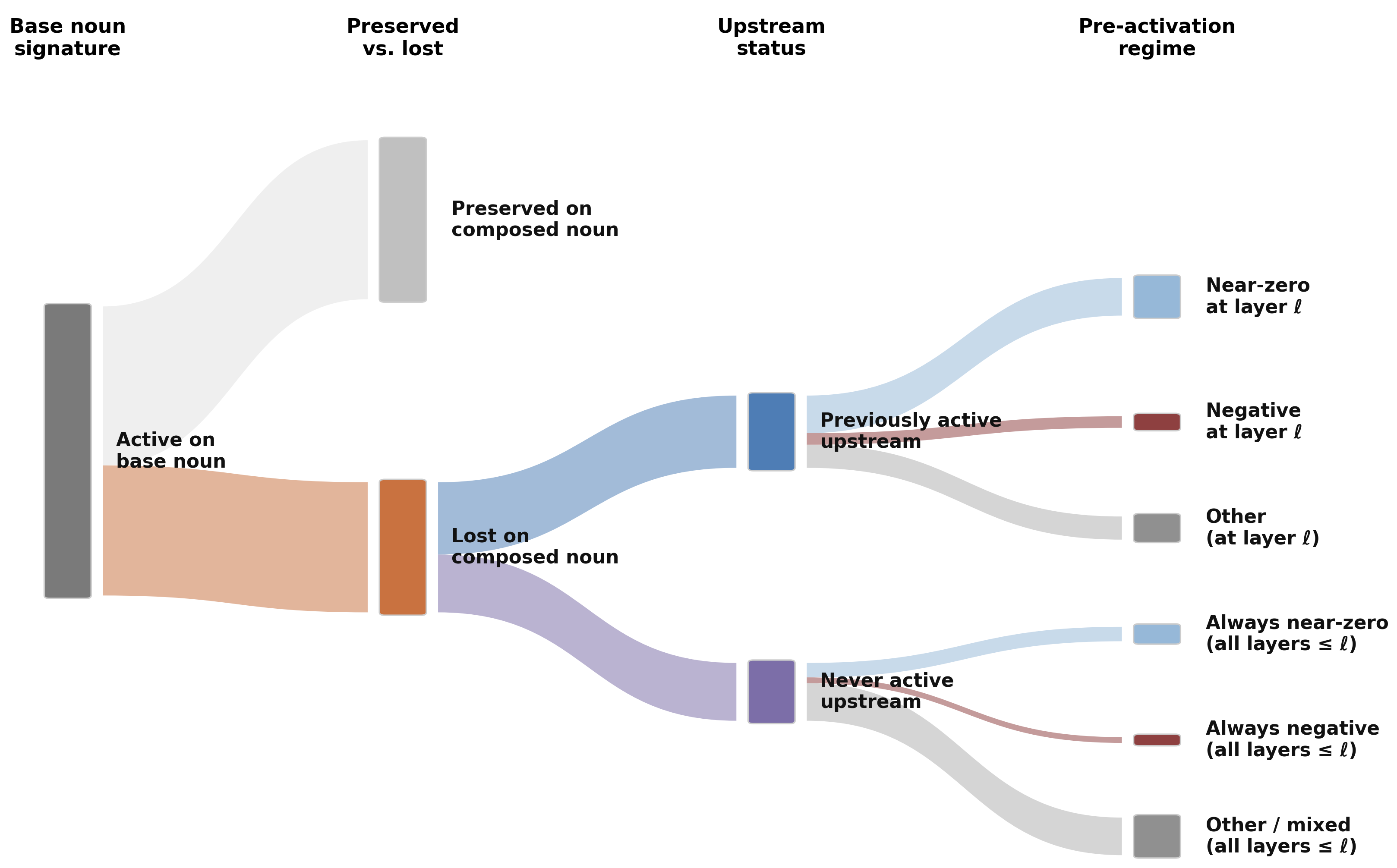}
    \caption{Schematic of steps~(i)--(iii). Lost latents are grouped by upstream presence and pre-activation regime. For previously active latents, pre-activations are evaluated at layer \(l\); for never-previously-active latents, across the upstream trajectory.}
    \label{fig:lost_latent_sankey}
\end{figure}

Our analysis proceeds in four steps; Figure~\ref{fig:lost_latent_sankey} outlines steps~(i)--(iii). Fixed prompt template tokens are excluded from the evaluation.

\paragraph{(i) Measuring lost latents}
We identify SAE latents active for the base prompt \(t_0\) but inactive for the more specific prompt \(t_k\), measuring how much of the target-noun signature is removed by adding compatible adjectives.

\paragraph{(ii) Upstream occurrence}
We ask whether these lost latents appeared earlier in the computation, separating previously active latents from those not detected upstream.

\paragraph{(iii) Pre-activation analysis}
For both groups, we inspect SAE pre-activations before the nonlinearity,
\begin{equation}
    \alpha_{\mathrm{pre}} = W_e (x-b_d) + b_e.
    \label{eq:preactivation}
\end{equation}
We distinguish near-zero latents, indicating weak alignment \citep{improving_rajamanoharan_2024}, from strongly negative pre-activations, which we interpret as stronger suppression \citep{mayne2024can, elhage2021mathematical}.

\paragraph{(iv) Qualitative interpretation}
Finally, we qualitatively discuss which types of latents are lost.

\begin{figure*}[t]
    \centering
    \includegraphics[width=\textwidth]{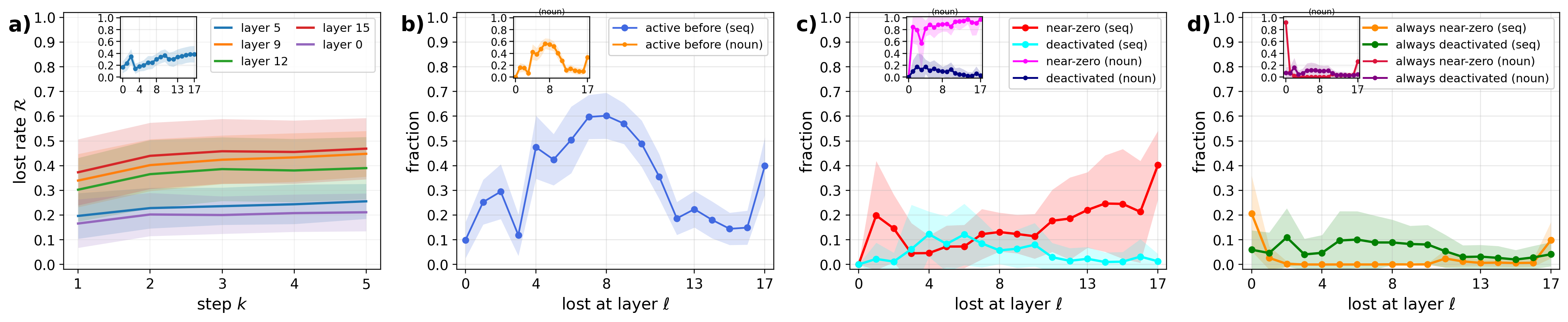}
    \caption{
    Set-consistency analysis for Gemma~3 270M with the 16k \texttt{big} SAE. Panels show dataset means with one-standard-deviation bands. Panel~(a) reports the lost-latent rate \(\mathcal{R}\) from Equation~\eqref{eq:lost_rate} as adjectives are added; the inset shows the layerwise trend for \(k=1\). Panel~(b) shows, among latents lost on the noun signature at layer \(\ell\), the fraction previously active upstream when probing all tokens of the constructed phrase; the inset restricts this check to noun signatures. Panel~(c) decomposes this previously active subset at the layer \(\ell\) where loss is measured, separating near-zero from negative pre-activations over the phrase tokens; the inset shows the noun-only variant. Panel~(d) applies the same near-zero/negative diagnostic to never-previously-active lost latents across the upstream trajectory of the constructed phrase; the inset again restricts to noun signatures. Panels~(b)--(d) use \(k=1\), representative across investigated models as shown in Appendix~\ref{app:k_dependence}. Additional models and SAE variants are shown in Appendix~\ref{app:sweep_lost_feats}.} 
    \label{fig:2}
\end{figure*}

\subsubsection{Set-Consistency Degradation}
\label{sec:degradation_from_compositional_consistency}
We quantify degradation from compositional consistency by comparing the SAE signature of the noun in the base prompt \(t_0\) with the corresponding noun signature in progressively more specific prompts \(t_k\), obtained by prefixing adjectives. Thus, the comparison is always made on the noun token positions. This measures how much of the original noun-related signature is lost as the surrounding description becomes more specific. We define the lost-latent rate \(\mathcal{R}^{(k)}_\ell\) as the fraction of latents active for the base prompt that are no longer active for the more specific prompt: \begin{equation} \mathcal{R}^{(k)}_\ell = \frac{\left|\overline{F}_\ell(t_0)\setminus \overline{F}_\ell(t_k)\right|}{\left|\overline{F}_\ell(t_0)\right|}. \label{eq:lost_rate} \end{equation} Here, \(\overline{F}_\ell(t_k)\) denotes the signature layer \(\ell\) over the noun token positions under the prompt containing \(t_k\). Under the compositionality assumption from Section~\ref{sec:motivation_set_consistency}, this set of latents should largely be preserved when compatible adjectives are added.\\ As shown in Figure~\ref{fig:2}~(a), lost-latent rates are substantial, typically between 20\% and 60\%, increasing with \(k\) and often also with model depth, as indicated by the inset. A more detailed breakdown by model, layer and \(k\) is provided in Appendix~\ref{app:sweep_lost_feats}. The central observation is that adding a semantically compatible adjective can deactivate a large fraction of latents that were active for the less specific noun prompt. Although the resulting signatures are not entirely disjoint, this directly contradicts the simple \emph{union-of-properties} expectation, under which the more specific prompt should preserve object-related noun latents while adding adjective-related ones. It also suggests a possible intuition for why human semantic coherence correlate poorly with SAE activation based distance metrics, as discussed in Section~\ref{sec:typicality_judgements}. This raises the question of why the anticipated behavior fails and what mechanism drives the observed loss of latents.

\subsection{Upstream Presence of Lost Latents}
\label{sec:info_present_upstream}
Having established that SAE set compositional consistency is violated, we relax this strict local criterion and ask whether lost latents were nevertheless recoverable earlier in the computation before disappearing at the inspected noun position.
For latents active on the base prompt \(t_0\) but absent from the noun signature of \(t_k\) at layer \(\ell\), we use the layer-\(\ell\) SAE as a cross-layer detector, since this SAE is known to resolve the lost latent. Following related uses of SAEs as transferable probes \citep{crosscoders_2024, balagansky2025mechanistic}, we apply this detector to residual-stream states from layers \(0,\dots,\ell\). This separates latents that were used earlier and later disappear from the local noun signature from latents not detected along the probed trajectory.\\
Figure~\ref{fig:2}~(b) shows the fraction of lost latents that were active upstream when probing the full sequence. This fraction varies with the layer at which the latent is lost, as it is low in early layers, rises to roughly \(60\%\) in middle layers, and decreases again in later layers. Thus, a substantial subset of lost latents was previously recoverable from the computation, but disappears at layer \(\ell\) at the noun position where loss is measured. The inset restricts the upstream check to noun signatures in previous layers. Its close agreement with the main curve suggests that upstream-active lost latents were usually already active on the noun itself, rather than only elsewhere in the prompt. 
We interpret this to mean that--despite violating compositional consistency in the stricter sense--a large fraction of lost latents remain in use by the model. Their disappearance could as such be a result of SAEs only probing a single point in the residual stream, rather than a failure to resolve composition. However, the results also show that this explanation is insufficient to account for all lost latents.
Additional results are provided in Appendix~\ref{app:sweep_lost_feats}.

\subsection{Pre-activation regimes of lost latents}
\label{sec:what_happened_to_lost_feats}
To characterize lost latents beyond binary activation, we inspect their SAE pre-activations from Equation~\eqref{eq:preactivation}. We attempt to distinguish cases where the latent is inhibited by a strong negative alignment with the residual-stream from cases where it is simply below its activation threshold. To this end we define a near-zero regime,
\begin{equation}
    \alpha_\mathrm{pre}^i \in [-\theta,\theta],
    \label{eq:ortho}
\end{equation}
as well as a negative regime,\color{black}
\begin{equation}
    \alpha_\mathrm{pre}^i < -\theta.
    \label{eq:negative}
\end{equation}
For JumpReLU SAEs, the learned offset provides a natural threshold \(\theta\); for other SAE types, we use the pragmatic choice \(\theta=0.1\), motivated in Appendix~\ref{app:threshold_choice}.\\
For lost latents that were previously active, Figure~\ref{fig:2}~(c) evaluates these regimes at layer \(\ell\), where the loss is observed. Across the full sequence at this layer, the negative fraction is relatively small, typically \(10\%\)--\(20\%\), while the near-zero fraction is larger and tends to increase in later layers. 
Thus, previously active lost latents are not categorically suppressed across the layer-\(\ell\) sequence, nor are they consistently near-zero. On the noun signature, however, the picture is much clearer. Around \(80\%\) of latents fall into the near-zero regime, with only few negative latents.
This suggests that, at the position where loss is measured, active latents are removed, rather than suppressed by a negative component.\color{black}\\
For lost latents that were never previously active, Figure~\ref{fig:2}~(d) applies the same diagnostic across the upstream trajectory, i.e.\ over the full content sequence and all layers up to \(\ell\). These categories test whether a latent is consistently near-zero or consistently negative before the loss point. Both fractions are generally low, indicating that most such latents do not follow a simple pattern of being persistently near zero or persistently in the negative regime. The noun-restricted inset is broadly similar, although the extended sweeps in Appendix~\ref{app:sweep_lost_feats} show more pronounced near-zero behavior on the noun for several models and SAE types, especially in middle layers. 
Overall, binary loss masks a more graded picture: many lost latents are recoverable upstream or show weak rather than negative alignment. Thus, latent disappearance often reflects a local change in the SAE signature rather than a single uniform mechanism of latent removal.\color{black}

\subsection{Qualitative View on Lost Latents}
This raises the question of what could be the driver behind these fluctuations in the latent signature\color{black}. One possibility is that no information is lost in a semantic sense, and that the apparent loss instead reflects geometric interference. Since SAE latents are not exactly orthogonal, adding an adjective may perturb projections onto nearby latent directions, as suggested in Section~\ref{sec:terms_and_defs} and by \citet{scaling_templeton_2024}. However, as discussed in Appendix~\ref{app:interference_vs_neg_feats}, interference alone is unlikely to explain the observed magnitude and structure of lost latents. A second possibility is feature absorption, where the more specific prompt may activate a more specific latent while deactivating a more general one \citep{absorption_chanin_2024}. This explanation also appears incomplete, as if absorption were the dominant mechanism, we would expect qualitatively different lost-latent rates for Gemma Matryoshka SAEs, which are designed to mitigate feature absorption \citep{learning_bussmann_2025}, compared to other SAE families. Instead, the qualitative trends are broadly consistent across SAE types, as shown in Appendix~\ref{app:sweep_lost_feats}.\\
We additionally characterize lost latents using the SAEBench auto-interpretability pipeline \citep{karvonen2025saebench}. As shown in Appendix~\ref{app:autointerp_lost_latents}, lost latents typically receive slightly lower auto-interpretability scores than latents active in both \(t_0\) and \(t_1\). We interpret this as weak evidence that lost latents are less well captured by current automatic explanations. One possible reason is that they participate in more distributed or context-dependent computations, or encode properties whose relevance is tied more to downstream model behavior than to directly verbalizable input patterns.\color{black}

\section{Related Work}
\label{sec:related_work}

\paragraph{Linear Representation Hypothesis}
\label{sec:LRH_critique}
Section~\ref{sec:terms_and_defs} introduced a deliberately minimal formulation of the strong LRH. At the same time, a growing body of work suggests that this picture is incomplete \citep{strong_smith_2024}. In particular, the features recovered by SAEs appear to exhibit geometric structure that is not captured by a simple linear superposition view \citep{sae_mendel_2024, geometry_park_2024}. \\
A related line of work has shown that transformer representations can also organize computation along circular \citep{engels2025not} and helical \citep{language_kantamneni_2025} structures, motivating the broader perspective that some features may inhabit higher-dimensional manifolds rather than isolated linear directions \citep{circuits_research_2024}. In addition, \citet{decomposing_engels_2024} provide evidence that transformer embeddings may contain a pathologically non-linear component.

\paragraph{Feature Composition}
\label{sec:feature_composition}
A cornerstone assumption of mechanistic interpretability is decomposability: Independent features are represented independently within the model \citep{elhage2022toy}. This assumption makes internal analysis tractable \citep{circuits_research_2022}, but how features are organized in practice remains open. Several theoretical frameworks have been proposed for the representational strategies models might use \citep{olah2023distributedcomposition}, alongside mechanistic accounts of how transformers could implement feature composition through architectural operations \citep{wattenberg2024relational}.\\
Empirical findings on SAEs complicate this picture. SAEs exhibit forms of distributed coding that deviate from the idealized decomposable regime, either as side-effects of the training objective \citep{absorption_chanin_2024} or because recovered features are not atomic representational units \citep{sparse_leask_2025, bricken2023towards}. Whether these effects reflect genuine model structure or SAE artifacts remains unsettled. More broadly, it is unclear to what extent a model's hypothesized ``true'' features compose, whether SAEs faithfully recover this structure, or whether they introduce spurious feature combinations absent from the model itself \citep{strong_smith_2024, circuits_research_2024}.

\paragraph{Sparse Autoencoder Variants}
SAE architecture design remains an active research area. TopK \citep{gao2024scalingevaluatingsparseautoencoders} and BatchTopK \citep{batchtopk_bussmann_2024} enforce sparsity directly through the activation mechanism, avoiding an explicit sparsity penalty. Gated \citep{improving_rajamanoharan_2024}, ProReLU \citep{prolu_taggart_2024}, and JumpReLU \citep{jumping_rajamanoharan_2024} SAEs separate interference filtering from latent-magnitude estimation. Matryoshka SAEs \citep{learning_bussmann_2025} instead use grouped training objectives to capture features at multiple levels of conceptual granularity.

\section{Conclusion}
\label{sec:conclusion}

We investigated whether the interpretability of individual SAE latents extends to active latent signatures. Preliminary studies show that these sets can be meaningful: SAEs recover union-like compositional structure in controlled toy models, and latent-set overlap induces semantically coherent neighborhoods in natural text.\\
However, in the setup of \citet{shani2025tokensthoughtsllmshumans}, SAE set similarities do not align with human category boundaries or typicality judgments better than dense model representations. Moreover, semantically compatible refinements, such as adding an adjective to a noun, can deactivate many previously active latents, violating a simple union-of-properties expectation.\\
Overall, our results constrain a naive bag-of-features interpretation of SAE signatures. Active latent sets reveal meaningful structure, but their dynamics are context-sensitive and model-internal, rather than straightforwardly equivalent to human concepts or symbolic semantic feature sets.\color{black}
\section{Limitations}
\label{sec:limitations}

Our analysis covers multiple model scales and SAE variants, from 270M to 27B parameters, with broadly consistent qualitative trends. Still, the conclusions are tied to the datasets, semantic constructions, and SAE families considered here. The typicality experiments rely on human psychology datasets and are therefore limited to their concepts, items, and judgment structure. The lost-latent experiments study one controlled form of semantic refinement, adjective modification of nouns, and may not directly generalize to paraphrases, task-relevant abstractions, or other forms of semantic consistency. Finally, we focus primarily on residual-stream SAEs and do not exhaustively evaluate other hook locations or SAE families, such as BatchTopK or gated SAEs. Whether the same set-level instabilities hold in these settings remains open. See Appendix~\ref{sec:resources_ai_use} for resources and AI-use disclosure.

\bibliography{custom}

@inproceedings{language_kantamneni_2025,
    title={Language Models Use Trigonometry to Do Addition},
    author={Subhash Kantamneni and Max Tegmark},
    booktitle={ICLR 2025 Workshop on Building Trust in Language Models and Applications},
    year={2025},
    url={https://openreview.net/forum?id=CqViN4dQJk}
}

@article{
decomposing_engels_2024,
title={Decomposing The Dark Matter of Sparse Autoencoders},
author={Joshua Engels and Logan Riggs Smith and Max Tegmark},
journal={Transactions on Machine Learning Research},
issn={2835-8856},
year={2025},
url={https://openreview.net/forum?id=sXq3Wb3vef},
note={}
}

@article{circuits_research_2024,
	title = {Circuits Updates - July 2024},
	author = {Olah, Chris and Jermyn, Adam},
	journal = {Transformer Circuits Thread},
	year = {2024},
    url = {https://transformer-circuits.pub/2024/july-update/index.htmls},
}

@article{circuits_research_2022,
	title = {Mechanistic Interpretability, Variables, and the Importance of Interpretable Bases},
	author = {Olah, Chris},
	journal = {Transformer Circuits Thread},
	year = {2022},
    url = {https://transformer-circuits.pub/2022/mech-interp-essay/index.html},
}

@article{olah2023distributedcomposition,
	title = {Distributed Representations: Composition \& Superposition},
	author = {Olah, Chris},
	journal = {Transformer Circuits Thread},
	year = {2024},
    url = {https://transformer-circuits.pub/2023/superposition-composition},
	litmapsId = {1377773}
}

@misc{jumping_rajamanoharan_2024,
	title = {Jumping Ahead: Improving Reconstruction Fidelity with JumpReLU Sparse Autoencoders},
	doi = {10.48550/arxiv.2407.14435},
	author = {Rajamanoharan, Senthooran and Lieberum, Tom and Sonnerat, Nicolas and Conmy, Arthur and Varma, Vikrant and Kramár, János and Nanda, Neel},
	year = {2024},
	eprint={2407.14435},
    archivePrefix={arXiv},
}

@article{prolu_taggart_2024,
	title = {ProLU: A Nonlinearity for Sparse Autoencoders},
	author = {Taggart, Glen},
	journal = {LessWrong},
    url = {https://www.lesswrong.com/posts/HEpufTdakGTTKgoYF/prolu-a-pareto-improvement-for-sparse-autoencoders},
	year = {2024},
	litmapsId = {1451841}
}

@article{strong_smith_2024,
	title = {The ‘strong’ feature hypothesis could be wrong},
	author = {Smith, Lewis},
	year = {2024},
    journal = {LessWrong},
    url = {https://www.lesswrong.com/posts/tojtPCCRpKLSHBdpn/the-strong-feature-hypothesis-could-be-wrong},
	litmapsId = {1451847}
}

@article{sae_mendel_2024,
	title = {SAE feature geometry is outside the superposition hypothesis},
	author = {Mendel, Jake},
	journal = {AI Alignment Forum},
	year = {2024},
    url = {https://www.alignmentforum.org/posts/MFBTjb2qf3ziWmzz6/sae-feature-geometry-is-outside-the-superposition-hypothesis},
	litmapsId = {1377771}
}

@inproceedings{
    geometry_park_2024,
    title={The Geometry of Categorical and Hierarchical Concepts in Large Language Models},
    author={Kiho Park and Yo Joong Choe and Yibo Jiang and Victor Veitch},
    booktitle={The Thirteenth International Conference on Learning Representations},
    year={2025},
    url={https://openreview.net/forum?id=bVTM2QKYuA}
}

@article{scaling_templeton_2024,
    title = {Scaling Monosemanticity: Extracting Interpretable Features from {Claude 3 Sonnet}},
    author = {Templeton, Adly and Conerly, Tom and Marcus, Jonathan and Lindsey, Jack and Bricken, Trenton and Chen, Brian and Pearce, Adam and Citro, Craig and Ameisen, Emmanuel and Jones, Andy and Cunningham, Hoagy and Turner, Nicholas L and McDougall, Callum and MacDiarmid, Monte and Tamkin, Alex and Durmus, Esin and Hume, Tristan and Mosconi, Francesco and Freeman, C. Daniel and Sumers, Theodore R. and Rees, Edward and Batson, Joshua and Jermyn, Adam and Carter, Shan and Olah, Chris and Henighan, Tom},
    journal = {Transformer Circuits Thread},
    year = {2024},
    url = {https://transformer-circuits.pub/2024/scaling-monosemanticity/#related-work},
    litmapsId = {1359871}
}

@inproceedings{batchtopk_bussmann_2024,
	title = {{BatchTopK} Sparse Autoencoders},
    author={Bart Bussmann and Patrick Leask and Neel Nanda},
    booktitle={NeurIPS 2024 Workshop on Scientific Methods for Understanding Deep Learning},
    year={2024},
    url={https://openreview.net/forum?id=d4dpOCqybL}
}

@inproceedings{improving_rajamanoharan_2024,
title={Improving Sparse Decomposition of Language Model Activations with Gated Sparse Autoencoders},
author={Senthooran Rajamanoharan and Arthur Conmy and Lewis Smith and Tom Lieberum and Vikrant Varma and Janos Kramar and Rohin Shah and Neel Nanda},
booktitle={The Thirty-eighth Annual Conference on Neural Information Processing Systems},
year={2024},
url={https://openreview.net/forum?id=zLBlin2zvW}
}

@inproceedings{learning_bussmann_2025,
    title = {Learning Multi-Level Features with Matryoshka Sparse Autoencoders},
    doi = {10.48550/arxiv.2503.17547},
    author = {Bussmann, Bart and Nabeshima, Noa and Karvonen, Adam and Nanda, Neel},
    booktitle = {Proceedings of the 42nd International Conference on Machine Learning},
    series = {ICML'25},
    year = {2025},
    litmapsId = {285786464}
}

@inproceedings{
sparse_leask_2025,
title={Sparse Autoencoders Do Not Find Canonical Units of Analysis},
author={Patrick Leask and Bart Bussmann and Michael T Pearce and Joseph Isaac Bloom and Curt Tigges and Noura Al Moubayed and Lee Sharkey and Neel Nanda},
booktitle={The Thirteenth International Conference on Learning Representations},
year={2025},
url={https://openreview.net/forum?id=9ca9eHNrdH}
}

@inproceedings{absorption_chanin_2024,
    title = {A is for Absorption: Studying Feature Splitting and Absorption in Sparse Autoencoders},
    doi = {10.48550/arxiv.2409.14507},
    author = {Chanin, David and Wilken-Smith, James and Dulka, Tom\'{a}\v{s} and Bhatnagar, Hardik and Bloom, Joseph},
    booktitle = {Advances in Neural Information Processing Systems},
    volume = {38},
    year = {2024},
    litmapsId = {280231508}
}

@misc{feature_chanin_2025,
	title = {Feature Hedging: Correlated Features Break Narrow Sparse Autoencoders},
	eprint = {2505.11756},
	author = {Chanin, David and Dulka, Tomáš and Garriga-Alonso, Adrià},
	archivePrefix={arXiv},
	year = {2025},
	litmapsId = {287522554}
}

@ARTICLE{6771362,
  author={Wyner, A. D.},
  journal={The Bell System Technical Journal}, 
  title={Random packings and coverings of the unit n-sphere}, 
  year={1967},
  volume={46},
  number={9},
  pages={2111-2118},
  doi={10.1002/j.1538-7305.1967.tb04246.x}}

@article{ml_math_2023,
	title = {Some ML-Related Math I Now Understand Better},
	author = {Roger, Fabien},
	journal = {LessWrong},
    url = {https://www.lesswrong.com/posts/Tu8DYNCg63F4HYbXE/some-ml-related-math-i-now-understand-better},
	year = {2023}
}

@inproceedings{
shani2025tokensthoughtsllmshumans,
title={From Tokens to Thoughts: How {LLM}s and Humans Trade Compression for Meaning},
author={Chen Shani and Liron Soffer and Dan Jurafsky and Yann LeCun and Ravid Shwartz-Ziv},
booktitle={The Fourteenth International Conference on Learning Representations},
year={2026},
url={https://openreview.net/forum?id=rkthPeHvAX}
}

@inproceedings{10.5555/3692070.3693675,
author = {Park, Kiho and Choe, Yo Joong and Veitch, Victor},
title = {The linear representation hypothesis and the geometry of large language models},
year = {2024},
publisher = {JMLR.org},
booktitle = {Proceedings of the 41st International Conference on Machine Learning},
articleno = {1605},
numpages = {24},
location = {Vienna, Austria},
series = {ICML'24},
url = {https://dl.acm.org/doi/10.5555/3692070.3693675}
}

@article{elhage2022toy,
   title={Toy Models of Superposition},
   author={Elhage, Nelson and Hume, Tristan and Olsson, Catherine and Schiefer, Nicholas and Henighan, Tom and Kravec, Shauna and Hatfield-Dodds, Zac and Lasenby, Robert and Drain, Dawn and Chen, Carol and Grosse, Roger and McCandlish, Sam and Kaplan, Jared and Amodei, Dario and Wattenberg, Martin and Olah, Christopher},
   year={2022},
   journal={Transformer Circuits Thread},
   url={https://transformer-circuits.pub/2022/toy\_model/index.html}
}

@article{bricken2023towards,
   title={Towards Monosemanticity: Decomposing Language Models With Dictionary Learning},
   author={Bricken, Trenton and Templeton, Adina and Batson, Joshua and Chen, Brian and Jermyn, Adam and Conerly, Tom and Turner, Nick and Anil, Cem and Cheng, Cheng and Kravec, Sasha and Schiefer, Nicholas and Hume, Tristan and Engstrom, Logan and Olah, Chris and others},
   year={2023},
   journal={Transformer Circuits Thread},
   url={https://transformer-circuits.pub/2023/monosemantic-features/index.html}
}

@inproceedings{cunningham2023sparseautoencodershighlyinterpretable,
    title = {Sparse Autoencoders Find Highly Interpretable Features in Language Models}, 
    author = {Hoagy Cunningham and Aidan Ewart and Logan Riggs and Robert Huben and Lee Sharkey},
    booktitle = {The Twelfth International Conference on Learning Representations},
    year = {2024},
    url = {https://openreview.net/forum?id=F76bwRSLeK}
}

@article{10.1109/TIT.2006.871582,
author = {Donoho, D. L.},
title = {Compressed sensing},
year = {2006},
issue_date = {April 2006},
publisher = {IEEE Press},
volume = {52},
number = {4},
issn = {0018-9448},
url = {https://doi.org/10.1109/TIT.2006.871582},
doi = {10.1109/TIT.2006.871582},
journal = {IEEE Trans. Inf. Theor.},
month = apr,
pages = {1289–1306},
numpages = {18}
}

@article{10.1109/TIT.2005.862083,
author = {Candes, E. J. and Romberg, J. and Tao, T.},
title = {Robust uncertainty principles: exact signal reconstruction from highly incomplete frequency information},
year = {2006},
issue_date = {February 2006},
publisher = {IEEE Press},
volume = {52},
number = {2},
issn = {0018-9448},
url = {https://doi.org/10.1109/TIT.2005.862083},
doi = {10.1109/TIT.2005.862083},
journal = {IEEE Trans. Inf. Theor.},
month = feb,
pages = {489–509},
numpages = {21}
}

@article{olshausen1996emergence,
  title={Emergence of simple-cell receptive field properties by learning a sparse code for natural images},
  author={Olshausen, Bruno A and Field, David J},
  journal={Nature},
  volume={381},
  number={6583},
  pages={607--609},
  year={1996},
  publisher={Nature Publishing Group}
}

@inproceedings{
wattenberg2024relational,
title={Relational Composition in Neural Networks: A Survey and Call to Action},
author={Martin Wattenberg and Fernanda Vi{\'e}gas},
booktitle={ICML 2024 Workshop on Mechanistic Interpretability},
year={2024},
url={https://openreview.net/forum?id=zzCEiUIPk9}
}

@inproceedings{NEURIPS2022_c32319f4,
 author = {Kusupati, Aditya and Bhatt, Gantavya and Rege, Aniket and Wallingford, Matthew and Sinha, Aditya and Ramanujan, Vivek and Howard-Snyder, William and Chen, Kaifeng and Kakade, Sham and Jain, Prateek and Farhadi, Ali},
 booktitle = {Advances in Neural Information Processing Systems},
 editor = {S. Koyejo and S. Mohamed and A. Agarwal and D. Belgrave and K. Cho and A. Oh},
 pages = {30233--30249},
 publisher = {Curran Associates, Inc.},
 title = {Matryoshka Representation Learning},
 url = {https://proceedings.neurips.cc/paper_files/paper/2022/file/c32319f4868da7613d78af9993100e42-Paper-Conference.pdf},
 volume = {35},
 year = {2022}
}

@inproceedings{lieberum-etal-2024-gemma,
    title = "Gemma Scope: Open Sparse Autoencoders Everywhere All At Once on Gemma 2",
    author = "Lieberum, Tom  and
      Rajamanoharan, Senthooran  and
      Conmy, Arthur  and
      Smith, Lewis  and
      Sonnerat, Nicolas  and
      Varma, Vikrant  and
      Kramar, Janos  and
      Dragan, Anca  and
      Shah, Rohin  and
      Nanda, Neel",
    editor = "Belinkov, Yonatan  and
      Kim, Najoung  and
      Jumelet, Jaap  and
      Mohebbi, Hosein  and
      Mueller, Aaron  and
      Chen, Hanjie",
    booktitle = "Proceedings of the 7th BlackboxNLP Workshop: Analyzing and Interpreting Neural Networks for NLP",
    month = nov,
    year = "2024",
    address = "Miami, Florida, US",
    publisher = "Association for Computational Linguistics",
    url = "https://aclanthology.org/2024.blackboxnlp-1.19/",
    doi = "10.18653/v1/2024.blackboxnlp-1.19",
    pages = "278--300"
}

@inproceedings{
gao2024scalingevaluatingsparseautoencoders,
title={Scaling and evaluating sparse autoencoders},
author={Leo Gao and Tom Dupre la Tour and Henk Tillman and Gabriel Goh and Rajan Troll and Alec Radford and Ilya Sutskever and Jan Leike and Jeffrey Wu},
booktitle={The Thirteenth International Conference on Learning Representations},
year={2025},
url={https://openreview.net/forum?id=tcsZt9ZNKD}
}

@misc{he2024llamascopeextractingmillions,
      title={Llama Scope: Extracting Millions of Features from Llama-3.1-8B with Sparse Autoencoders}, 
      author={Zhengfu He and Wentao Shu and Xuyang Ge and Lingjie Chen and Junxuan Wang and Yunhua Zhou and Frances Liu and Qipeng Guo and Xuanjing Huang and Zuxuan Wu and Yu-Gang Jiang and Xipeng Qiu},
      year={2024},
      eprint={2410.20526},
      archivePrefix={arXiv},
      primaryClass={cs.LG},
      url={https://arxiv.org/abs/2410.20526},
}

@techreport{google2024gemmascope2,
  title={Gemma Scope 2 - Technical Paper},
  author={Callum McDougall and Arthur Conmy and János Kramár and Tom Lieberum and Senthooran Rajamanoharan and Neel Nanda},
  year={2025},
  institution={Google DeepMind},
  url={https://storage.googleapis.com/deepmind-media/DeepMind.com/Blog/gemma-scope-2-helping-the-ai-safety-community-deepen-understanding-of-complex-language-model-behavior/Gemma_Scope_2_Technical_Paper.pdf}
}

@misc{gao2020pile800gbdatasetdiverse,
      title={The Pile: An 800GB Dataset of Diverse Text for Language Modeling}, 
      author={Leo Gao and Stella Biderman and Sid Black and Laurence Golding and Travis Hoppe and Charles Foster and Jason Phang and Horace He and Anish Thite and Noa Nabeshima and Shawn Presser and Connor Leahy},
      year={2020},
      eprint={2101.00027},
      archivePrefix={arXiv},
      primaryClass={cs.CL},
      note={\url{https://arxiv.org/abs/2101.00027}}, 
}

@inproceedings{
engels2025not,
title={Not All Language Model Features Are One-Dimensionally Linear},
author={Joshua Engels and Eric J Michaud and Isaac Liao and Wes Gurnee and Max Tegmark},
booktitle={The Thirteenth International Conference on Learning Representations},
year={2025},
url={https://openreview.net/forum?id=d63a4AM4hb}
}

@techreport{gemmateam2025gemma3technicalreport,
      title={Gemma 3 Technical Report}, 
      author={{Gemma Team} and Aishwarya Kamath and Johan Ferret and Shreya Pathak and Nino Vieillard and Ramona Merhej and Sarah Perrin and Tatiana Matejovicova and Alexandre Ramé and Morgane Rivière and Louis Rouillard and Thomas Mesnard and Geoffrey Cideron and Jean-bastien Grill and Sabela Ramos and Edouard Yvinec and Michelle Casbon and Etienne Pot and Ivo Penchev and Gaël Liu and Francesco Visin and Kathleen Kenealy and Lucas Beyer and Xiaohai Zhai and Anton Tsitsulin and Robert Busa-Fekete and Alex Feng and Noveen Sachdeva and Benjamin Coleman and Yi Gao and Basil Mustafa and Iain Barr and Emilio Parisotto and David Tian and Matan Eyal and Colin Cherry and Jan-Thorsten Peter and Danila Sinopalnikov and Surya Bhupatiraju and Rishabh Agarwal and Mehran Kazemi and Dan Malkin and Ravin Kumar and David Vilar and Idan Brusilovsky and Jiaming Luo and Andreas Steiner and Abe Friesen and Abhanshu Sharma and Abheesht Sharma and Adi Mayrav Gilady and Adrian Goedeckemeyer and Alaa Saade and Alex Feng and Alexander Kolesnikov and Alexei Bendebury and Alvin Abdagic and Amit Vadi and András György and André Susano Pinto and Anil Das and Ankur Bapna and Antoine Miech and Antoine Yang and Antonia Paterson and Ashish Shenoy and Ayan Chakrabarti and Bilal Piot and Bo Wu and Bobak Shahriari and Bryce Petrini and Charlie Chen and Charline Le Lan and Christopher A. Choquette-Choo and CJ Carey and Cormac Brick and Daniel Deutsch and Danielle Eisenbud and Dee Cattle and Derek Cheng and Dimitris Paparas and Divyashree Shivakumar Sreepathihalli and Doug Reid and Dustin Tran and Dustin Zelle and Eric Noland and Erwin Huizenga and Eugene Kharitonov and Frederick Liu and Gagik Amirkhanyan and Glenn Cameron and Hadi Hashemi and Hanna Klimczak-Plucińska and Harman Singh and Harsh Mehta and Harshal Tushar Lehri and Hussein Hazimeh and Ian Ballantyne and Idan Szpektor and Ivan Nardini and Jean Pouget-Abadie and Jetha Chan and Joe Stanton and John Wieting and Jonathan Lai and Jordi Orbay and Joseph Fernandez and Josh Newlan and Ju-yeong Ji and Jyotinder Singh and Kat Black and Kathy Yu and Kevin Hui and Kiran Vodrahalli and Klaus Greff and Linhai Qiu and Marcella Valentine and Marina Coelho and Marvin Ritter and Matt Hoffman and Matthew Watson and Mayank Chaturvedi and Michael Moynihan and Min Ma and Nabila Babar and Natasha Noy and Nathan Byrd and Nick Roy and Nikola Momchev and Nilay Chauhan and Noveen Sachdeva and Oskar Bunyan and Pankil Botarda and Paul Caron and Paul Kishan Rubenstein and Phil Culliton and Philipp Schmid and Pier Giuseppe Sessa and Pingmei Xu and Piotr Stanczyk and Pouya Tafti and Rakesh Shivanna and Renjie Wu and Renke Pan and Reza Rokni and Rob Willoughby and Rohith Vallu and Ryan Mullins and Sammy Jerome and Sara Smoot and Sertan Girgin and Shariq Iqbal and Shashir Reddy and Shruti Sheth and Siim Põder and Sijal Bhatnagar and Sindhu Raghuram Panyam and Sivan Eiger and Susan Zhang and Tianqi Liu and Trevor Yacovone and Tyler Liechty and Uday Kalra and Utku Evci and Vedant Misra and Vincent Roseberry and Vlad Feinberg and Vlad Kolesnikov and Woohyun Han and Woosuk Kwon and Xi Chen and Yinlam Chow and Yuvein Zhu and Zichuan Wei and Zoltan Egyed and Victor Cotruta and Minh Giang and Phoebe Kirk and Anand Rao and Kat Black and Nabila Babar and Jessica Lo and Erica Moreira and Luiz Gustavo Martins and Omar Sanseviero and Lucas Gonzalez and Zach Gleicher and Tris Warkentin and Vahab Mirrokni and Evan Senter and Eli Collins and Joelle Barral and Zoubin Ghahramani and Raia Hadsell and Yossi Matias and D. Sculley and Slav Petrov and Noah Fiedel and Noam Shazeer and Oriol Vinyals and Jeff Dean and Demis Hassabis and Koray Kavukcuoglu and Clement Farabet and Elena Buchatskaya and Jean-Baptiste Alayrac and Rohan Anil and Dmitry and Lepikhin and Sebastian Borgeaud and Olivier Bachem and Armand Joulin and Alek Andreev and Cassidy Hardin and Robert Dadashi and Léonard Hussenot},
      year={2025},
      eprint={2503.19786},
      archivePrefix={arXiv},
      primaryClass={cs.CL},
      institution = {Google DeepMind}
}

@misc{grattafiori2024llama3herdmodels,
      title={The Llama 3 Herd of Models}, 
      author={Aaron Grattafiori and Abhimanyu Dubey and Abhinav Jauhri and Abhinav Pandey and Abhishek Kadian and Ahmad Al-Dahle and Aiesha Letman and Akhil Mathur and Alan Schelten and Alex Vaughan and Amy Yang and Angela Fan and Anirudh Goyal and Anthony Hartshorn and Aobo Yang and Archi Mitra and Archie Sravankumar and Artem Korenev and Arthur Hinsvark and Arun Rao and Aston Zhang and Aurelien Rodriguez and Austen Gregerson and Ava Spataru and Baptiste Roziere and Bethany Biron and Binh Tang and Bobbie Chern and Charlotte Caucheteux and Chaya Nayak and Chloe Bi and Chris Marra and Chris McConnell and Christian Keller and Christophe Touret and Chunyang Wu and Corinne Wong and Cristian Canton Ferrer and Cyrus Nikolaidis and Damien Allonsius and Daniel Song and Danielle Pintz and Danny Livshits and Danny Wyatt and David Esiobu and Dhruv Choudhary and Dhruv Mahajan and Diego Garcia-Olano and Diego Perino and Dieuwke Hupkes and Egor Lakomkin and Ehab AlBadawy and Elina Lobanova and Emily Dinan and Eric Michael Smith and Filip Radenovic and Francisco Guzmán and Frank Zhang and Gabriel Synnaeve and Gabrielle Lee and Georgia Lewis Anderson and Govind Thattai and Graeme Nail and Gregoire Mialon and Guan Pang and Guillem Cucurell and Hailey Nguyen and Hannah Korevaar and Hu Xu and Hugo Touvron and Iliyan Zarov and Imanol Arrieta Ibarra and Isabel Kloumann and Ishan Misra and Ivan Evtimov and Jack Zhang and Jade Copet and Jaewon Lee and Jan Geffert and Jana Vranes and Jason Park and Jay Mahadeokar and Jeet Shah and Jelmer van der Linde and Jennifer Billock and Jenny Hong and Jenya Lee and Jeremy Fu and Jianfeng Chi and Jianyu Huang and Jiawen Liu and Jie Wang and Jiecao Yu and Joanna Bitton and Joe Spisak and Jongsoo Park and Joseph Rocca and Joshua Johnstun and Joshua Saxe and Junteng Jia and Kalyan Vasuden Alwala and Karthik Prasad and Kartikeya Upasani and Kate Plawiak and Ke Li and Kenneth Heafield and Kevin Stone and Khalid El-Arini and Krithika Iyer and Kshitiz Malik and Kuenley Chiu and Kunal Bhalla and Kushal Lakhotia and Lauren Rantala-Yeary and Laurens van der Maaten and Lawrence Chen and Liang Tan and Liz Jenkins and Louis Martin and Lovish Madaan and Lubo Malo and Lukas Blecher and Lukas Landzaat and Luke de Oliveira and Madeline Muzzi and Mahesh Pasupuleti and Mannat Singh and Manohar Paluri and Marcin Kardas and Maria Tsimpoukelli and Mathew Oldham and Mathieu Rita and Maya Pavlova and Melanie Kambadur and Mike Lewis and Min Si and Mitesh Kumar Singh and Mona Hassan and Naman Goyal and Narjes Torabi and Nikolay Bashlykov and Nikolay Bogoychev and Niladri Chatterji and Ning Zhang and Olivier Duchenne and Onur Çelebi and Patrick Alrassy and Pengchuan Zhang and Pengwei Li and Petar Vasic and Peter Weng and Prajjwal Bhargava and Pratik Dubal and Praveen Krishnan and Punit Singh Koura and Puxin Xu and Qing He and Qingxiao Dong and Ragavan Srinivasan and Raj Ganapathy and Ramon Calderer and Ricardo Silveira Cabral and Robert Stojnic and Roberta Raileanu and Rohan Maheswari and Rohit Girdhar and Rohit Patel and Romain Sauvestre and Ronnie Polidoro and Roshan Sumbaly and Ross Taylor and Ruan Silva and Rui Hou and Rui Wang and Saghar Hosseini and Sahana Chennabasappa and Sanjay Singh and Sean Bell and Seohyun Sonia Kim and Sergey Edunov and Shaoliang Nie and Sharan Narang and Sharath Raparthy and Sheng Shen and Shengye Wan and Shruti Bhosale and Shun Zhang and Simon Vandenhende and Soumya Batra and Spencer Whitman and Sten Sootla and Stephane Collot and Suchin Gururangan and Sydney Borodinsky and Tamar Herman and Tara Fowler and Tarek Sheasha and Thomas Georgiou and Thomas Scialom and Tobias Speckbacher and Todor Mihaylov and Tong Xiao and Ujjwal Karn and Vedanuj Goswami and Vibhor Gupta and Vignesh Ramanathan and Viktor Kerkez and Vincent Gonguet and Virginie Do and Vish Vogeti and Vítor Albiero and Vladan Petrovic and Weiwei Chu and Wenhan Xiong and Wenyin Fu and Whitney Meers and Xavier Martinet and Xiaodong Wang and Xiaofang Wang and Xiaoqing Ellen Tan and Xide Xia and Xinfeng Xie and Xuchao Jia and Xuewei Wang and Yaelle Goldschlag and Yashesh Gaur and Yasmine Babaei and Yi Wen and Yiwen Song and Yuchen Zhang and Yue Li and Yuning Mao and Zacharie Delpierre Coudert and Zheng Yan and Zhengxing Chen and Zoe Papakipos and Aaditya Singh and Aayushi Srivastava and Abha Jain and Adam Kelsey and Adam Shajnfeld and Adithya Gangidi and Adolfo Victoria and Ahuva Goldstand and Ajay Menon and Ajay Sharma and Alex Boesenberg and Alexei Baevski and Allie Feinstein and Amanda Kallet and Amit Sangani and Amos Teo and Anam Yunus and Andrei Lupu and Andres Alvarado and Andrew Caples and Andrew Gu and Andrew Ho and Andrew Poulton and Andrew Ryan and Ankit Ramchandani and Annie Dong and Annie Franco and Anuj Goyal and Aparajita Saraf and Arkabandhu Chowdhury and Ashley Gabriel and Ashwin Bharambe and Assaf Eisenman and Azadeh Yazdan and Beau James and Ben Maurer and Benjamin Leonhardi and Bernie Huang and Beth Loyd and Beto De Paola and Bhargavi Paranjape and Bing Liu and Bo Wu and Boyu Ni and Braden Hancock and Bram Wasti and Brandon Spence and Brani Stojkovic and Brian Gamido and Britt Montalvo and Carl Parker and Carly Burton and Catalina Mejia and Ce Liu and Changhan Wang and Changkyu Kim and Chao Zhou and Chester Hu and Ching-Hsiang Chu and Chris Cai and Chris Tindal and Christoph Feichtenhofer and Cynthia Gao and Damon Civin and Dana Beaty and Daniel Kreymer and Daniel Li and David Adkins and David Xu and Davide Testuggine and Delia David and Devi Parikh and Diana Liskovich and Didem Foss and Dingkang Wang and Duc Le and Dustin Holland and Edward Dowling and Eissa Jamil and Elaine Montgomery and Eleonora Presani and Emily Hahn and Emily Wood and Eric-Tuan Le and Erik Brinkman and Esteban Arcaute and Evan Dunbar and Evan Smothers and Fei Sun and Felix Kreuk and Feng Tian and Filippos Kokkinos and Firat Ozgenel and Francesco Caggioni and Frank Kanayet and Frank Seide and Gabriela Medina Florez and Gabriella Schwarz and Gada Badeer and Georgia Swee and Gil Halpern and Grant Herman and Grigory Sizov and Guangyi and Zhang and Guna Lakshminarayanan and Hakan Inan and Hamid Shojanazeri and Han Zou and Hannah Wang and Hanwen Zha and Haroun Habeeb and Harrison Rudolph and Helen Suk and Henry Aspegren and Hunter Goldman and Hongyuan Zhan and Ibrahim Damlaj and Igor Molybog and Igor Tufanov and Ilias Leontiadis and Irina-Elena Veliche and Itai Gat and Jake Weissman and James Geboski and James Kohli and Janice Lam and Japhet Asher and Jean-Baptiste Gaya and Jeff Marcus and Jeff Tang and Jennifer Chan and Jenny Zhen and Jeremy Reizenstein and Jeremy Teboul and Jessica Zhong and Jian Jin and Jingyi Yang and Joe Cummings and Jon Carvill and Jon Shepard and Jonathan McPhie and Jonathan Torres and Josh Ginsburg and Junjie Wang and Kai Wu and Kam Hou U and Karan Saxena and Kartikay Khandelwal and Katayoun Zand and Kathy Matosich and Kaushik Veeraraghavan and Kelly Michelena and Keqian Li and Kiran Jagadeesh and Kun Huang and Kunal Chawla and Kyle Huang and Lailin Chen and Lakshya Garg and Lavender A and Leandro Silva and Lee Bell and Lei Zhang and Liangpeng Guo and Licheng Yu and Liron Moshkovich and Luca Wehrstedt and Madian Khabsa and Manav Avalani and Manish Bhatt and Martynas Mankus and Matan Hasson and Matthew Lennie and Matthias Reso and Maxim Groshev and Maxim Naumov and Maya Lathi and Meghan Keneally and Miao Liu and Michael L. Seltzer and Michal Valko and Michelle Restrepo and Mihir Patel and Mik Vyatskov and Mikayel Samvelyan and Mike Clark and Mike Macey and Mike Wang and Miquel Jubert Hermoso and Mo Metanat and Mohammad Rastegari and Munish Bansal and Nandhini Santhanam and Natascha Parks and Natasha White and Navyata Bawa and Nayan Singhal and Nick Egebo and Nicolas Usunier and Nikhil Mehta and Nikolay Pavlovich Laptev and Ning Dong and Norman Cheng and Oleg Chernoguz and Olivia Hart and Omkar Salpekar and Ozlem Kalinli and Parkin Kent and Parth Parekh and Paul Saab and Pavan Balaji and Pedro Rittner and Philip Bontrager and Pierre Roux and Piotr Dollar and Polina Zvyagina and Prashant Ratanchandani and Pritish Yuvraj and Qian Liang and Rachad Alao and Rachel Rodriguez and Rafi Ayub and Raghotham Murthy and Raghu Nayani and Rahul Mitra and Rangaprabhu Parthasarathy and Raymond Li and Rebekkah Hogan and Robin Battey and Rocky Wang and Russ Howes and Ruty Rinott and Sachin Mehta and Sachin Siby and Sai Jayesh Bondu and Samyak Datta and Sara Chugh and Sara Hunt and Sargun Dhillon and Sasha Sidorov and Satadru Pan and Saurabh Mahajan and Saurabh Verma and Seiji Yamamoto and Sharadh Ramaswamy and Shaun Lindsay and Shaun Lindsay and Sheng Feng and Shenghao Lin and Shengxin Cindy Zha and Shishir Patil and Shiva Shankar and Shuqiang Zhang and Shuqiang Zhang and Sinong Wang and Sneha Agarwal and Soji Sajuyigbe and Soumith Chintala and Stephanie Max and Stephen Chen and Steve Kehoe and Steve Satterfield and Sudarshan Govindaprasad and Sumit Gupta and Summer Deng and Sungmin Cho and Sunny Virk and Suraj Subramanian and Sy Choudhury and Sydney Goldman and Tal Remez and Tamar Glaser and Tamara Best and Thilo Koehler and Thomas Robinson and Tianhe Li and Tianjun Zhang and Tim Matthews and Timothy Chou and Tzook Shaked and Varun Vontimitta and Victoria Ajayi and Victoria Montanez and Vijai Mohan and Vinay Satish Kumar and Vishal Mangla and Vlad Ionescu and Vlad Poenaru and Vlad Tiberiu Mihailescu and Vladimir Ivanov and Wei Li and Wenchen Wang and Wenwen Jiang and Wes Bouaziz and Will Constable and Xiaocheng Tang and Xiaojian Wu and Xiaolan Wang and Xilun Wu and Xinbo Gao and Yaniv Kleinman and Yanjun Chen and Ye Hu and Ye Jia and Ye Qi and Yenda Li and Yilin Zhang and Ying Zhang and Yossi Adi and Youngjin Nam and Yu and Wang and Yu Zhao and Yuchen Hao and Yundi Qian and Yunlu Li and Yuzi He and Zach Rait and Zachary DeVito and Zef Rosnbrick and Zhaoduo Wen and Zhenyu Yang and Zhiwei Zhao and Zhiyu Ma},
      year={2024},
      eprint={2407.21783},
      archivePrefix={arXiv},
      primaryClass={cs.AI},
      institution = {Meta AI},
}

@techreport{radford2019language,
  title={Language models are unsupervised multitask learners},
  author={Radford, Alec and Wu, Jeffrey and Child, Rewon and Luan, David and Amodei, Dario and Sutskever, Ilya},
  journal={OpenAI blog},
  volume={1},
  number={8},
  year={2019},
  institution = {OpenAI},
  url= {https://cdn.openai.com/better-language-models/language_models_are_unsupervised_multitask_learners.pdf}
}

@INPROCEEDINGS{Olsen2025Probing,
  author={Olson, Matthew L. and Hinck, Musashi and Ratzlaff, Neale and Li, Changbai and Howard, Phillip and Lal, Vasudev and Tseng, Shao-Yen},
  booktitle={2025 IEEE/CVF International Conference on Computer Vision Workshops (ICCVW)}, 
  title={Probing the Representational Power of Sparse Autoencoders in Vision Models}, 
  year={2025},
  volume={},
  number={},
  pages={6226-6236},
  doi={10.1109/ICCVW69036.2025.00648}}

@misc{jiang2023mistral7b,
      title={Mistral 7B}, 
      author={Albert Q. Jiang and Alexandre Sablayrolles and Arthur Mensch and Chris Bamford and Devendra Singh Chaplot and Diego de las Casas and Florian Bressand and Gianna Lengyel and Guillaume Lample and Lucile Saulnier and Lélio Renard Lavaud and Marie-Anne Lachaux and Pierre Stock and Teven Le Scao and Thibaut Lavril and Thomas Wang and Timothée Lacroix and William El Sayed},
      year={2023},
      eprint={2310.06825},
      archivePrefix={arXiv},
      primaryClass={cs.CL},
      institution = {Mistral AI}
}

@inproceedings{park-etal-2025-decoding,
    title = "Decoding Dense Embeddings: Sparse Autoencoders for Interpreting and Discretizing Dense Retrieval",
    author = "Park, Seongwan  and
      Kim, Taeklim  and
      Ko, Youngjoong",
    editor = "Christodoulopoulos, Christos  and
      Chakraborty, Tanmoy  and
      Rose, Carolyn  and
      Peng, Violet",
    booktitle = "Proceedings of the 2025 Conference on Empirical Methods in Natural Language Processing",
    month = nov,
    year = "2025",
    address = "Suzhou, China",
    publisher = "Association for Computational Linguistics",
    url = "https://aclanthology.org/2025.emnlp-main.1345/",
    doi = "10.18653/v1/2025.emnlp-main.1345",
    pages = "26468--26485",
    ISBN = "979-8-89176-332-6"
}

@misc{SAELens,
   title = {{SAELens}},
   author = {Bloom, Joseph and Tigges, Curt and Duong, Anthony and Chanin, David},
   year = {2024},
   howpublished = {\url{https://github.com/decoderesearch/SAELens}},
}

@misc{TransformerLens,
    title = {{TransformerLens}},
    author = {Neel Nanda and Joseph Bloom},
    year = {2022},
    howpublished = {\url{https://github.com/TransformerLensOrg/TransformerLens}},
}

@incollection{rosch1973internal,
  author       = {Rosch, Eleanor H.},
  title        = {On the Internal Structure of Perceptual and Semantic Categories},
  booktitle    = {Cognitive Development and the Acquisition of Language},
  editor       = {Moore, Timothy E.},
  pages        = {111--144},
  publisher    = {Academic Press},
  address      = {New York},
  year         = {1973},
  doi          = {10.1016/B978-0-12-505850-6.50010-4},
  url          = {https://doi.org/10.1016/B978-0-12-505850-6.50010-4},
  howpublished = {\url{https://doi.org/10.1016/B978-0-12-505850-6.50010-4}}
}

@article{rosch1975cognitive,
  author       = {Rosch, Eleanor},
  title        = {Cognitive Representations of Semantic Categories},
  journal      = {Journal of Experimental Psychology: General},
  volume       = {104},
  number       = {3},
  pages        = {192--233},
  year         = {1975},
  doi          = {10.1037/0096-3445.104.3.192},
  url          = {https://doi.org/10.1037/0096-3445.104.3.192},
  howpublished = {\url{https://doi.org/10.1037/0096-3445.104.3.192}}
}

@article{mccloskey1978natural,
  author       = {McCloskey, Michael E. and Glucksberg, Sam},
  title        = {Natural Categories: Well Defined or Fuzzy Sets?},
  journal      = {Memory \& Cognition},
  volume       = {6},
  number       = {4},
  pages        = {462--472},
  year         = {1978},
  doi          = {10.3758/BF03197480},
  url          = {https://doi.org/10.3758/BF03197480},
  howpublished = {\url{https://doi.org/10.3758/BF03197480}}
}

@article{crosscoders_2024,
	title = {Sparse Crosscoders for Cross-Layer Features and Model Diffing - October 2024},
	author = {Jack Lindsey and Adly Templeton and Jonathan Marcus and Thomas Conerly and Joshua Batson and Christopher Olah},
	journal = {Transformer Circuits Thread},
	year = {2024},
    url = {https://transformer-circuits.pub/2024/crosscoders/},
}

@inproceedings{balagansky2025mechanistic,
title={Mechanistic Permutability: Match Features Across Layers},
author={Nikita Balagansky and Ian Maksimov and Daniil Gavrilov},
booktitle={The Thirteenth International Conference on Learning Representations},
year={2025},
url={https://openreview.net/forum?id=MDvecs7EvO}
}

@inproceedings{
mayne2024can,
title={Can sparse autoencoders be used to decompose and interpret steering vectors?},
author={Harry Mayne and Yushi Yang and Adam Mahdi},
booktitle={Interpretable AI: Past, Present and Future},
year={2024},
url={https://openreview.net/forum?id=6VGkENHc1J}
}

@article{elhage2021mathematical,
  title={A Mathematical Framework for Transformer Circuits},
  author={Elhage, Nelson and Nanda, Neel and Olsson, Catherine and Henighan, Tom and Joseph, Nicholas and Mann, Ben and Askell, Amanda and Bai, Yuntao and Chen, Anna and Conerly, Tom and DasSarma, Nova and Drain, Dawn and Ganguli, Deep and Hatfield-Dodds, Zac and Hernandez, Danny and Jones, Andy and Kernion, Jackson and Lovitt, Liane and Ndousse, Kamal and Amodei, Dario and Brown, Tom and Clark, Jack and Kaplan, Jared and McCandlish, Sam and Olah, Chris},
  journal={Transformer Circuits Thread},
  year={2021},
  note={\url{https://transformer-circuits.pub/2021/framework/index.html}}
}

@inproceedings{
karvonen2025saebench,
title={{SAEB}ench: A Comprehensive Benchmark for Sparse Autoencoders in Language Model Interpretability},
author={Adam Karvonen and Can Rager and Johnny Lin and Curt Tigges and Joseph Isaac Bloom and David Chanin and Yeu-Tong Lau and Eoin Farrell and Callum Stuart McDougall and Kola Ayonrinde and Demian Till and Matthew Wearden and Arthur Conmy and Samuel Marks and Neel Nanda},
booktitle={Forty-second International Conference on Machine Learning},
year={2025},
url={https://openreview.net/forum?id=qrU3yNfX0d}
}

@inproceedings{geva-etal-2022-transformer,
    title = "Transformer Feed-Forward Layers Build Predictions by Promoting Concepts in the Vocabulary Space",
    author = "Geva, Mor  and
      Caciularu, Avi  and
      Wang, Kevin  and
      Goldberg, Yoav",
    editor = "Goldberg, Yoav  and
      Kozareva, Zornitsa  and
      Zhang, Yue",
    booktitle = "Proceedings of the 2022 Conference on Empirical Methods in Natural Language Processing",
    month = dec,
    year = "2022",
    address = "Abu Dhabi, United Arab Emirates",
    publisher = "Association for Computational Linguistics",
    url = "https://aclanthology.org/2022.emnlp-main.3/",
    doi = "10.18653/v1/2022.emnlp-main.3",
    pages = "30--45"
}
\newpage
\appendix

\newcommand{\sweepfig}[3]{
\begin{figure*}[p]
    \centering
    \includegraphics[width=\textwidth]{figures/#1}
    \caption{#2}
    \label{#3}
\end{figure*}
}

\FloatBarrier
\section{Models and Sparse Autoencoders Used}
\label{app:saes_used}

We evaluate residual-stream SAEs across a set of pretrained language models spanning several model families and parameter scales. Table~\ref{tab:models_saes} summarizes the backbone models, SAE suites, dictionary widths, and SAE architectures used in this work. For Gemma~3, we use Gemma Scope~2 SAEs, which are JumpReLU SAEs trained with an additional Matryoshka loss. We report results for all the available sparsity regimes (\texttt{small}, \texttt{medium}, \texttt{big}) of Gemma~Scope~2 variants. The public model cards specify target average-$L_0$ ranges of approximately 10--20 for \texttt{small}, 30--60 for \texttt{medium}, and 60--150 for \texttt{big}. Note that \texttt{medium} sparsity models are only available for a limited amount of layers, while \texttt{small} and \texttt{big} sparsity SAEs are available for each layer. For GPT-2 Small and Llama~3.1~8B, we use TopK-style SAE suites, while the Mistral-7B SAEs use a ReLU nonlinearity.

\begin{table}[h]
    \centering
    \scriptsize
    \setlength{\tabcolsep}{3pt}
    \renewcommand{\arraystretch}{1.12}
    \caption{Language models and residual-stream SAEs used in this work. Widths are given in the common shorthand. For Gemma Scope~2, S/M/L denote the small, medium, and large target average-$L_0$ variants.}
    \label{tab:models_saes}
    \begin{tabular}{@{}%
        >{\raggedright\arraybackslash}p{0.27\linewidth}
        >{\raggedright\arraybackslash}p{0.20\linewidth}
        >{\raggedright\arraybackslash}p{0.13\linewidth}
        >{\raggedright\arraybackslash}p{0.30\linewidth}@{}}
        \hline
        \textbf{Backbone model} & \textbf{SAE suite} & \textbf{Width} & \textbf{SAE type} \\
        \hline
        Gemma~3 270M, 1B, 4B, 12B, 27B \citep{gemmateam2025gemma3technicalreport}
        & Gemma Scope~2 \citep{google2024gemmascope2}
        & 16k, 262k
        & \makecell[tl]{Matryoshka BatchTopK;\\ JumpReLU;\\ target-$L_0$: S/M/L} \\
        GPT-2 Small \citep{radford2019language}
        & OpenAI v5 \citep{gao2024scalingevaluatingsparseautoencoders}
        & 32k, 128k
        & TopK \\
        Llama~3.1~8B \citep{grattafiori2024llama3herdmodels}
        & Llama Scope \citep{he2024llamascopeextractingmillions}
        & 32k, 128k
        & JumpReLU \\
        Mistral-7B \citep{jiang2023mistral7b}
        & Engels et al. \citep{engels2025not}
        & 65k
        & ReLU \\
        \hline
    \end{tabular}
\end{table}
\FloatBarrier
\paragraph{Matryoshka SAEs}
A substantial part of our experimental results, including all figures shown in the main text, use Gemma Scope 2 SAEs. These are \emph{Matryoshka} SAEs \citep{learning_bussmann_2025, NEURIPS2022_c32319f4}, trained to capture latents at multiple conceptual resolutions within a single run. SAEs trained with a Matryoshka loss, are trained across several nested widths \(m_1<\cdots<m_K\). For each \(k\), let \(F^{(k)}\) denote the first \(m_k\) activations and \(W_d^{(k)}\) the corresponding decoder columns, yielding \(\hat{x}^{(k)} = W_d^{(k)}F^{(k)}(x)+b_d\). Training then minimizes a weighted sum of prefix-wise SAE losses,
\begin{equation}
\mathcal{L}_{\mathrm{Mat}}(x)
=\sum_{k=1}^K \alpha_k\Big(\|x-\hat{x}^{(k)}\|_2^2+\lambda_k\|F^{(k)}(x)\|_1\Big).
\label{eq:matryoshka_loss}
\end{equation}
Smaller widths tend to capture coarse, broadly reused latents, whereas larger widths allocate capacity to more specific concepts. The parameter \(\lambda_k\) controls the sparsity--fidelity trade-off at each width.\\
They are also known to mitigate feature absorption \citep{learning_bussmann_2025}, as the nested-width training objective encourages coarse, reusable latents to be represented already for smaller prefixes.

\section{Preliminary Studies}
\label{app:preliminary_studies}

\FloatBarrier
\subsection{Union of Properties on LRH Aligned Toy-Model}
\label{app:toy_union_setup}

\paragraph{Motivation and setup.} We evaluate SAE compositionality in a controlled LRH-aligned toy setting, following \citet{feature_chanin_2025}. The toy model acts as a frozen residual stream simulator in which activations are sparse linear superpositions of known ground-truth feature directions. This gives direct access to the true features and allows us to test whether SAE latent activations compose predictably when multiple features are combined.
We define $N=\{10, 50, 100\}$ ground-truth features in a $D=50$ dimensional activation space. The feature directions are the columns of a frozen linear map $W_{\mathrm{toy}} \in \mathbb{R}^{D \times N}$, optimized to be approximately pairwise orthogonal and of unit-norm. For a feature coefficient vector $a \in \mathbb{R}^{N}$, the toy activation is
\begin{equation}
h = W_{\mathrm{toy}} a .
\end{equation}
Each feature fires independently with probability $p=0.2$ and unit magnitude when active. We repeated the experiment at both higher and lower sparsity levels and observed qualitatively similar results. Thus, multi-feature activations arise naturally through chance co-occurrence, without imposing an explicit compositional rule.
We train standard ReLU SAEs on these activations using the architecture defined Equation~\eqref{eq:encode}~and~\eqref{eq:decode}. These are trained with reconstruction loss and a norm-scaled L1 sparsity penalty.

\subsubsection{Union Consistency Evaluation}
\label{sec:union_consistency_eval}

After training, we test whether the SAE latent code satisfies a union property. If two ground-truth features $i$ and $j$ are represented independently, then the signatures for their combined input should approximately match the union of the signatures for the individual inputs. Since the underlying feature directions are approximately orthogonal and the input is their linear superposition, a successful SAE should resolve the two contributions without suppressing or recruiting additional unrelated latents.\\
To probe this directly, we construct controlled activations from the toy model, simulating LRH-aligned residual stream states in which either one or two ground-truth features are active. Let $e_i$ denote the one-hot coefficient vector for feature $i$. Applying the frozen toy-model map to $e_i$ isolates the activation induced by feature $i$ alone:
\begin{align}
    h_i &= W_{\mathrm{toy}} e_i, \\
    h_j &= W_{\mathrm{toy}} e_j, \\
    h_{ij} &= W_{\mathrm{toy}}(e_i + e_j) = h_i + h_j .
\end{align}
We then pass each controlled activation through the SAE encoder. Let $F_{\ell}(h)$ denote the activation of SAE latent $\ell$ for input $h$, as defined by the encoder in Eq.~\ref{eq:encode}. Because the encoder uses a ReLU nonlinearity, we define a latent as active whenever its activation is positive:
\begin{equation}
S_x = \{\ell : F_{\ell}(h_x) > 0\}, \qquad x \in \{i,j,ij\}.
\end{equation}
The expected signature under perfect union consistency is $U_{ij}=S_i \cup S_j$. We compare the observed set $S_{ij}$ to $U_{ij}$ using
\begin{align}
    \mathrm{Jaccard}_{ij}
    &= \frac{|S_{ij} \cap U_{ij}|}{|S_{ij} \cup U_{ij}|}, \label{eq:jaccard} \\
    \mathrm{Recall}_{ij}
    &= \frac{|S_{ij} \cap U_{ij}|}{|U_{ij}|}, \label{eq:recall} \\
    \mathrm{Precision}_{ij}
    &= \frac{|S_{ij} \cap U_{ij}|}{|S_{ij}|}. \label{eq:precission}
\end{align}
Recall measures whether expected latents are preserved under composition, while precision measures whether the combined input recruits only expected latents. Thus, low recall indicates lost latents, whereas low precision indicates extra latents. Jaccard similarity summarizes both failure modes in a single set-overlap score. For each trained SAE, we average the metrics over all feature pairs. We then report mean and standard deviation across three independently seeded SAE training runs.

\FloatBarrier
\subsubsection{Results}
\label{sec:toy_union_results}

\begin{figure}[t]
    \centering
    \includegraphics[width=\linewidth]{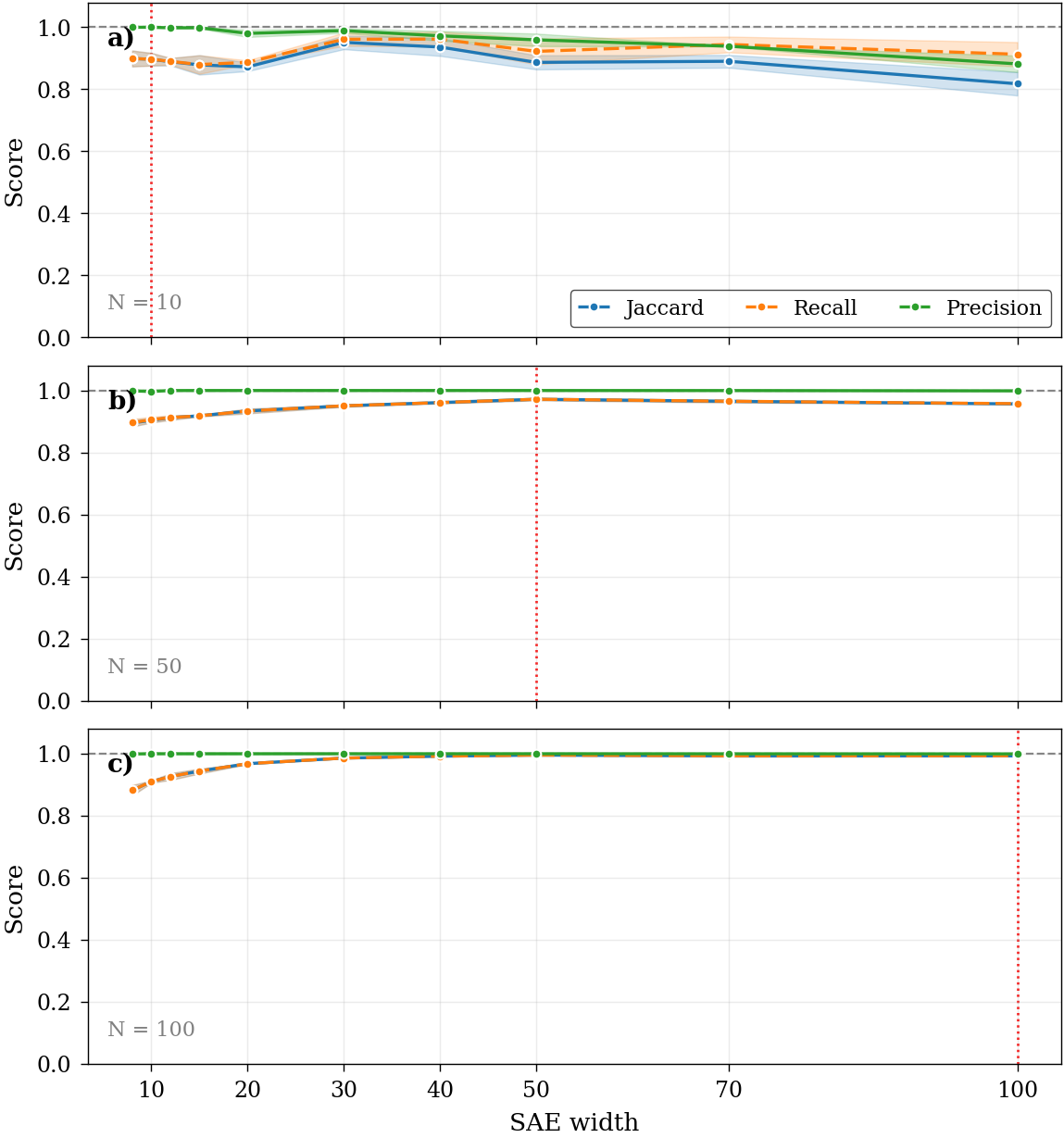}
    \caption{Jaccard, Recall, and Precision scores, as defined in Equations~\eqref{eq:jaccard}~to~\eqref{eq:precission}, for LRH-aligned feature-pair combinations, shown for increasing SAE width. Scores are averaged over all ground-truth feature pairs, with semitransparent regions denoting one standard deviation across random seeds. Panels show results for different numbers of ground-truth features: (a) $N=10$, (b) $N=50$, and (c) $N=100$. The red vertical line marks the point where the SAE width equals the number of ground-truth features.}
    \label{fig:toy_model}
\end{figure}
Results are shown in Figure~\ref{fig:toy_model} and Figure~\ref{fig:prelim_study} in the main text. Overall, all three metrics are high, typically ranging between $0.8$ and $1.0$. This indicates that, in the controlled LRH-aligned toy setting, the trained SAEs largely preserve the expected union structure of signatures.\\
The strongest, although still moderate, deviation from union consistency appears for $N=10$ ground-truth features, shown in Figure~\ref{fig:toy_model}~(a). For narrow SAEs, the degradation is mainly recall-driven. Precision remains close to one, while Recall and Jaccard decrease. Thus, the composed input typically does not recruit many unexpected latents, but instead some latents active for the individual features are lost under composition. This suggests that the dominant failure mode is not extra-latent recruitment, but a mild suppression of expected latents.
A plausible interpretation is that the mild degradation reflects a hedging-like distortion. Since Precision remains close to one, the combined input rarely activates latents outside the expected union. Instead, the lower Recall suggests that some latents active for isolated features are suppressed when features are combined. A plausible explanation might be feature hedging \citep{feature_chanin_2025}, as mixed feature components in capacity-limited latents may cause some expected latents to fall below the ReLU threshold when features are combined.\\
For larger ground truth feature sets $N=50$ and $N=100$ shown in Figure~\ref{fig:toy_model}~(b) and (c) respectively, we observe overall higher Jaccard similarity and recall, especially for wider SAEs. For small SAE widths we observe similar effects as for $N=10$.\\
Under realistic LRH-style assumptions, the number of underlying features is expected to exceed both the residual stream dimension and the number of available SAE latents \citep{10.5555/3692070.3693675, bricken2023towards}, making the $N=100$ setting the most "realistic" configuration. In this regime, we still observe only mild deviations from set-consistent behavior, indicating that within this controlled linear setting, SAE latent activations mostly satisfy the expected union property.\color{black}

\subsubsection{Scaling the Toy-Model Setup}
\label{sec:toy_union_scaling}

\begin{figure}[!tbp]
    \centering
    \includegraphics[width=\linewidth]{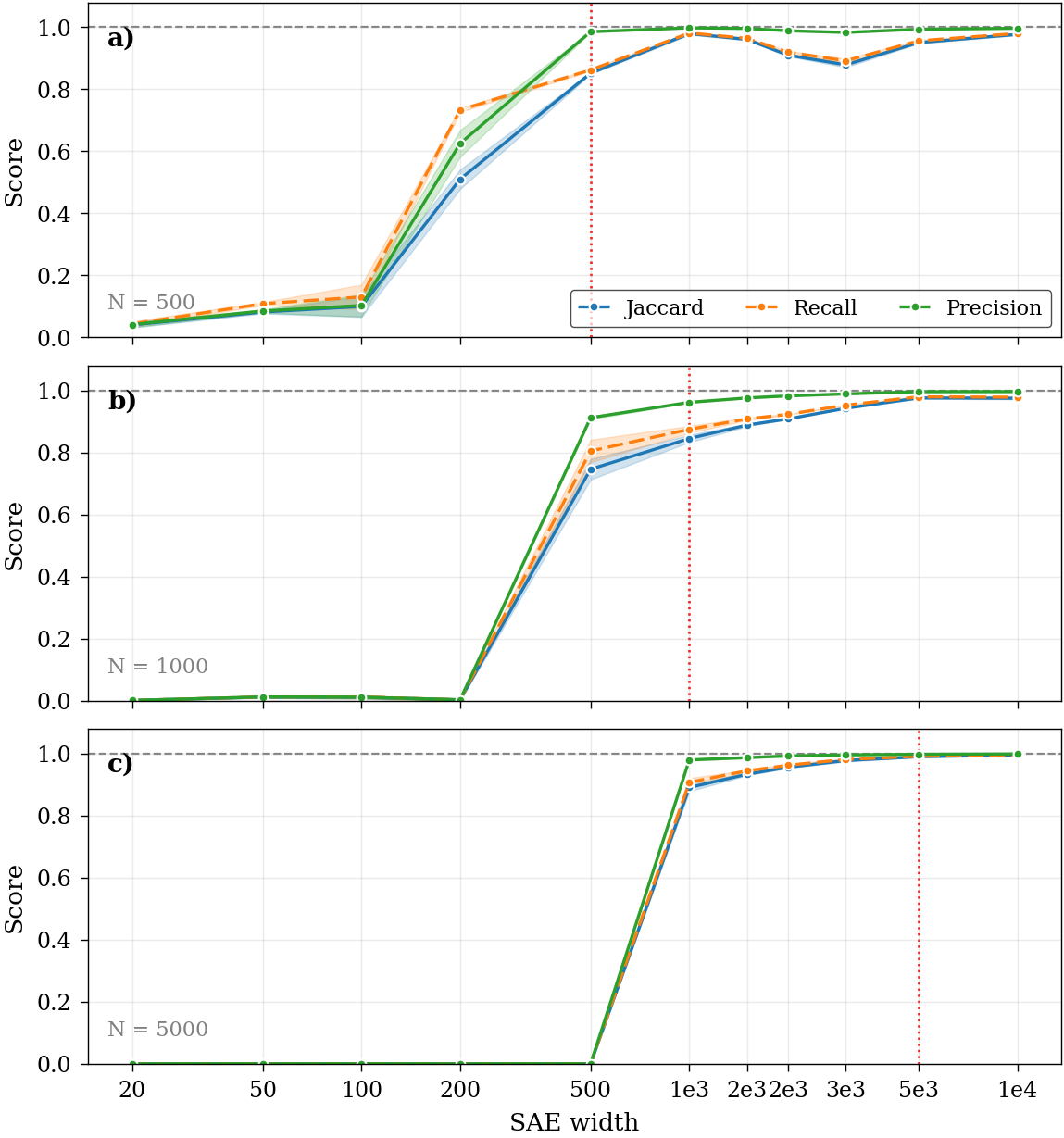}
    \caption{Jaccard, Recall, and Precision scores, as defined in Equations~\eqref{eq:jaccard}--\eqref{eq:precission}, for LRH-aligned feature-pair combinations as a function of SAE width in the larger \(D=1000\) scaling setup. Scores are averaged over randomly sampled ground-truth feature pairs, with semitransparent regions denoting one standard deviation across random seeds. Panels show results for increasing numbers of ground-truth features: (a) \(N=500\), (b) \(N=1\mathrm{e}^3\), and (c) \(N=5\mathrm{e}^3\). The red vertical line marks the point where SAE width equals the number of ground-truth features. The \(x\)-axis is logarithmic.}
    \label{fig:toy_scaling}
\end{figure}

To test whether the union-consistency results persist beyond the small \(D=50\) toy setting, we repeat the same evaluation in a larger activation space with \(D=1000\). We vary the number of ground-truth features over \(N \in \{500,1000,5000\}\) and sweep SAE widths $L\in [50,1\mathrm{e}^4]$, as delinearated in Appendix~\ref{sec:toy_implementation_details}.
This sweep covers both severely capacity-limited regimes, where \(L \ll N\), and strongly overcomplete regimes, where \(L > N\). As before, features fire independently with probability \(p=0.2\). Each SAE is trained on 50M toy activations with three random seeds. For large \(N\), evaluating all feature pairs becomes infeasible, so we estimate the union-consistency metrics using up to 5000 randomly sampled feature pairs per SAE.\\
Results are shown in Figure~\ref{fig:toy_scaling}. \\
In the strongly undercomplete regime, union consistency is poor, which can be considered to be expected. When the SAE has far fewer latents than ground-truth features, it cannot allocate independent latents to all directions, and feature-pair composition is not reliably resolved. As width increases, however, all three metrics improve sharply. The transition generally occurs around the point where SAE capacity becomes comparable to the ground-truth feature dictionary. However, for larger \(N=5\mathrm{e}^3\), we also observe regimes with \(L<N\) in which the SAE already recovers the union property reliably. This is important because LRH-motivated accounts typically assume that, although SAE dictionaries are overcomplete relative to the activation dimension, the number of underlying ground-truth features may be even larger \citep{10.5555/3692070.3693675, bricken2023towards}. This suggests that set-consistent behaviour does not require a strict one-latent-per-feature allocation, as there may plausibly be a regime of sufficient SAE-width in which SAEs recover compositional structure despite having fewer latents than ground-truth feature directions. Once the SAE is sufficiently wide, Jaccard similarity, recall, and precision remain high, indicating that the union property can persist in substantially larger settings. These results support the interpretation that the small-scale toy-model findings are not merely artifacts of the low-dimensional setup, but may reflect a more general behaviour of SAEs in controlled LRH-aligned representations.

\subsubsection{Latent Ground-Truth Features}

To make the simulation of residual-stream states closer to realistic model geometry, we also sample ground-truth feature directions from the dictionary of a trained Gemma-Scope SAE, whose decoder directions are themselves trained to reconstruct residual-stream states. Specifically, we use the residual-stream SAE of Gemma-3-270M with width \(16\mathrm{k}\) at layer \(9\). $D$ is thereby set to the dimensionality of the model, which is $640$ in this case. From its decoder directions, we greedily select a near-orthogonal subset with pairwise cosine similarity below \(0.2\), and use the first \(N\) selected directions as the ground-truth feature dictionary.\footnote{The threshold of \(0.2\) is selected pragmatically to obtain a sufficiently large set of directions satisfying this criterion.}\\
As before, simulated hidden states are sparse linear combinations of these directions, with each feature firing independently with probability \(p=0.2\). We then train L1-regularized SAEs on these activations and evaluate the same set-consistency metrics. 
\begin{figure}[htb]
    \centering
    \includegraphics[width=\linewidth]{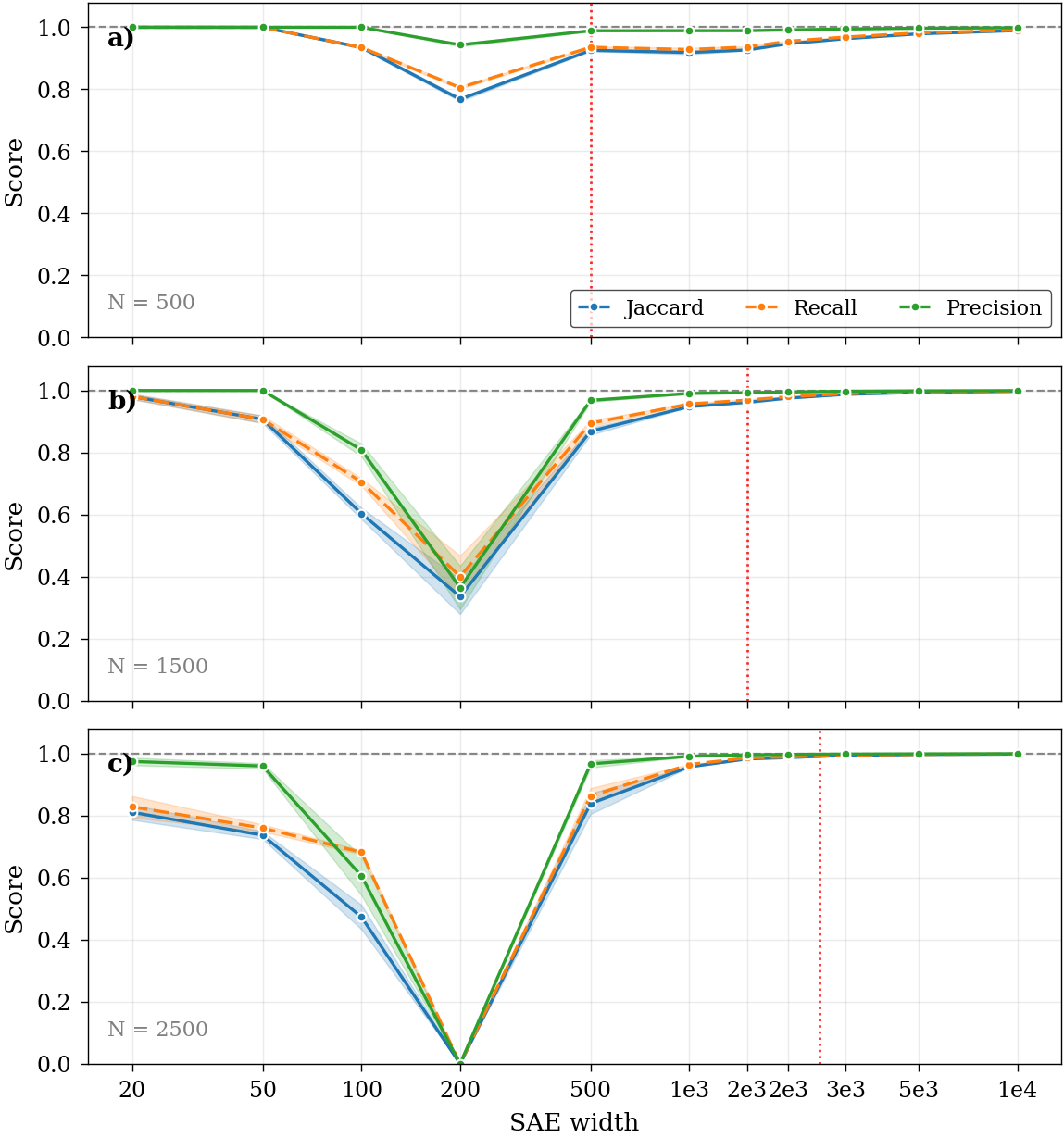}
    \caption{Jaccard, Recall, and Precision for the union-consistency evaluation using ground-truth directions sampled from a trained Gemma-Scope SAE decoder dictionary. Scores are shown across SAE widths for increasing dictionary sizes \(N\), with shaded regions denoting one standard deviation across seeds. The red vertical line marks width equal to \(N\) and the \(x\)-axis is on logarithmic scale.}
    \label{fig:toy_gemma_sampling}
\end{figure}
Evaluation is performed on up to \(5\mathrm{e}^3\) randomly sampled feature pairs.\\
The results show that union consistency is typically recovered for sufficiently wide SAEs, which alligns essentially with the results from Appendix~\ref{sec:toy_union_scaling}. Interestingly, the metrics can also be high in the strongly compressed regime. This should be interpreted cautiously, as when the SAE is severely capacity-limited, many inputs may collapse onto coarse shared latent patterns, so apparent set consistency need not imply that individual feature directions are faithfully resolved. The strongest failure mode appears at intermediate widths, around SAE width \(200\), and becomes more pronounced as \(N\) increases. In this regime, the SAE might have enough capacity to begin differentiating feature directions, but not enough to represent the full dictionary consistently. As a result, composed inputs may suppress latents active for individual features while also recruiting additional latents, leading to reduced performance on the metrics, more comparable to results in Section~\ref{sec:toy_union_scaling} for capacity limited SAEs. At larger widths, the metrics recover, suggesting that set-consistent behaviour can still emerge under more realistic feature geometry once sufficient capacity is available.
\FloatBarrier

\subsubsection{Implementation Details}
\label{sec:toy_implementation_details}

Across the toy-model studies above, we evaluate three configurations; a small near-orthogonal setup, a larger scaling setup, and a setup where ground-truth features are sampled from SAE latents using directions from a pretrained Gemma-Scope SAE. Unless stated otherwise, features fire independently with probability \(p=0.2\), active features have coefficient \(1\), and SAEs use a reconstruction loss with norm-scaled \(L_1\) sparsity. Table~\ref{tab:toy_experimental_configurations} summarizes the implementation details.
\begin{table}[h]
\centering
\scriptsize
\setlength{\tabcolsep}{4pt}
\renewcommand{\arraystretch}{1.08}
\caption{Experimental configurations for the LRH-aligned toy-model studies. \(D\) denotes the activation dimension, \(N\) the number of ground-truth features, and \(L\) the SAE width.}
\label{tab:toy_experimental_configurations}
\begin{tabularx}{\columnwidth}{@{}>{\raggedright\arraybackslash}p{0.31\columnwidth}>{\raggedright\arraybackslash}X@{}}
\toprule
\textbf{Detail} & \textbf{Value} \\
\midrule

\multicolumn{2}{@{}l}{\textbf{Global setup}} \\
Feature firing probability & \(p=0.2\) \\
SAE objective & reconstruction loss plus norm-scaled \(L_1\) penalty \\
SAE seeds & \(\{0,1,2\}\) \\

\midrule
\multicolumn{2}{@{}l}{\textbf{Orthogonal toy model}} \\
Feature geometry & optimized near-orthogonal directions \\
Activation dimension & \(D=50\) \\
Ground-truth features & \(N \in \{10,50,70,100\}\) \\
SAE widths & \(L \in \{8,10,12,15,20,30,40,50,70,100\}\) \\
Training activations & 15M \\
Evaluation & all feature pairs \\

\midrule
\multicolumn{2}{@{}l}{\textbf{Scaling setup}} \\
Feature geometry & optimized near-orthogonal directions \\
Activation dimension & \(D=1000\) \\
Ground-truth features & \(N \in \{500,1000,5000\}\) \\
SAE widths & \(L \in \{20,50,200,500,1\mathrm{e}^3,2\mathrm{e}^3,3\mathrm{e}^3,5\mathrm{e}^3,1\mathrm{e}^4\}\) \\
Training activations & 50M \\
Evaluation & up to 5000 sampled feature pairs \\

\midrule
\multicolumn{2}{@{}l}{\textbf{Decoder-sampled setup}} \\
Feature geometry & near-orthogonal directions sampled from a Gemma-Scope decoder dictionary \\
Activation dimension & \(D=640\) \\
Ground-truth features & \(N \in \{500,1500,2500\}\) \\
SAE widths & \(L \in \{20,50,200,500,1\mathrm{e}^3,2\mathrm{e}^3,3\mathrm{e}^3,5\mathrm{e}^3,1\mathrm{e}^4\}\) \\
Training activations & 50M \\
Evaluation & up to 5000 sampled feature pairs \\

\bottomrule
\end{tabularx}
\end{table}

\subsection{Interactive Exploration of SAE Set-Level Structure}
\label{app:interactive_sae_clusters}

To provide intuition for whether joint SAE activations can serve as a meaningful unit of analysis, we complement the controlled evaluations above with an exploratory clustering experiment on natural text. The goal to ask whether texts that activate similar \emph{sets} of SAE latents also form interpretable groups, placing the signature in the center of analysis instead of individual latents.\\
We sample \(30{,}000\) text snippets from \texttt{The Pile}~\citep{gao2020pile800gbdatasetdiverse}, truncate each snippet to 100 characters, and prepend the fixed prefix \mbox{"\texttt{Fact: }"}. For each snippet, we run the prompted text through the model and encode the corresponding residual-stream states with a fixed SAE. SAE activations are binarized token-wise, yielding an active/inactive latent signature
\[
    a(x,t) = \mathbf{1}\{F(x,t) > 0\}
\]
for each input \(x\) and token position \(t\). Here $F$ is defined according to Equation~\eqref{eq:encode}.
A simple union over signatures for each token position in a sequence would typically produce very large sets up to the entire dictionary, making comparisons between longer snippets less nuanced and obscuring which tokens contribute to the similarity. We therefore use a token-level average pairwise maximality score (APM). For two sequences \(u\) and \(v\), APM compares each token in \(u\) to the best-matching token in \(v\), where token similarity is measured by the intersection size of their SAE signature:
\begin{equation}
\begin{split}
&\operatorname{APM}(u,v)
:=\\
&\frac{1}{T}
\sum_{t=1}^{T}
\max_{l \in \{1,\dots,L\}}
\left\{
\left\lVert a(u,t) \odot a(v,l) \right\rVert_0
\right\}.
\end{split}
\label{eq:apm_set_main}
\end{equation}

\begin{figure*}[t]
    \centering

    \begin{subfigure}[t]{0.9\textwidth}
        \centering
        \includegraphics[width=\linewidth]{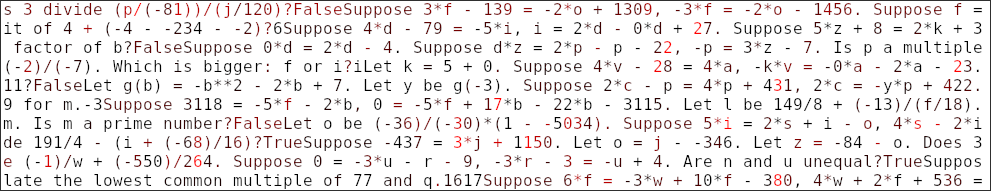}
        \caption{First ten entries of cluster 14381 obtained with the latent-set browser. The cluster appears to group contexts related to \emph{mathematical proofs}. Notably, the highest overlap appears to be individual digits. While all sentences share the word "Suppose" this does not meaningfully contribute to the APM.}
        \label{fig:cluster_example_1}
    \end{subfigure}

    \vspace{0.75em}

    \begin{subfigure}[t]{0.9\textwidth}
        \centering
        \includegraphics[width=\linewidth]{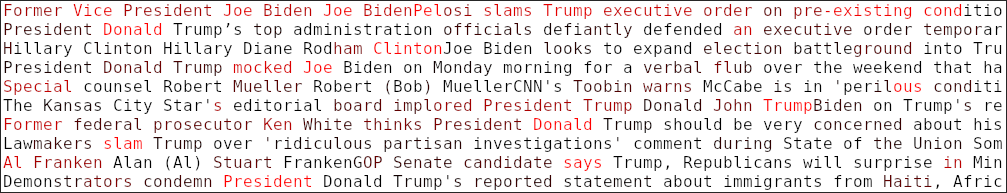}
        \caption{First then entries of cluster 388 obtained with the latent-set browser. The cluster appears to group contexts related to \emph{political actors talking about President Donald Trump}. Notably, the highest overlap tokens do not always correspond to "Donald Trump" and often highlight the sentence verb.}
        \label{fig:cluster_example_2}
    \end{subfigure}

    \vspace{0.75em}

    \begin{subfigure}[t]{0.9\textwidth}
        \centering
        \includegraphics[width=\linewidth]{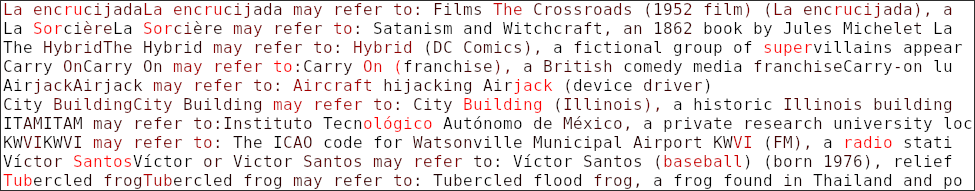}
        \caption{First ten entries of cluster 17405 obtained with the latent-set browser. The cluster appears to group contexts related to \emph{proper noun disambiguation}. Notably, while all samples share the phrase "may refer to" the largest overlap is achieved by the proper nouns themselves, despite being different in each sample.}
        \label{fig:cluster_example_4}
    \end{subfigure}

    \caption{Representative clusters obtained with the latent-set browser at \(\tau=25\). Each panel shows the first ten members of each cluster. Token-level heatmaps indicate which tokens have high latent signature overlap with the seed snippet.}
    \label{fig:cluster_examples}
\end{figure*}
Here, \(T\) and \(L\) are the sequence lengths of \(u\) and \(v\), \(\odot\) denotes elementwise multiplication, and \(\|\cdot\|_0\) counts the number of shared active latents. Since this score is typically asymmetric, we use the symmetrized version
\begin{equation}
\begin{split}
    &\operatorname{APM}^{\mathrm{sym}}(u,v)=\\
    &\frac{
    \operatorname{APM}(u,v)
    +
    \operatorname{APM}(v,u)
    }{2}.
    \label{eq:apm_set_sym}
\end{split}
\end{equation}
Using this metric, we operationalize the similarity between snippets as follows: On average, each token in one snippet should share many active SAE latents with some best-matching token in the other snippet. This also preserves token-level information and allows us to identify which parts of a snippet contribute most strongly to the similarity.\\
We use this setup for an interactive cluster browser. After precomputing APM scores for all sampled snippeds pairwise, one can select a seed snippet \(x_i\) around which a cluster is formed. All snippets \(x_j\) satisfying
\begin{equation}
    \operatorname{APM}^{\mathrm{sym}}(x_i,x_j) \geq \tau
\end{equation}
are assigned to the displayed cluster, where \(\tau\) is a user-controlled threshold. Higher thresholds produce smaller and typically more tightly related clusters, while lower thresholds reveal broader neighborhoods. For each cluster member, the browser displays a token-level heatmap indicating which tokens have high latent-set overlap with the seed. The user can then select any sampled snipped as a new seed and continue browsing the induced neighborhood structure.\\
We include representative examples for the Gemma-3-270M residual-stream SAE with width 16k and \texttt{big} sparsity target at layer 9. To support inspection beyond the selected examples, we provide an interactive browser as supplementary material, allowing readers to explore induced clusters under different seeds and thresholds. This experiment provides qualitative evidence that SAE signatures can expose interpretable structure in natural text. The code and browser will be released upon acceptance.\\
Figure~\ref{fig:cluster_examples} shows representative examples of typical clusters obtained with the latent-set browser. The examples were drawn at random and are representative of the overall quality of clusters. Each line shows a single cluster member and the lines are sorted from highest APM to lowest. We omit the shared "\texttt{Fact: }" prefix for readability.
\FloatBarrier
\section{Category-level clustering with AMI.}
\label{app:category_clustering_ami}
\begin{figure*}[t]
    \centering
    \includegraphics[width=0.95\textwidth]{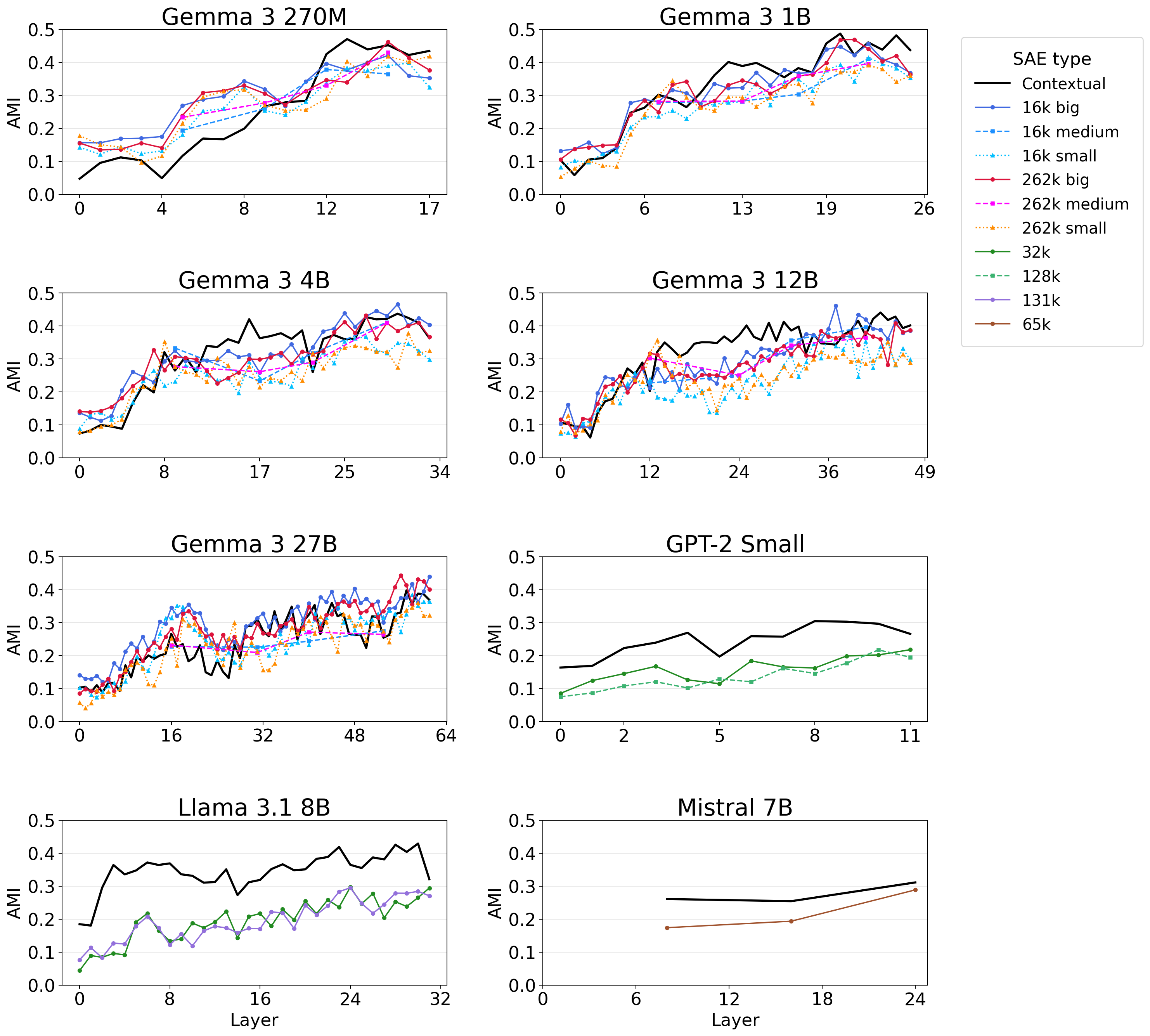}
    \caption{Layerwise category-boundary agreement for contextualized model representations, measured by AMI. For each layer, pairwise similarities between category instances are converted into distances \(d=1-s\), clustered with \(K\)-medoids using \(K\) equal to the ground-truth number of human categories, and compared against the ground-truth category labels. Results are shown for contextual residual-stream representations using cosine similarity \(s=s_{\mathrm{cos}}^\ell\) and for contextual SAE latent signatures using Jaccard similarity \(s=s_{\mathrm{Jac}}^\ell\). Higher AMI indicates that the corresponding layerwise representation induces clusters more closely aligned with human category boundaries.}
    \label{fig:ami_contextual_layerwise}
\end{figure*}

We ask whether model-induced geometries recover human category boundaries. The human-concepts dataset provides both category labels and typicality scores. Here, we use the labels to test whether items from the same human category also form coherent clusters under model-based similarities. Following the clustering analysis of \citet{shani2025tokensthoughtsllmshumans}, we cluster all items into \(K\) groups, where \(K\) is the number of ground-truth categories, and compare the resulting partition \(V\) against the human category partition \(U\).\\
We use Adjusted Mutual Information (AMI), which measures agreement between two partitions while correcting for chance agreement \citep{shani2025tokensthoughtsllmshumans}. Let \(U=\{U_i\}_{i=1}^R\) and \(V=\{V_j\}_{j=1}^C\) be two partitions of \(N\) items, with cluster sizes \(a_i=|U_i|\), \(b_j=|V_j|\), and contingency counts \(n_{ij}=|U_i\cap V_j|\). Their mutual information is
\begin{equation}
    I(U;V)
    =
    \sum_{i=1}^{R}\sum_{j=1}^{C}
    \frac{n_{ij}}{N}
    \log
    \frac{N n_{ij}}{a_i b_j}.
    \label{eq:ami_mi}
\end{equation}
For any partition \(C=\{C_j\}_{j=1}^{M}\) with cluster sizes \(c_j=|C_j|\), we write its entropy as
\begin{equation}
    H(C)
    =
    -\sum_{j=1}^{M}
    \frac{c_j}{N}
    \log \frac{c_j}{N}.
    \label{eq:ami_entropy}
\end{equation}
This definition applies to \(C=U\) and \(C=V\), respectively.
AMI subtracts the expected mutual information \(\mathbb{E}[I(U;V)]\) under a fixed-marginal random baseline and normalizes the result:
\begin{equation}
    \begin{split}
    &\operatorname{AMI}(U,V)
    =\\
    &\frac{I(U;V)-\mathbb{E}[I(U;V)]}
         {\max(H(U),H(V))-\mathbb{E}[I(U;V)]}.
    \label{eq:ami}
\end{split}
\end{equation}
Thus, \(\operatorname{AMI}=1\) indicates perfect agreement with human category labels, whereas \(\operatorname{AMI}\approx 0\) indicates chance-level agreement.\\
For each representation, we convert pairwise similarities into distances \(d=1-s\) and apply \(K\)-medoids clustering, which operates directly on precomputed distances and is therefore suitable for both cosine- and Jaccard-based similarities. We evaluate four representation families. First, static vocabulary embeddings use cosine similarity between average-pooled raw token embeddings from \(W_E\), corresponding to the uncontextualized version of \(s_{\mathrm{cos}}\) in Equation~\eqref{eq:typicality_cosine} with the layer index dropped. Second, contextual residual-stream states use \(s_{\mathrm{cos}}^\ell\) from Equation~\eqref{eq:typicality_cosine}, computed from the prompted template ``\texttt{This is a \{word\}.\_}''. Third, contextual SAE signatures use the binary target-word signatures from Equation~\eqref{eq:sae_signature} and their Jaccard similarity \(s_{\mathrm{Jac}}^\ell\) from Equation~\eqref{eq:typicality_jaccard}. Finally, uncontextualized SAE signatures are obtained by applying the lowest-layer SAE to raw token embeddings and computing the same Jaccard score. Static embedding and uncontextualized SAE results are summarized in Table~\ref{tab:ami_static_scalar}; layerwise contextual residual-stream and SAE results are shown in Figure~\ref{fig:ami_contextual_layerwise}. \\
\begin{table}[!t]
\centering
\scriptsize
\setlength{\tabcolsep}{3pt}
\renewcommand{\arraystretch}{1.02}
\caption{Static category-boundary agreement measured by AMI. Static embedding AMI is reported once per model. Static SAE-Jaccard AMI is reported separately for each SAE variant by applying the lowest-layer SAE to raw token embeddings.}
\label{tab:ami_static_scalar}
\begin{tabularx}{\columnwidth}{@{}%
>{\raggedright\arraybackslash}p{0.28\columnwidth}
>{\raggedright\arraybackslash}X
>{\raggedleft\arraybackslash}p{0.13\columnwidth}
>{\raggedleft\arraybackslash}p{0.16\columnwidth}@{}}
\toprule
\textbf{Model} & \textbf{SAE variant} & \textbf{Emb.} & \textbf{SAE-Jac.} \\
\midrule
\multirow{6}{*}{Gemma 3 270M}
& 16k big    & \multirow{6}{*}{0.1074} & 0.0294 \\
& 16k medium &                         & 0.0306 \\
& 16k small  &                         & 0.0429 \\
& 262k big   &                         & 0.0555 \\
& 262k medium&                         & 0.0457 \\
& 262k small &                         & 0.0305 \\
\midrule
\multirow{6}{*}{Gemma 3 1B}
& 16k big    & \multirow{6}{*}{0.1222} & 0.0183 \\
& 16k medium &                         & 0.0294 \\
& 16k small  &                         & 0.0229 \\
& 262k big   &                         & 0.0279 \\
& 262k medium&                         & 0.0982 \\
& 262k small &                         & 0.0085 \\
\midrule
\multirow{6}{*}{Gemma 3 4B}
& 16k big    & \multirow{6}{*}{0.1237} & 0.0244 \\
& 16k medium &                         & 0.0155 \\
& 16k small  &                         & 0.0196 \\
& 262k big   &                         & 0.0373 \\
& 262k medium&                         & 0.0734 \\
& 262k small &                         & 0.0141 \\
\midrule
\multirow{6}{*}{Gemma 3 12B}
& 16k big    & \multirow{6}{*}{0.1119} & 0.0136 \\
& 16k medium &                         & 0.0152 \\
& 16k small  &                         & 0.0174 \\
& 262k big   &                         & 0.0443 \\
& 262k medium&                         & 0.0034 \\
& 262k small &                         & 0.0248 \\
\midrule
\multirow{6}{*}{Gemma 3 27B}
& 16k big    & \multirow{6}{*}{0.1019} & 0.0057 \\
& 16k medium &                         & 0.0186 \\
& 16k small  &                         & 0.0089 \\
& 262k big   &                         & 0.0247 \\
& 262k medium&                         & 0.0529 \\
& 262k small &                         & 0.0208 \\
\midrule
\multirow{2}{*}{GPT-2 Small}
& 32k  & \multirow{2}{*}{0.0646} & 0.0121 \\
& 128k &                         & 0.0213 \\
\midrule
\multirow{2}{*}{Llama 3.1 8B}
& 32k  & \multirow{2}{*}{0.0601} & 0.0227 \\
& 131k &                         & 0.0174 \\
\midrule
Mistral 7B
& 65k & 0.1070 & 0.0103 \\
\bottomrule
\end{tabularx}
\end{table}

The static results in Table~\ref{tab:ami_static_scalar} show weak recovery of human category boundaries. Static embedding AMI is typically around \(0.1\), indicating that uncontextualized vocabulary embeddings contain minimal human-category-level structure, only weakly recover the human category partition in our setup. This is consistent with findings in \citet{shani2025tokensthoughtsllmshumans}. The uncontextualized SAE-Jaccard scores are lower still, suggesting that applying the lowest-layer SAE directly to raw token embeddings does not improve category-boundary recovery over the embedding geometry itself.\\
The contextualized layerwise results in Figure~\ref{fig:ami_contextual_layerwise} show a clearer category-boundary signal. For residual-stream cosine similarity \(s_{\mathrm{cos}}^\ell\), AMI ranges roughly from near zero to about \(0.5\), with an overall tendency to increase across layers. Contextual SAE signatures \(s_{\mathrm{Jac}}^\ell\) show a broadly similar increasing trend, although the detailed curves vary from the embedding related curve. Across SAE types, the layerwise AMI curves are mostly more similar to each other than to the residual-stream cosine curve, suggesting that SAE variants show comparable layerwise agreement with human category labels, but differ more strongly from the corresponding agreement pattern of dense residual-stream similarities. 
\begin{table}[!t]
\centering
\scriptsize
\setlength{\tabcolsep}{3pt}
\renewcommand{\arraystretch}{1.02}
\caption{Peak layerwise contextualized category-boundary agreement measured by AMI. Reported values correspond to the maximum AMI across layers for residual-stream cosine similarity and, separately for each SAE variant, the maximum AMI across layers for SAE-set Jaccard similarity.}
\label{tab:ami_peak}
\begin{tabularx}{\columnwidth}{@{}%
>{\raggedright\arraybackslash}p{0.28\columnwidth}
>{\raggedright\arraybackslash}X
>{\raggedleft\arraybackslash}p{0.13\columnwidth}
>{\raggedleft\arraybackslash}p{0.16\columnwidth}@{}}
\toprule
\textbf{Model} & \textbf{SAE variant} & \textbf{Emb.} & \textbf{SAE-Jac.} \\
\midrule
\multirow{6}{*}{Gemma 3 270M}
& 16k big     & \multirow{6}{*}{0.471} & 0.421 \\
& 16k medium  &                         & 0.379 \\
& 16k small   &                         & 0.398 \\
& 262k big    &                         & 0.463 \\
& 262k medium &                         & 0.431 \\
& 262k small  &                         & 0.419 \\
\midrule
\multirow{6}{*}{Gemma 3 1B}
& 16k big     & \multirow{6}{*}{0.488} & 0.456 \\
& 16k medium  &                         & 0.411 \\
& 16k small   &                         & 0.415 \\
& 262k big    &                         & 0.470 \\
& 262k medium &                         & 0.397 \\
& 262k small  &                         & 0.391 \\
\midrule
\multirow{6}{*}{Gemma 3 4B}
& 16k big     & \multirow{6}{*}{0.438} & 0.466 \\
& 16k medium  &                         & 0.412 \\
& 16k small   &                         & 0.367 \\
& 262k big    &                         & 0.432 \\
& 262k medium &                         & 0.410 \\
& 262k small  &                         & 0.378 \\
\midrule
\multirow{6}{*}{Gemma 3 12B}
& 16k big     & \multirow{6}{*}{0.441} & 0.461 \\
& 16k medium  &                         & 0.397 \\
& 16k small   &                         & 0.376 \\
& 262k big    &                         & 0.409 \\
& 262k medium &                         & 0.365 \\
& 262k small  &                         & 0.357 \\
\midrule
\multirow{6}{*}{Gemma 3 27B}
& 16k big     & \multirow{6}{*}{0.397} & 0.439 \\
& 16k medium  &                         & 0.275 \\
& 16k small   &                         & 0.385 \\
& 262k big    &                         & 0.444 \\
& 262k medium &                         & 0.271 \\
& 262k small  &                         & 0.359 \\
\midrule
\multirow{2}{*}{GPT-2 Small}
& 32k  & \multirow{2}{*}{0.304} & 0.218 \\
& 128k &                         & 0.217 \\
\midrule
\multirow{2}{*}{Llama 3.1 8B}
& 32k  & \multirow{2}{*}{0.429} & 0.297 \\
& 131k &                         & 0.295 \\
\midrule
Mistral 7B
& 65k & 0.311 & 0.289 \\
\bottomrule
\end{tabularx}
\end{table}
However, the peak AMI values, reported separately in Table~\ref{tab:ami_peak}, show that SAE-based similarities often remain below the corresponding dense residual-stream state similarities. The gap is smaller than in the static comparison in Table~\ref{tab:ami_static_scalar} and for the gemma SAEs there are cases where the trend is slightly reversed, but the overall pattern is consistent: SAE latent sets do not outperform dense model representations on the category-boundary task.

\section{Typicality Rankings}
\label{app:typicality_rankings}
Below, we discuss the metrics introduced in Section~\ref{sec:typicality_judgements} for all investigated models and SAE variants. The results are overall quite robust across models and SAE types, with exceptions discussed onward.
\begin{figure*}[t]
    \centering
    \includegraphics[width=\textwidth]{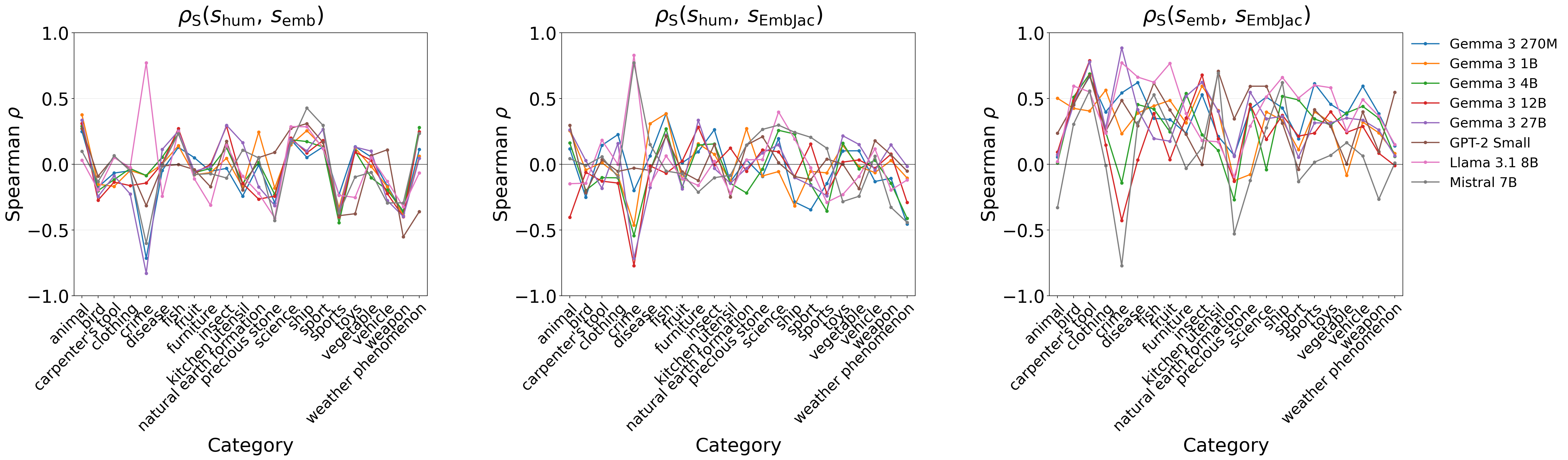}
    \caption{Spearman rank correlations between human and model-induced typicality rankings. Panel~(a) compares human typicality scores \(s_{\mathrm{hum}}\) with the embedding-based typicality score \(s_{\mathrm{emb}}\), computed from \(s_{\mathrm{cos}}^\ell\) in Equation~\eqref{eq:typicality_cosine}. Panel~(b) compares \(S_{\mathrm{hum}}\) with the uncontextualized SAE set-based Jaccard score \(s_{\mathrm{Jac}}^{0}\), computed from the layer-0 SAE representation. Panel~(c) compares \(S_{\mathrm{emb}}\) with the same SAE Jaccard score \(s_{\mathrm{Jac}}^{0}\). Results are shown across semantic categories for all investigated models.}
    \label{fig:typicality_vs_apt}
\end{figure*}
Figure~\ref{fig:typicality_vs_apt} shows, for each category, the Spearman correlation between human typicality ratings $s_{\text{hum}}$ provided by the human typicality dataset~\citep{shani2025tokensthoughtsllmshumans, rosch1973internal, rosch1975cognitive, mccloskey1978natural} and uncontextualized cosine similarity vocabulary embedding scores $s_{\text{emb}}$, computed from \(s_{\mathrm{cos}}^\ell\) in Equation~\eqref{eq:typicality_cosine} and uncontextualized jaccard based similarity scores on the sae of layer 0.  Panel a therefore measures the extent to which the model's embedding-based similarity structure agrees with the human ranking of how representative an instance is for its category. It shows that the correlations between human typicality judgments and the target-word cosine similarity scores \(s_{\mathrm{cos}}^\ell\) are generally weak and overall volatile around zero, depending on the category. "Crime" is standing out as an exception due to its strong anticorrelation for many models and its strong correlation for Llama~3.1~8B, however the overall impression remains. Therefore, these results indicate that the model's embedding geometry does not organize category membership in a way that closely matches human typicality judgments, consistent with prior observations in the literature, see \citep{shani2025tokensthoughtsllmshumans}. Interestingly the SAE activations show a comparably diffuse correlation with the human typicality rankings, again with stronger scores for the crime category, however a noisy trend overall, indicating that the structure on the sae latents does not seem to track human typicallity. Panel (c) shows that the scores on the embedding structure align much more closely with sae level analysis as indicated by the overall positive trend. Interestingly, rankings are not equal to one indicating that even though the sae aligns more closely here it decomposition of the latent directions still does something qulaitatively different.\\
\\
From Figure~\ref{fig:typicalitysweep_gemma_270m} to Figure~\ref{fig:typicalitysweep_llama31_8b}, each individual figure corresponds to one model, and each row within a figure corresponds to a different SAE type trained for that model. 
Each panel resolves within-category Spearman rank correlations at different levels of model depth.
Across all figures, the columns report complementary rank-correlation analyses. The first four columns show category-wise Spearman correlations, where each score induces a ranking over instances within a category. The first column shows \(\rho_S(s_{\mathrm{hum}}, s_{\mathrm{cos}}^\ell)\), i.e.\ the correlation between human typicality ratings and contextual residual-stream cosine similarity at each layer. The second column shows \(\rho_S(s_{\mathrm{hum}}, s_{\mathrm{Jac}}^\ell)\), measuring how well human category judgments align with SAE set-based Jaccard similarity. The third column shows \(\rho_S(s_{\mathrm{emb}}, s_{\mathrm{Jac}}^\ell)\), comparing static embedding cosine similarity with the SAE-induced Jaccard ranking. The fourth column shows \(\rho_S(s_{\mathrm{cos}}^\ell, s_{\mathrm{Jac}}^\ell)\), comparing contextual residual-stream cosine similarity with SAE set-based Jaccard similarity at the same layer. The final column shows the cross-layer consistency of the SAE-based scores, i.e.\ \(\rho_S(s_{\mathrm{Jac}}^{\ell_1}, s_{\mathrm{Jac}}^{\ell_2})\). This final panel is not category-resolved,  but aggregates over the dataset and reports the layer-by-layer agreement between SAE-induced typicality rankings. 
Since this matrix is symmetric, we display only the diagonal and lower-triangular part of the \(\ell_1 \times \ell_2\) matrix.
\FloatBarrier
\sweepfig{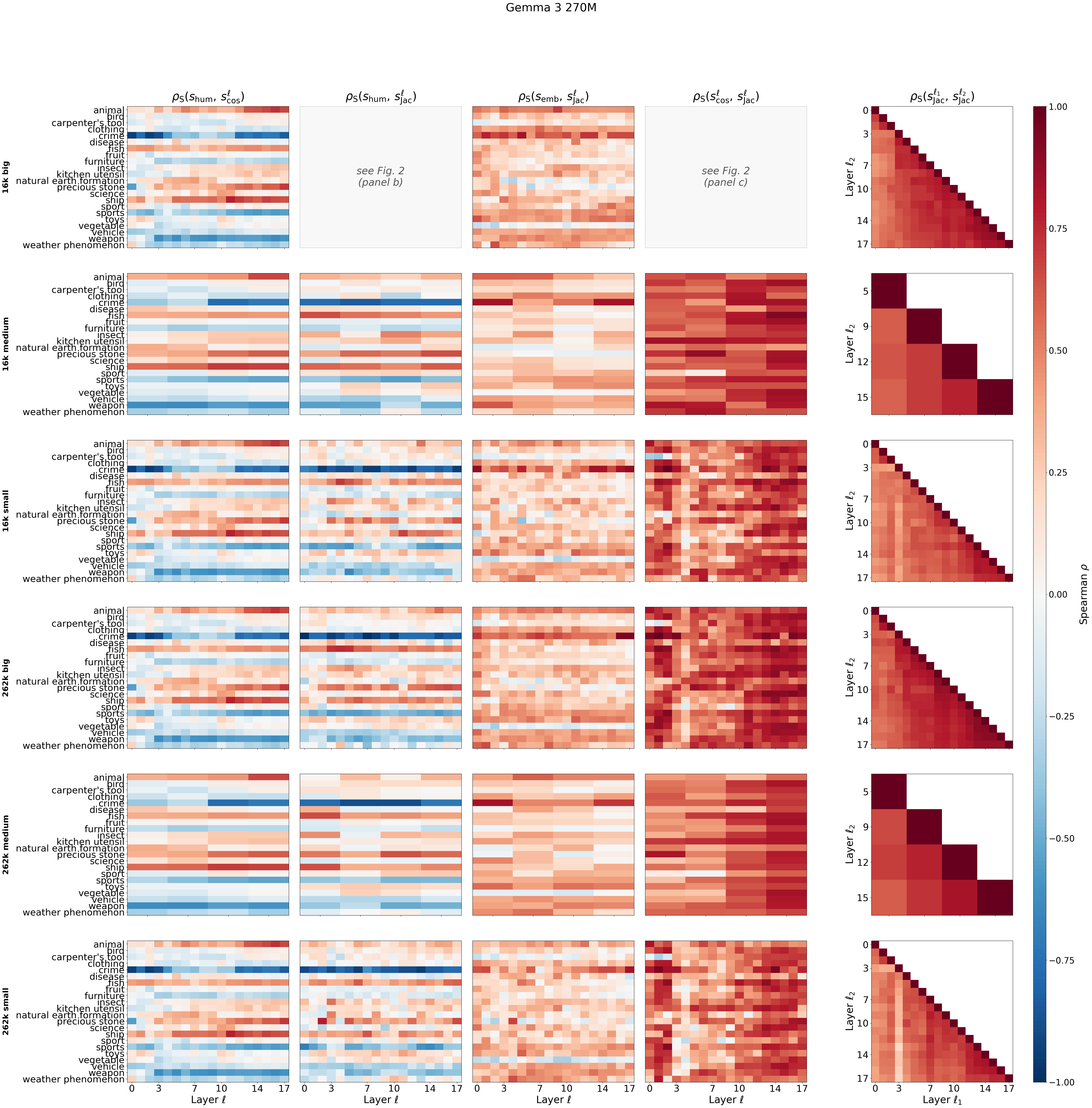}{Layer-dependent Spearman rank correlations for Gemma 3 270M.}{fig:typicalitysweep_gemma_270m}
\sweepfig{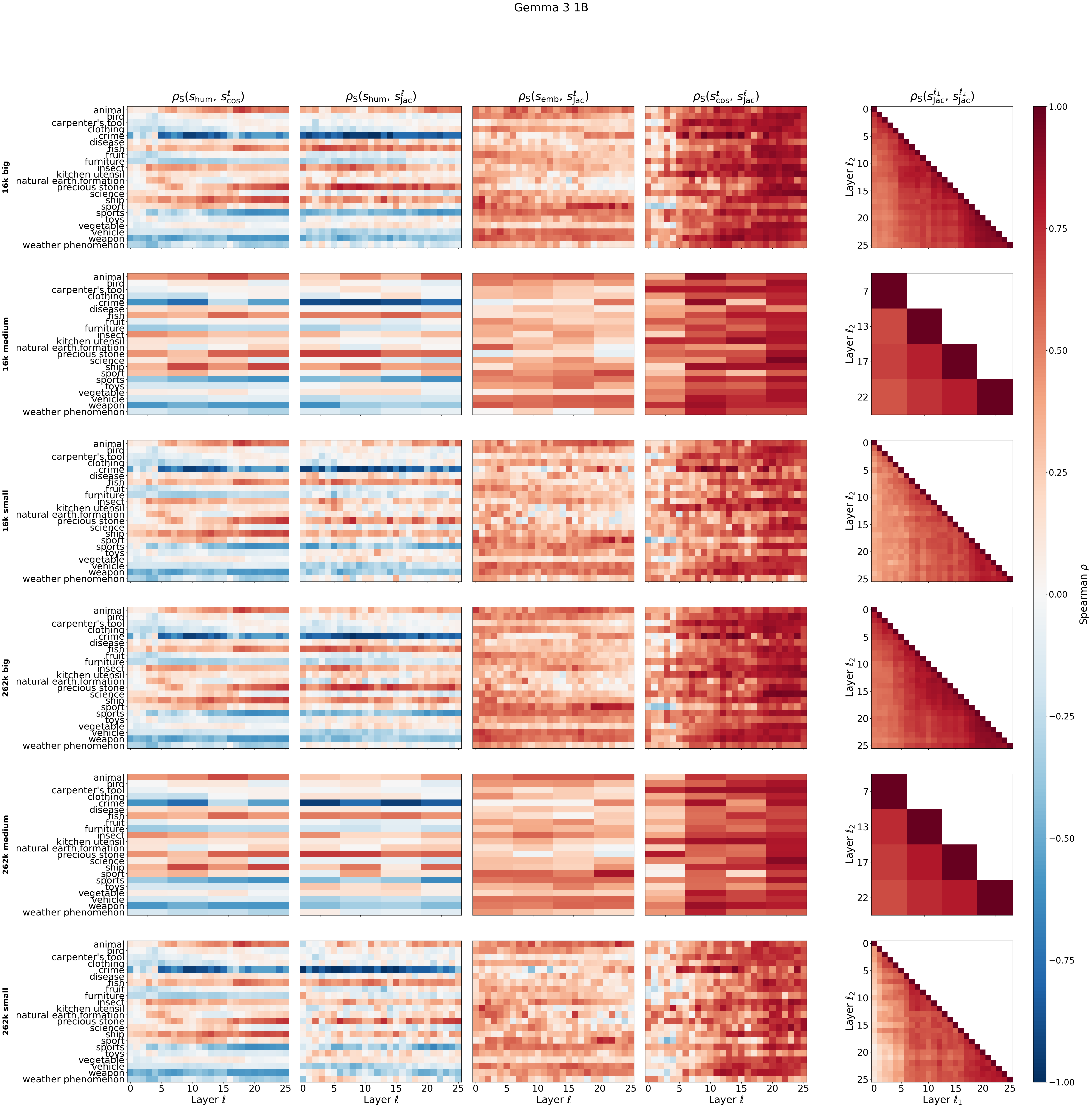}{Layer-dependent Spearman rank correlations for Gemma 3 1B.}{fig:typicalitysweep_gemma_1b}
\sweepfig{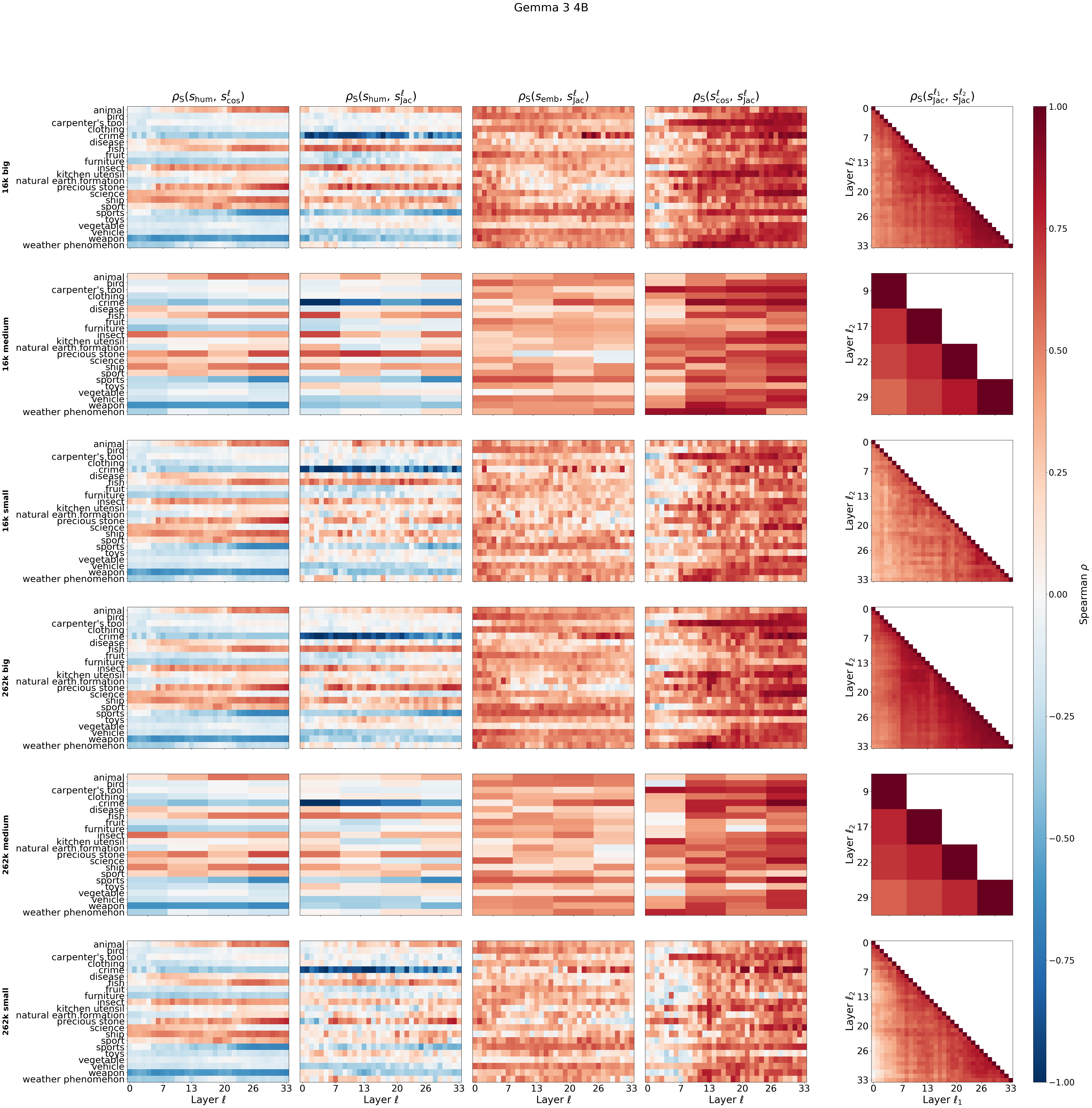}{Layer-dependent Spearman rank correlations for Gemma 3 4B.}{fig:typicalitysweep_gemma_4b}
\sweepfig{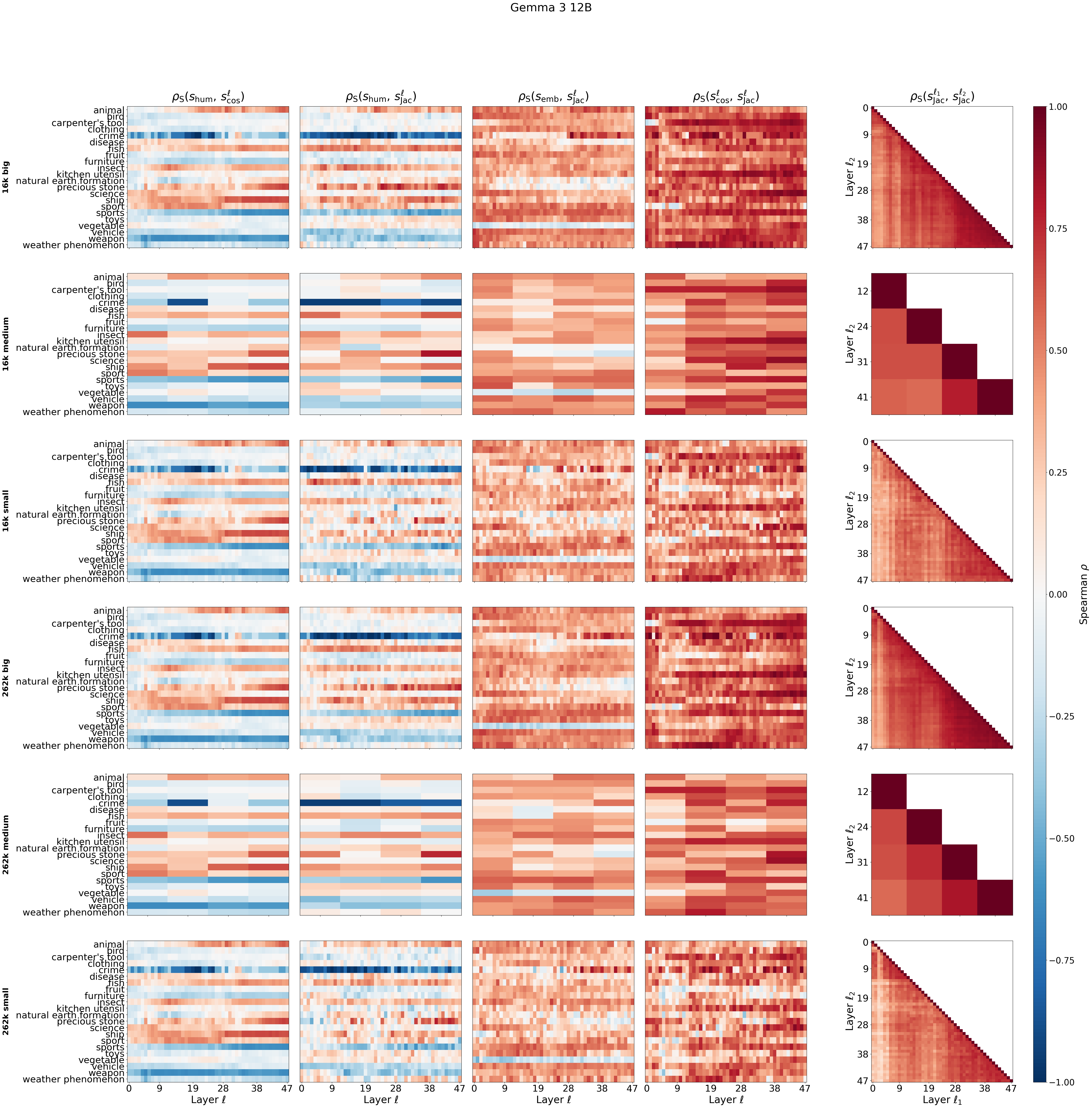}{Layer-dependent Spearman rank correlations for Gemma 3 12B.}{fig:typicalitysweep_gemma_12b}
\sweepfig{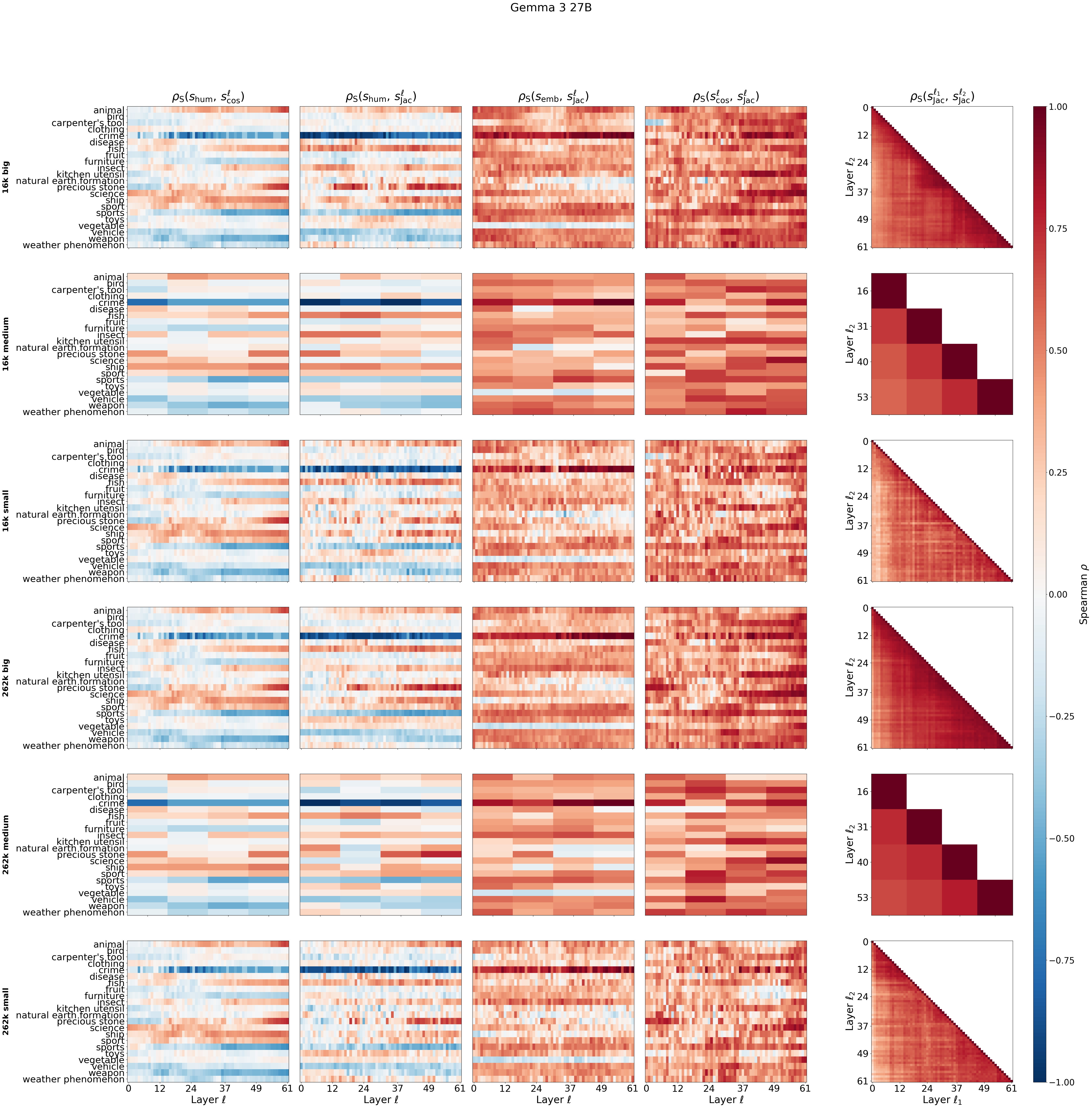}{Layer-dependent Spearman rank correlations for Gemma 3 27B.}{fig:typicalitysweep_gemma_27b}
\sweepfig{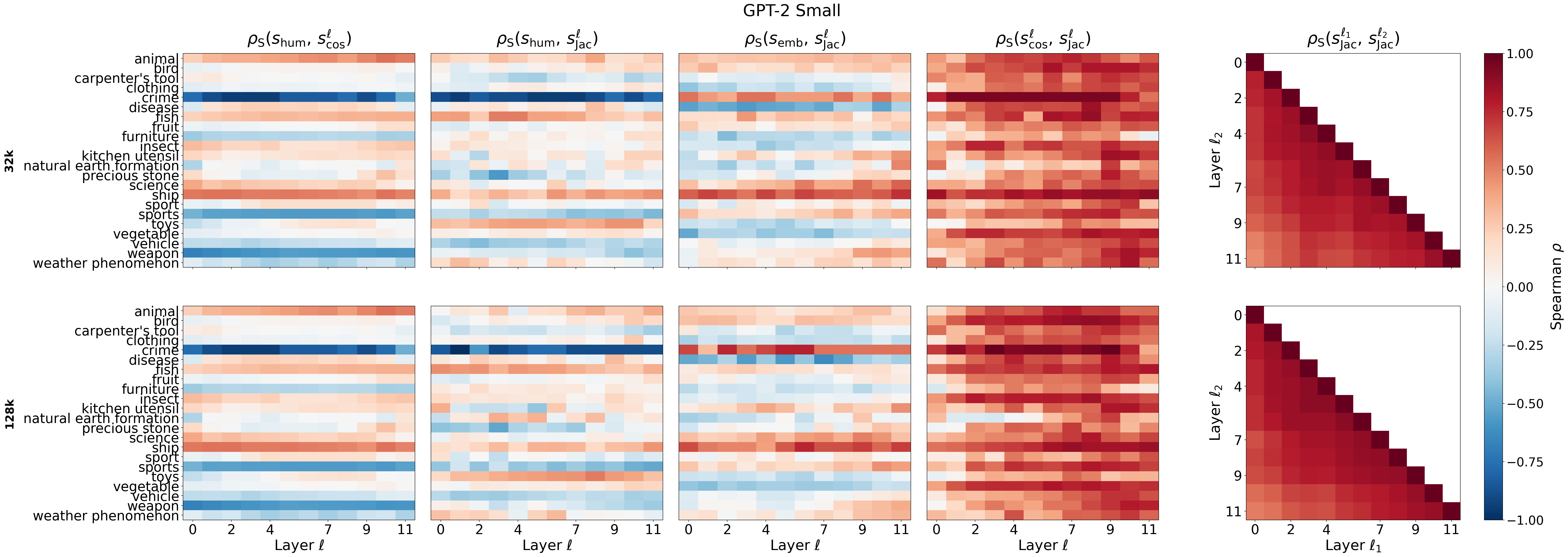}{Layer-dependent Spearman rank correlations for GPT-2 Small.}{fig:typicalitysweep_gpt2_small}
\sweepfig{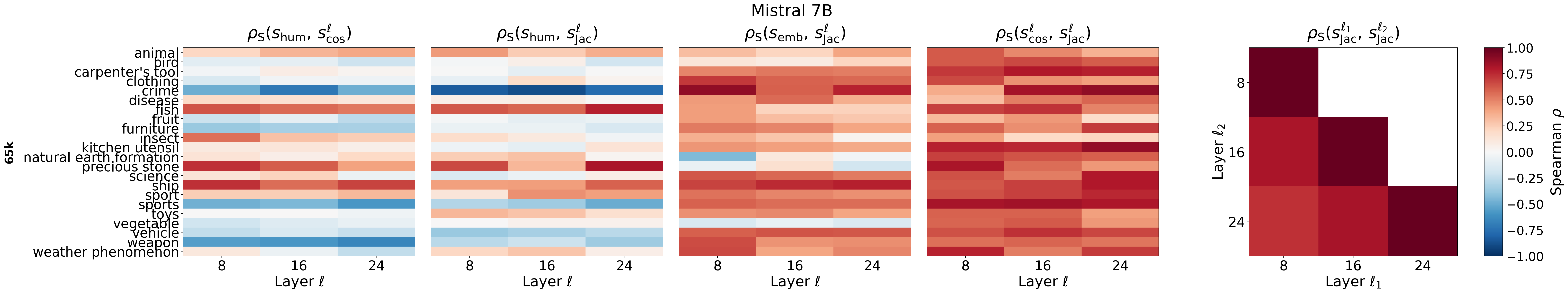}{Layer-dependent Spearman rank correlations for Mistral 7B.}{fig:typicalitysweep_mistral_7b}
\sweepfig{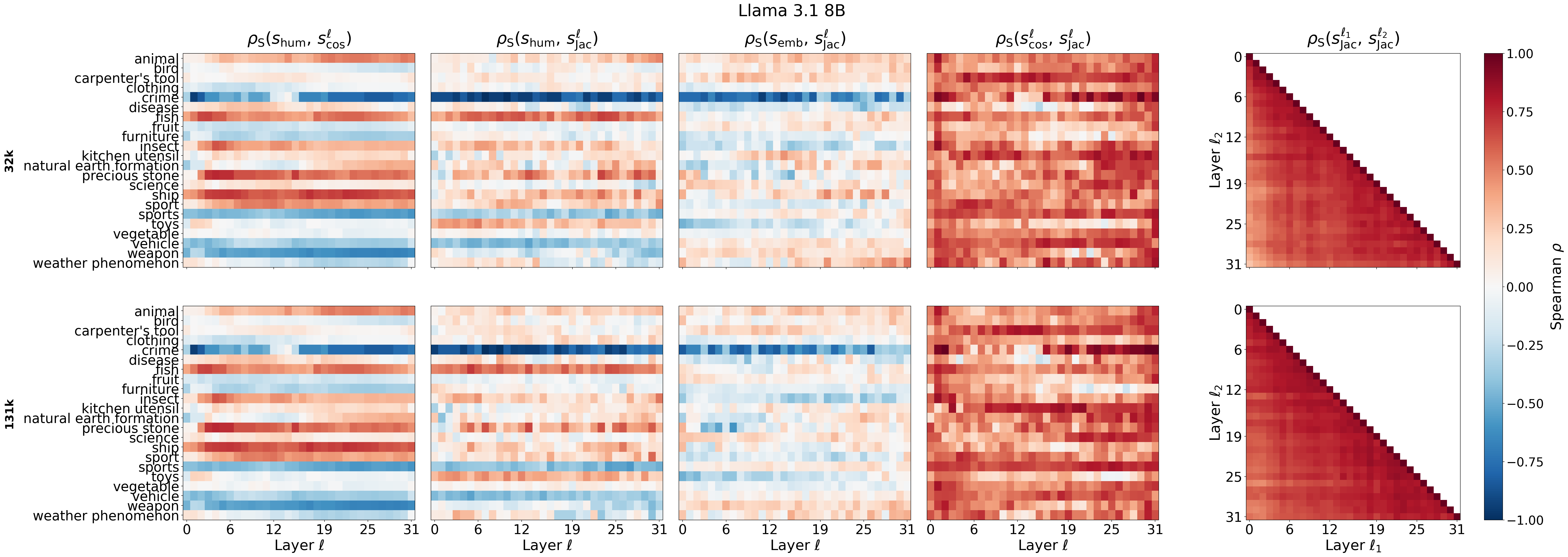}{Layer-dependent Spearman rank correlations for Llama 3.1 8B.}{fig:typicalitysweep_llama31_8b}
\color{white}.\color{black} \\
The comparison between human typicality scores and SAE activation patterns (i.e. the second collumn) is especially relevant. A central motivation in the SAE literature \citep{scaling_templeton_2024, bricken2023towards, cunningham2023sparseautoencodershighlyinterpretable} is that SAE latents recover interpretable concepts and may therefore provide a representation that aligns more naturally with human-understandable concepts. Under this expectation, one might expect that the correlation between human typicality and SAE set-based Jaccard similarity \(s_{\mathrm{Jac}}^\ell\) would be substantially stronger than the corresponding result on the hidden state model embeddings, shown in the first collumn in the plot. However as already discussed in Section~\ref{sec:typicality_judgements}, this is not observed. The correlations between human ratings and SAE set-based Jaccard similarity \(s_{\mathrm{Jac}}^\ell\) remain weak across models, layers, and SAE types, indicating that SAE activation patterns do not resolve conceptual similarity in a way that is substantially closer to human typicality than the raw model embeddings. By contrast, the third column of plots shows that cosine similarity scores on the raw model embeddings \(s_{\mathrm{emb}}\) and SAE set-based Jaccard scores \(s_{\mathrm{Jac}}^\ell\) are in most cases qualitatively positively correlated. Thus, although SAE activation patterns do not align well with human conceptualization, they do align closely with the model's own embedding-based conceptualization structure. 
GPT~2 and Llama~3.1~8B SAEs are an exception from this trend. Here, we do not observe a consistent trend toward positive rank correlation. Instead, the correlations fluctuate around zero across categories. 
A trend that holds reliably across SAEs is the strong positive correlation between contextual residual-stream similarity \(s_{\mathrm{cos}}^\ell\) and SAE set-based similarity \(s_{\mathrm{Jac}}^\ell\). This agreement is partly expected, since the SAE code is a dictionary-based decomposition of the residual-stream state at the same layer. However, the result should not be overemphasized, as SAE activations are meant to expose a sparse and more interpretable structure, not simply to reproduce the dense residual-stream geometry. The correlations are therefore informative as they are positive but also not close to one, suggesting that SAE set similarity tracks the model's residual-stream similarity structure while still measuring something non-identical.\\
The final column shows cross-layer agreement of SAE set-based Jaccard similarities \(s_{\mathrm{Jac}}^\ell\). These correlations are often strong, indicating that SAE-induced similarity rankings are relatively stable across depth. For Gemma Scope~2, the lower-sparsity (\texttt{big}) variants typically show higher cross-layer correlations than higher-sparsity variants, suggesting that denser SAE signatures yield smoother, more layer-stable rankings.\\
Overall, SAE activation sets do not faithfully recover human conceptual typicality. Instead, they more closely track the model's internal similarity structure, which is known to differ from human judgments \citep{shani2025tokensthoughtsllmshumans}.

\FloatBarrier
\section{Lost Latents}
\subsection{Choosing a threshold for non-JumpReLU SAEs}
\label{app:threshold_choice}

As discussed in Section~\ref{sec:what_happened_to_lost_feats}, for a latent that is inactive at a given residual state we distinguish between two cases; it is either \emph{not emitted}, meaning that its encoder direction is approximately orthogonal to the residual vector, or \emph{suppressed}, meaning that the corresponding pre-activation in Equation~\eqref{eq:preactivation} is sufficiently negative \citep{mayne2024can, elhage2021mathematical}. For JumpReLU SAEs, this distinction is natural, since the learned JumpReLU offset already provides a scale for the decision.
For the remaining SAEs, no learned threshold is available, so we introduce an angular tolerance \(\theta\) for deciding when an encoder direction should count as approximately orthogonal to the residual vector. To motivate this choice, we use a geometric heuristic based on spherical codes~\citep{ml_math_2023}. Let \(A(n,\phi)\) denote the maximum number of unit vectors in \(\mathbb{R}^n\) whose pairwise angle is at least \(\phi\). Known asymptotic bounds~\citep{6771362} imply that, for fixed \(\phi < \pi/2\),
\begin{equation}
    A(n,\phi) \approx \exp\ \!\bigl(n \log(1/\sin\phi) + o(n)\bigr),
\label{eq.geom_heuristic}
\end{equation}
up to lower-order terms.
Strictly speaking, this applies on the unit sphere, whereas SAE pre-activations depend on inner products with residual vectors of varying norm. However, for our purposes this mainly introduces a scale factor. The sign of the pre-activation is unchanged, and the qualitative transition between clearly non-orthogonal and approximately orthogonal directions is still governed by the angular term. Since we use the argument only as a heuristic for selecting a pragmatic cutoff, and focus on the location of the ``hockey-stick'' transition rather than on absolute values, this approximation is sufficient.
We therefore inspect
$\Lambda(\phi) := n \log(1/\sin\phi),$ 
ignoring the lower-order term, for the residual-space dimensions $n$ relevant to the non-JumpReLU models considered here. The resulting curves exhibit a pronounced ``hockey-stick'' shape; for very small \(\phi\), \(\Lambda(\phi)\) increases sharply, indicating a regime in which extremely many directions may still count as nearly orthogonal, making the distinction between true inactivity and weak alignment highly ambiguous. Beyond this range, the growth becomes less extreme. We therefore choose \(\theta = 0.1\). This provides geometric motivation for the choice, while we emphasize that it remains a pragmatic threshold for defining a distinction between \emph{deactivated} and \emph{suppressed} in the absence of a learned activation offset.

\begin{figure}[!tbp]
    \centering
    \includegraphics[width=\linewidth]{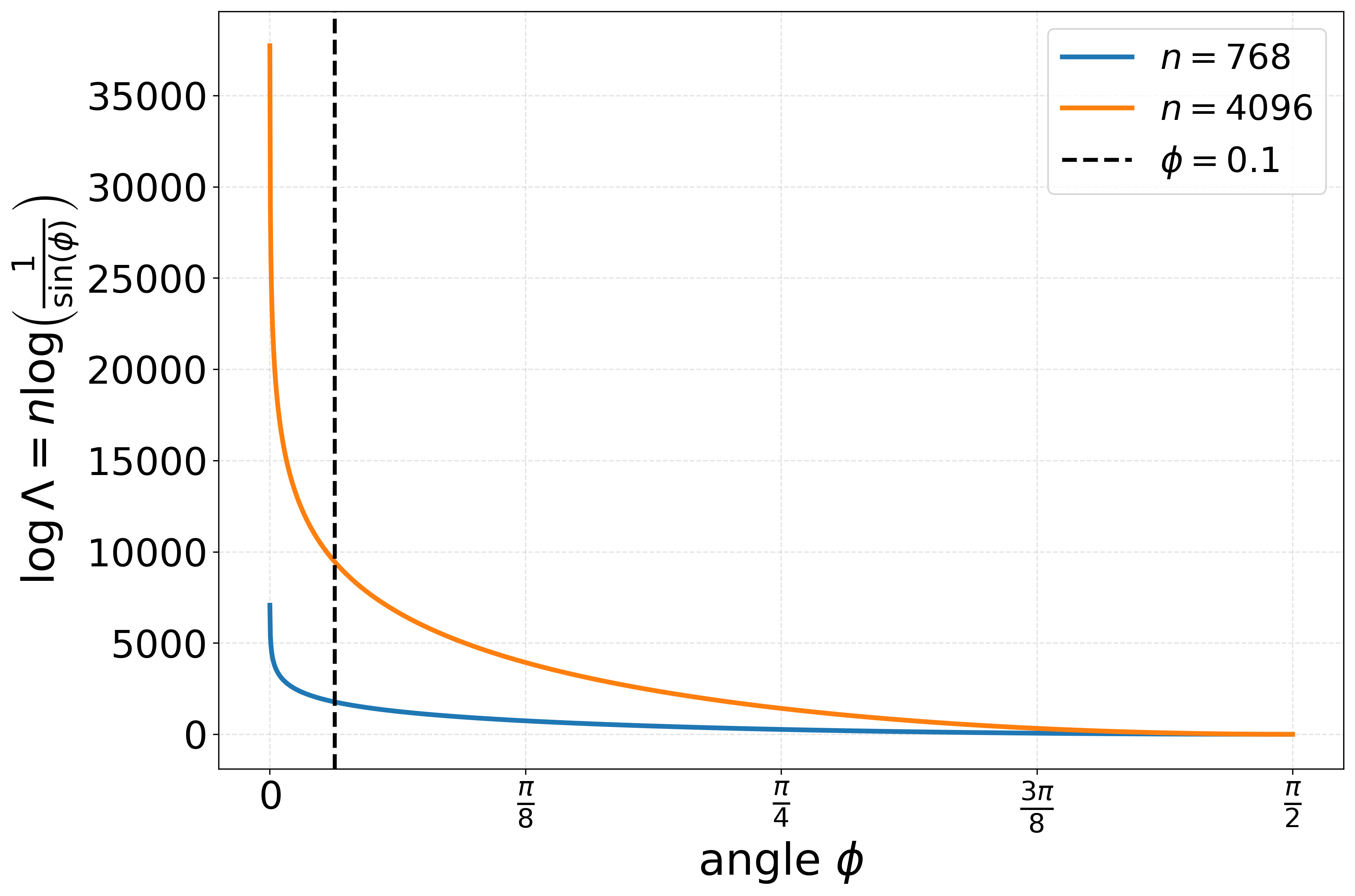}
    \caption{
    Value of \(\log \Lambda(\phi)=n\log(1/\sin\theta)\) as a function of the angle \(\theta\), shown for the residual-stream dimensions used by the non-JumpReLU models in our experiments: \(n=768\) for GPT-2 Small and \(n=4096\) for both Llama~3.1~8B and Mistral-7B. The dashed vertical line marks the chosen threshold \(\theta=0.1\).}
    \label{fig:orth-threshold}
\end{figure}
\FloatBarrier
\subsection{Additional Results}

We report the sweep results for all remaining investigated models and SAE variants, analogous to Figure~\ref{fig:2} in the main text. Each figure corresponds to one model, and each row within a figure corresponds to a different SAE type trained for that model. The panel layout follows Figure~\ref{fig:2}. Panel~(a) shows the lost-latent rate \(\mathcal{R}\) from Equation~\eqref{eq:lost_rate} as a function of the number of prefixed adjectives \(k\), for selected layers across model depth. The inset summarizes the full layerwise trend for \(k=1\). Panels~(b)--(d) focus on \(k=1\), which is representative because these quantities depend only weakly on \(k\) across the investigated models. See Appendix~\ref{app:k_dependence} for investigating \(k\)-dependence.\color{black}\\
Panel~(b) considers latents that are active for the base prompt \(t_0\) but lost on the noun signature of the more specific prompt at layer \(\ell\). It reports the fraction of these lost latents that were previously active upstream when probing the full content sequence with the SAE of layer \(\ell\); the inset restricts the same upstream check to noun signatures only. See Section~\ref{sec:info_present_upstream}. Panel~(c) analyzes the previously active subset from panel~(b), distinguishing whether the corresponding layer-\(\ell\) pre-activations fall into the near-zero regime from Equation~\eqref{eq:ortho} or the negative regime from Equation~\eqref{eq:negative}. The main plot evaluates this distinction over the full sequence at layer \(\ell\), while the inset restricts it to the noun signature.\\
Panel~(d) applies the same near-zero/negative distinction to the complementary subset of lost latents that were never previously active under the upstream probing criterion. The main plot evaluates whether these latents remain near-zero or negative throughout the full upstream computation, while the inset restricts the evaluation to noun signatures. Together, panels~(b)--(d) distinguish whether lost latents were previously recoverable, and whether their binary loss reflects near-threshold inactivity or stronger negative pre-activation.\\
\\
A shared observation across all investigated models (Figures~\ref{lost_sweepfig_gemma270m}~to~\ref{fig:lostsweep_llama31_8b}) is that the lost-latent rate \(\mathcal{R}\) increases with the number of prefixed adjectives \(k\), typically by roughly \(20\) percent over the five investigated construction steps. Interestingly, this increase is not linear, as the slope becomes noticeably less pronounced beyond \(k=2\). A possible interpretation, is that the model may distinguish between a noun modified by only a few adjectives and a qualitatively different regime resembling an extended adjective enumeration. Once the latter regime is reached, the precise number of additional adjectives may become less relevant. This interpretation is further suggested by Appendix~\ref{app:stepwise_lost_rates}. However, this interpretation should be understood as suggestive rather than conclusive.\\
A second observed trend is the increase of \(\mathcal{R}\) with model depth, shown in the inlay plots of collumn a), for the simplest deviation from the base prompt, i.e. at \(k=1\). Across cases, the lost-latent rate approaches a plateau around \(50\%\) in deeper layers. from the main curves in column~(a), one can further extrapolate this profile to shift upward by roughly another \(20\) percent as \(k\) increases, which has been verified but is not explicitly shown here. Therefore, the loss rate is consistently smaller in earlier layers, which are commonly associated more strongly with lexical processing \citep{elhage2021mathematical, geva-etal-2022-transformer}. In this regime, the SAEs appear to adhere to a more set-consistent behaviour between the compared inputs. However, at greater computational depth the lost rate $\mathcal{R}$ increases, suggesting that the model transitions toward a different internal conceptualization of the more specific prompt, indicating that the SAE might capture additional structure in higher layers not explained by a "bag-of-features"-view on the dictionary latents.\\
\label{app:sweep_lost_feats}
\renewcommand{\sweepfig}[3]{%
\begin{figure*}[!tp]
    \centering
    \includegraphics[
        width=\textwidth,
        height=0.72\textheight,
        keepaspectratio
    ]{figures/#1}
    \caption{#2}
    \label{#3}
\end{figure*}
}
\sweepfig{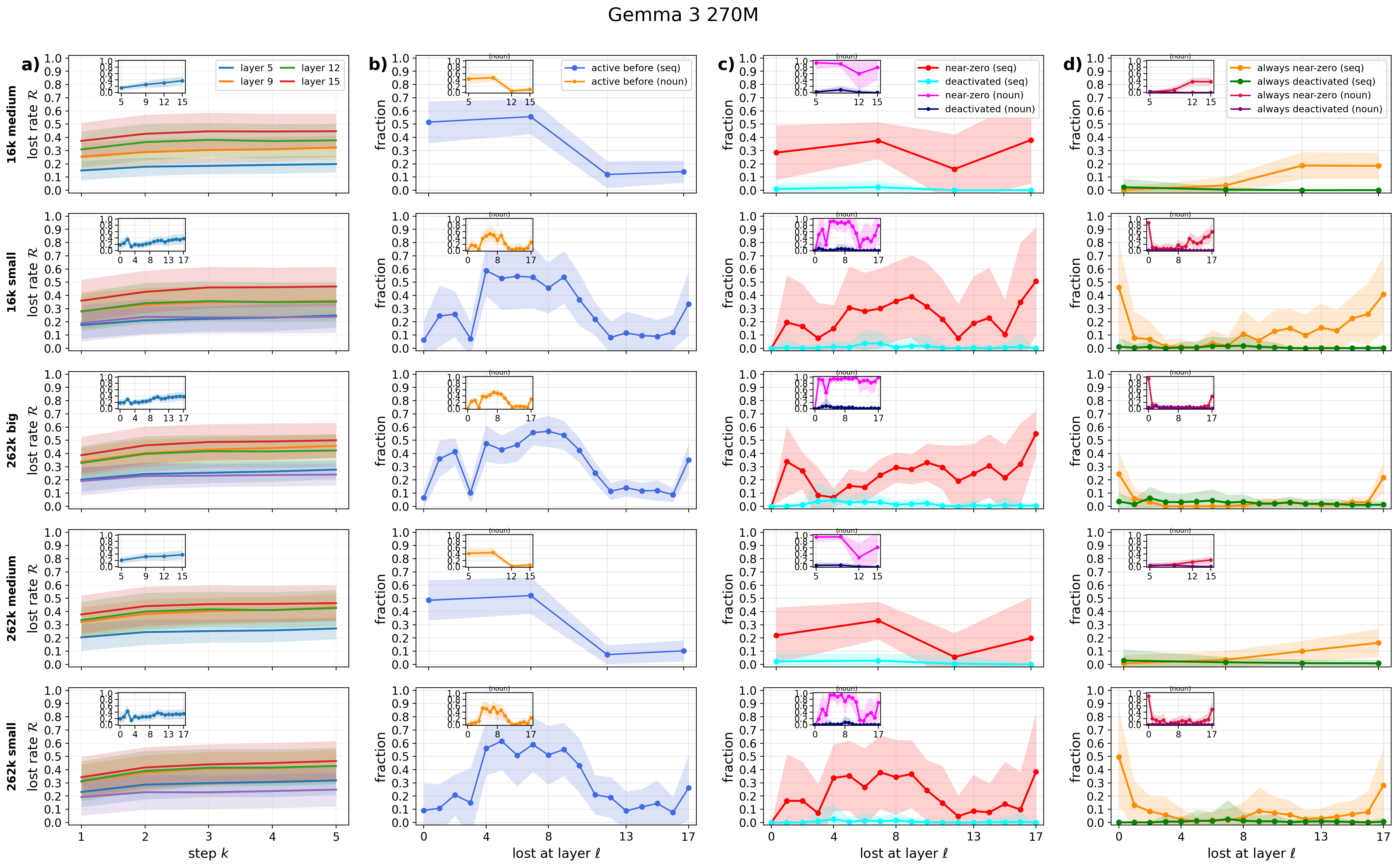}{Lost features under semantically consistent inputs for Gemma 3 270M.}{lost_sweepfig_gemma270m}
\color{white}.\color{black}\\
The results in panel~(b) show a structured dependence on model family and depth. For the Gemma SAEs and GPT-2 Small, the fraction of lost latents that had been active upstream generally rises from early layers to a maximum in the middle layers, typically around \(50\%\)--\(80\%\), then decreases again and often increases once more toward the final layers. The initial increase is partly expected, as for deeper layers, there are more preceding layers in which a latent could have appeared. This makes the subsequent decrease more noteworthy, since it indicates that lost latents in later layers are not simply more likely to have been active before. One possible interpretation is that different phases of the model computation use a greater variety of latent repertoires, so that later-layer losses are not always recoverable as earlier activations of the same SAE directions. Llama~3.1~8B shows a different pattern, with the previously-active fraction remaining comparatively stable at roughly \(40\%\) across layers. Mistral-7B reaches relatively high values of roughly $80\%$, indicating that the lost latents remain relevant in the computation. However, SAEs for only few layers are available, so we avoid drawing strong conclusions about the layerwise trend.\\
The main panel measures upstream activity over all token positions in the full prompt sequence. The inset repeats the analysis using only noun signatures in previous layers. For most Gemma SAEs, the inset closely tracks the main curve, suggesting that when a lost latent was previously active, it was usually active on the noun signature at a previous layer rather than only elsewhere in the sequence. GPT-2 Small and Llama~3.1~8B are partial exceptions. In early layers, the noun-restricted curves are substantially lower than the full-sequence curves, indicating that upstream activity of lost latents is less tightly coupled to the noun position in these models.\\
\\
The plots in column~(c) further decompose the previously active lost latents according to their layer-\(\ell\) pre-activations. For most Gemma SAEs, the fraction that is negative over the full sequence is non-zero but typically small, while a larger fraction falls into the near-zero regime. The precise layerwise trend varies across model sizes and SAE variants, but the inset shows a clearer pattern. When the analysis is restricted to the noun signature, the negative fraction remains low, whereas the near-zero fraction is often very high. Thus, for Gemma SAEs, many previously active lost latents show weak rather than negative alignment at the noun position.\\
A broadly similar pattern holds for Llama~3.1~8B, where noun-restricted near-zero rates are substantially higher than the corresponding negative rates. Mistral-7B is the strongest exception, as for both over the full sequence and on the noun signature, almost all previously active lost latents are classified as negative, suggesting that when such latents are lost, they are more strongly removed from the representation. GPT-2 Small shows no comparably clear pattern under this classification, as neither the near-zero nor the negative regime accounts for a substantial fraction of previously active lost latents.\\
\\
Finally, column~(d) analyzes lost latents that were never previously active under the upstream probing criterion. For Gemma SAEs, the fraction that is always negative across the full upstream computation is typically low. By contrast, the fraction that remains within the near-zero regime is often larger, frequently increases with layer depth. The inset restricts the same analysis to noun signatures and often strengthens this trend, particularly in earlier layers. This suggests that many latents classified as never previously active are not persistently negatively aligned throughout the computation. On the noun position, they instead often show only weak alignment with the current residual-stream state.\\
For Llama~3.1~8B, the main panel shows a similar tendency to the Gemma SAEs. The always-negative fraction remains low, while the near-zero fraction accounts for a larger share of never-previously-active lost latents. The noun-restricted inset further increases this fraction, suggesting that several of these latents show weak rather than clear negative alignment on the noun position. GPT-2 Small shows no clear separation, as both fractions remain small. Mistral-7B shows the clearest contrast, as in panel~(c), lost latents mostly fall into the negative regime, indicating stronger negative rather than weak subthreshold alignment.
\color{black}

\sweepfig{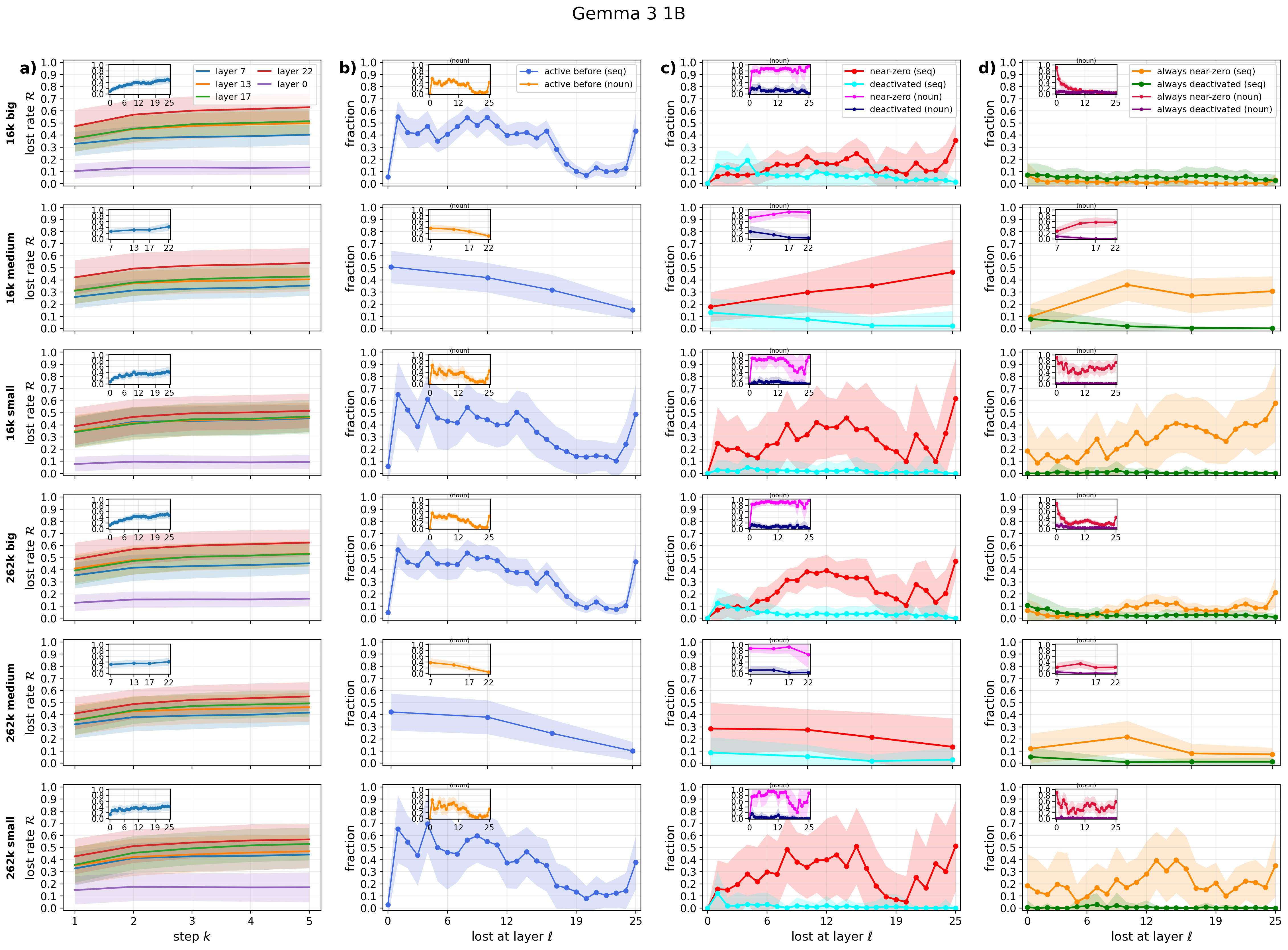}{Lost features under semantically consistent inputs for Gemma 3 1B.}{fig:lostsweep_gemma_1b}
\sweepfig{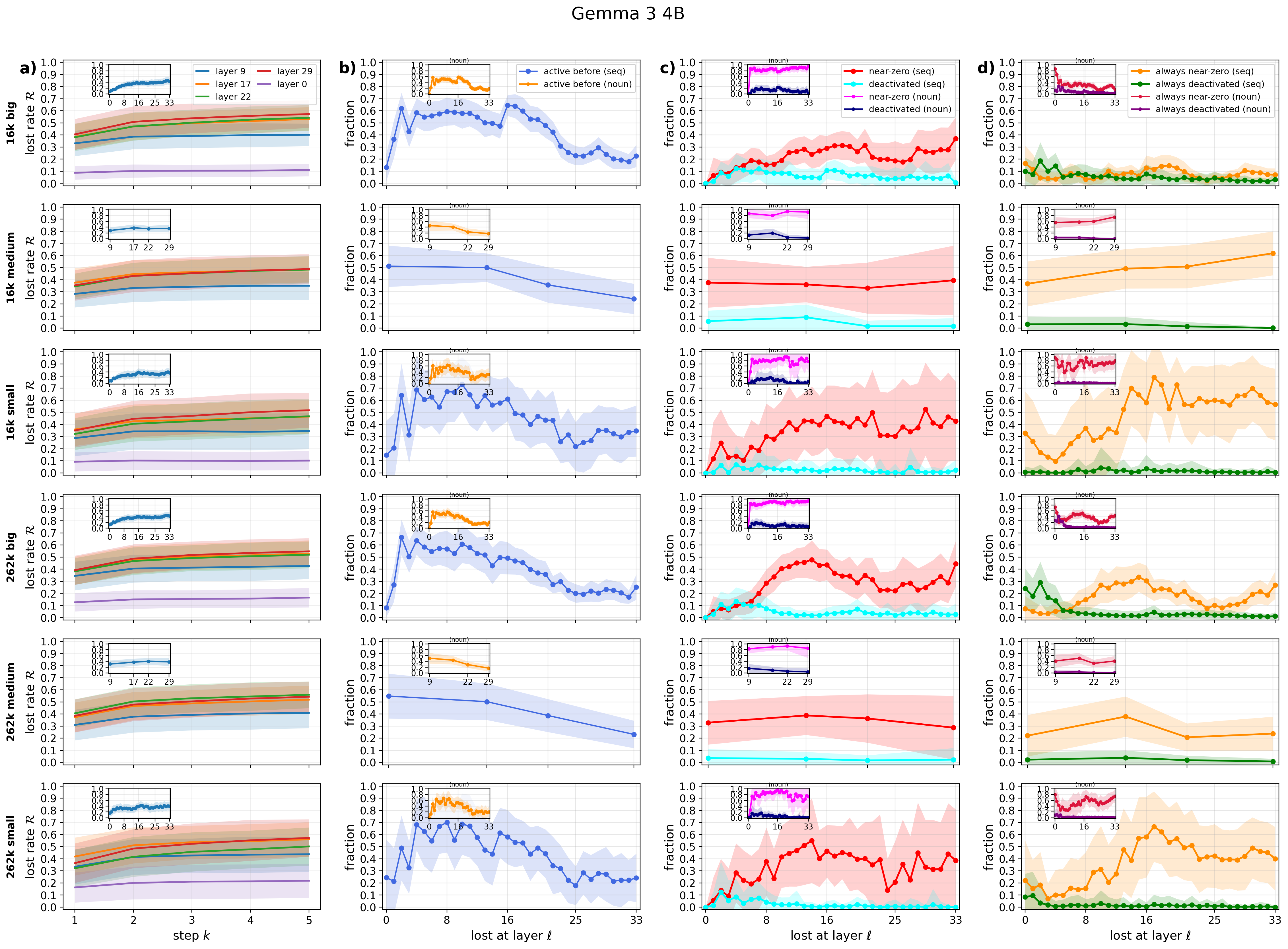}{Lost features under semantically consistent inputs for Gemma 3 4B.}{fig:lostsweep_gemma_4b}
\sweepfig{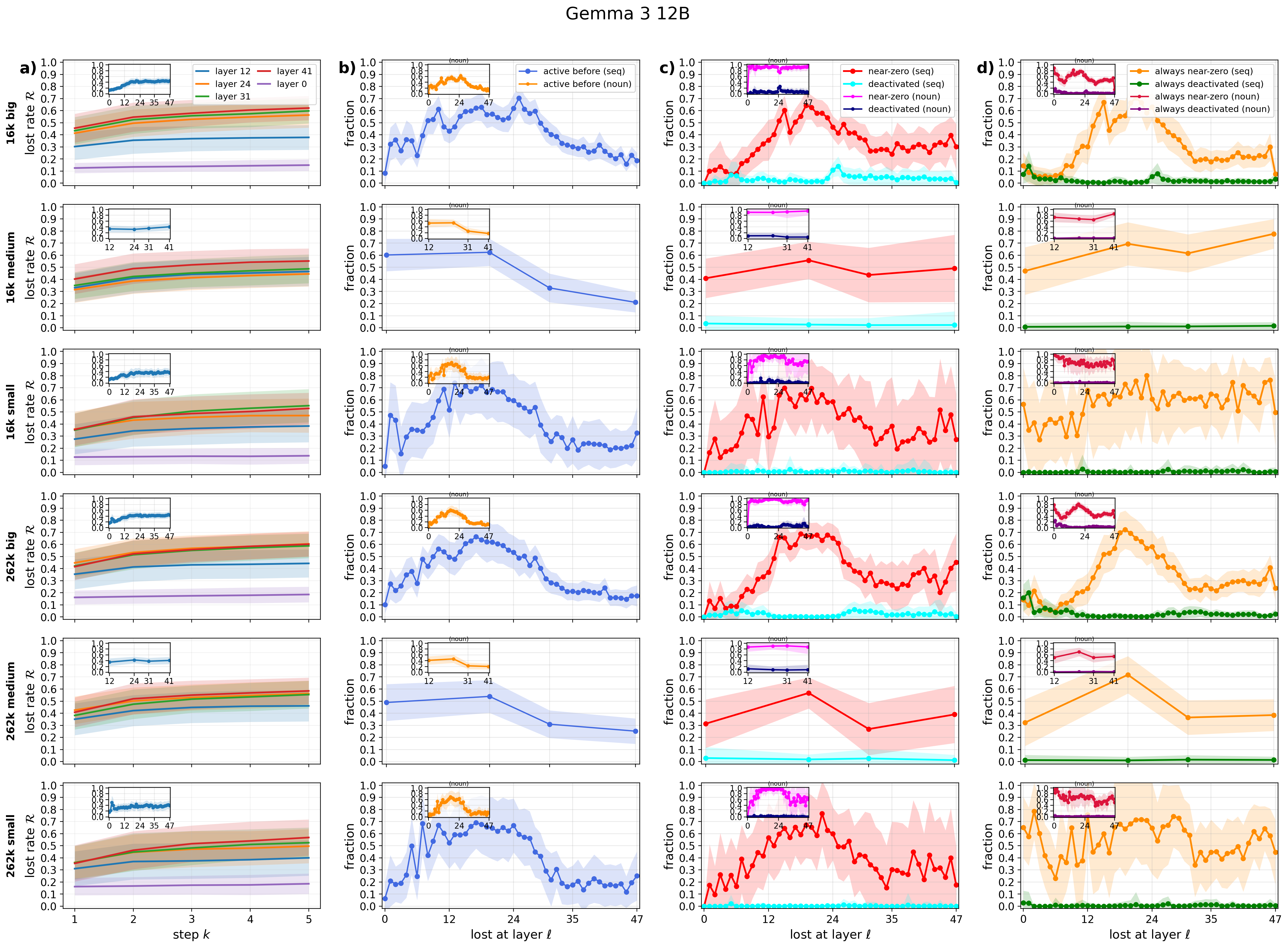}{Lost features under semantically consistent inputs for Gemma 3 12B.}{fig:lostsweep_gemma_12b}
\sweepfig{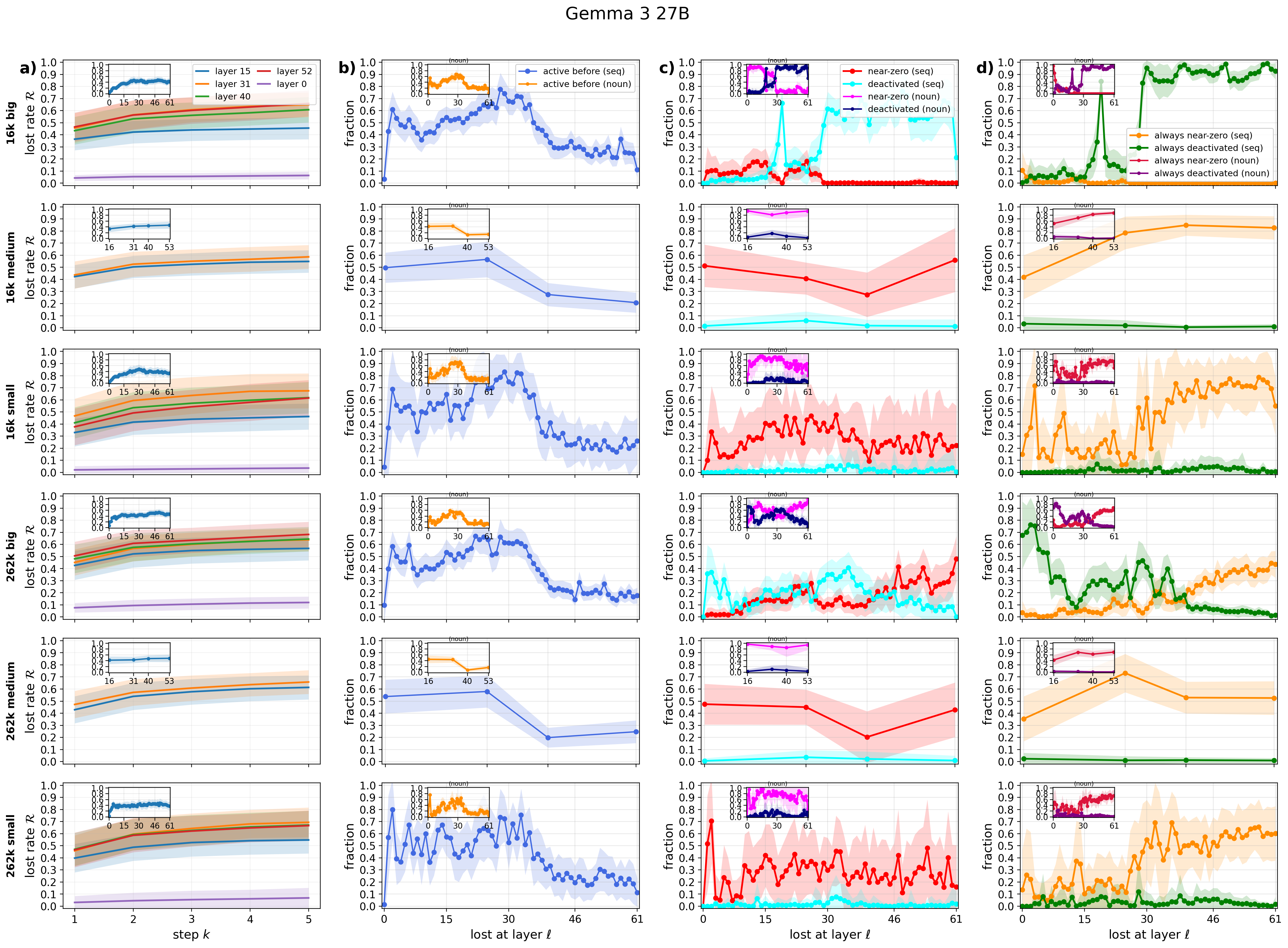}{Lost features under semantically consistent inputs for Gemma 3 27B.}{fig:lostsweep_gemma_27b}
\sweepfig{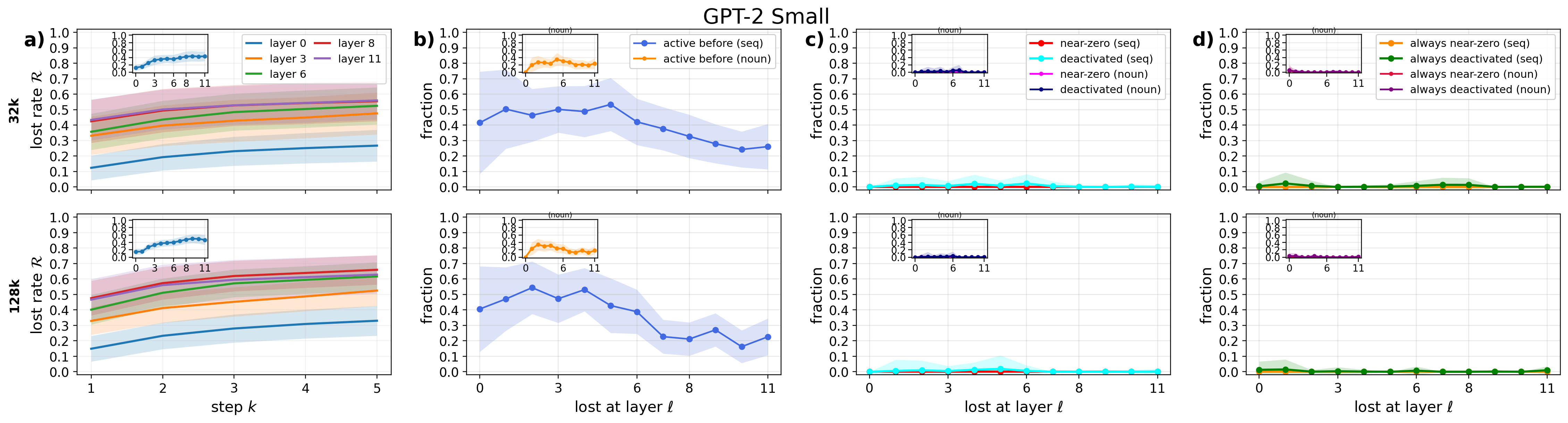}{Lost features under semantically consistent inputs for GPT-2 Small.}{fig:lostsweep_gpt2_small}
\sweepfig{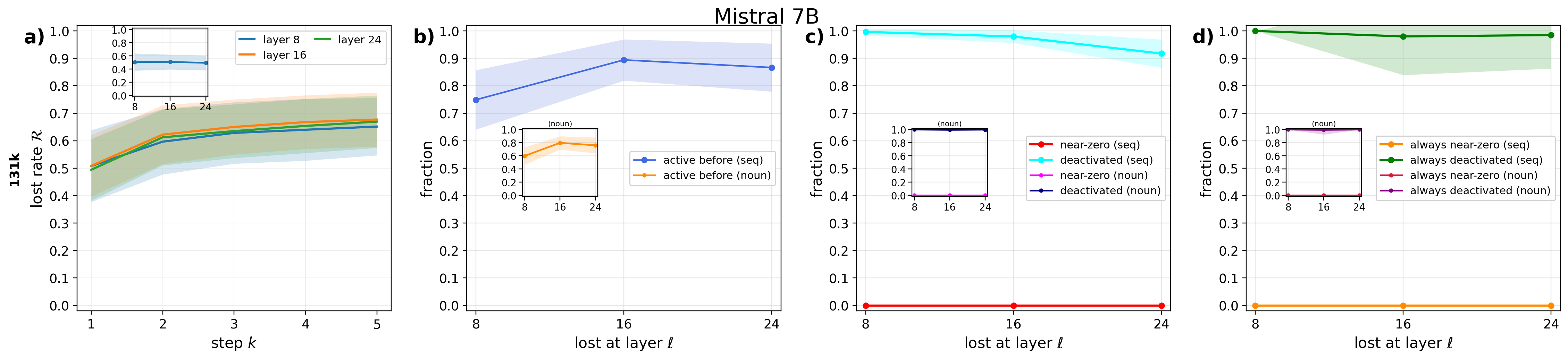}{Lost features under semantically consistent inputs for Mistral 7B.}{fig:lostsweep_mistral_7b}
\sweepfig{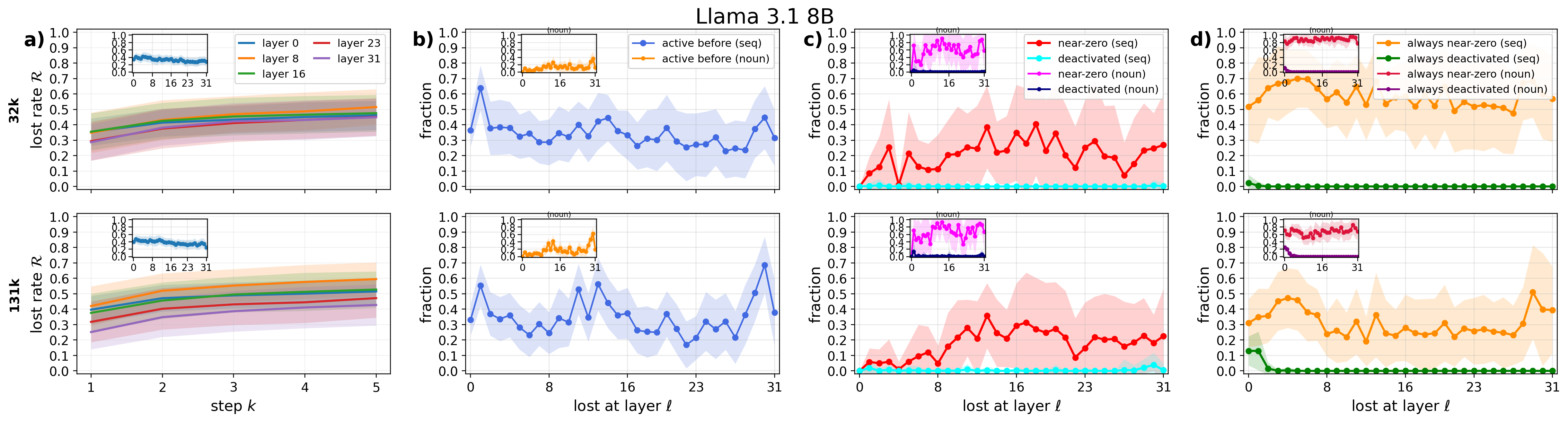}{Lost features under semantically consistent inputs for Llama 3.1 8B.}{fig:lostsweep_llama31_8b}

\FloatBarrier
\begin{figure*}[!t]
    \centering
    \includegraphics[width=\textwidth]{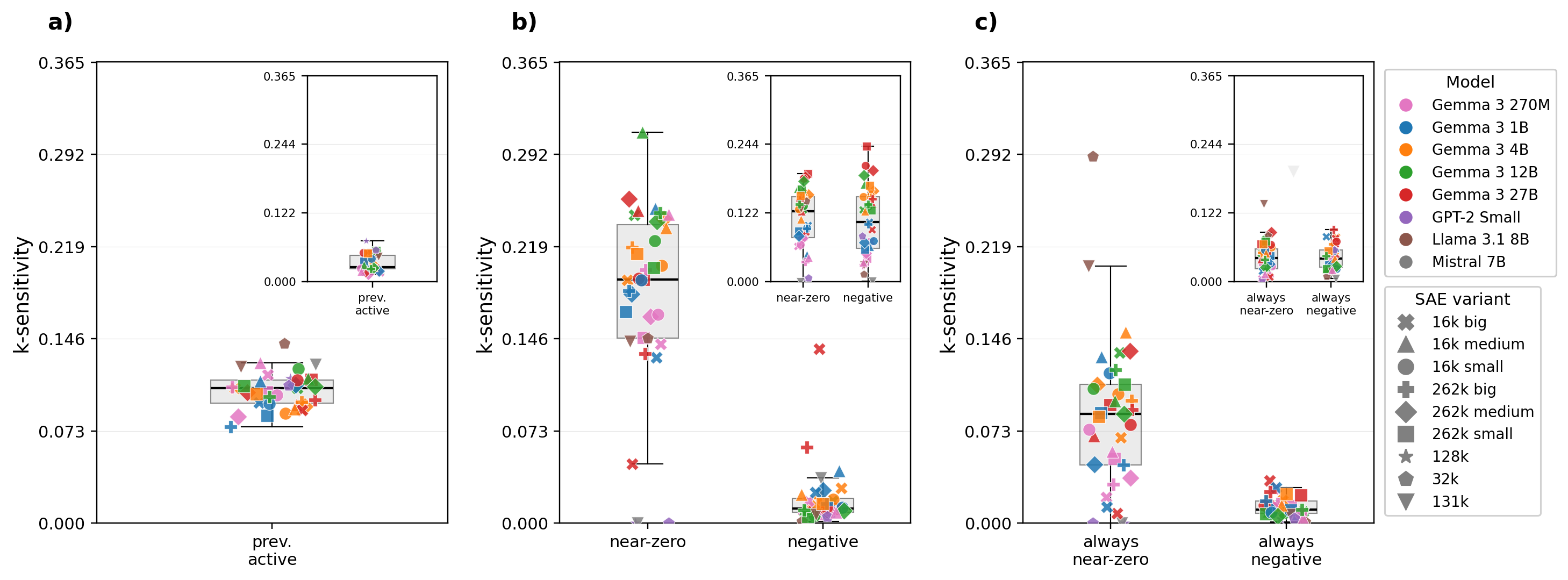}
    \caption{
    Sensitivity of the quantities in panels~(b)--(d) of Figure~\ref{fig:2}  and Figures~\ref{lost_sweepfig_gemma270m}--\ref{fig:lostsweep_llama31_8b} to the number of added adjectives \(k\). For each measured quantity \(q\), model, and SAE variant, we compute the layer-averaged range across \(k\), \(\Delta_k(q;m,s)\), as defined in Equation~\eqref{eq:k_sensitivity}. Each point corresponds to one model--SAE pair. Similar to the layout from Figures~\ref{lost_sweepfig_gemma270m}--\ref{fig:lostsweep_llama31_8b}, main panels show the full-sequence analyses, while insets show the corresponding noun-only analyses. 
    }
    \label{fig:k_sensitivity_lost_latents}
\end{figure*}
\subsubsection{Sensitivity on $k$}
\label{app:k_dependence}
In Figure~\ref{fig:2} and Figures~\ref{lost_sweepfig_gemma270m}--\ref{fig:lostsweep_llama31_8b}, panels~(b)--(d) are shown for \(k=1\). To verify that this choice is representative, we repeat the same analyses for all available construction steps \(k\). Let \(q_{m,s,l}^{(k)}\) denote one of the measured quantities for model \(m\), SAE variant \(s\), layer \(\ell\), and construction step \(k\). We define its \(k\)-sensitivity as the layer-averaged range across \(k\):
\begin{equation}
\begin{split}
    &\Delta_k(q;m,s)
    =\\\
    &\frac{1}{|\mathcal{L}_{m,s}|}
    \sum_{l \in \mathcal{L}_{m,s}}
    \left(
    \max_k q_{m,s,l}^{(k)}
    -
    \min_k q_{m,s,l}^{(k)}
    \right),
    \label{eq:k_sensitivity}
\end{split}
\end{equation}
where \(\mathcal{L}_{m,s}\) is the set of layers for which the corresponding SAE variant is available. This yields one \(k\)-sensitivity value per measured quantity, model, and SAE variant. Lower values indicate that the quantity depends only weakly on the number of added adjectives.\\
Figure~\ref{fig:k_sensitivity_lost_latents} summarizes these values for the quantities used in panels~(b)--(d). Panel~(a) corresponds to the upstream-activity analysis from panel~(b), panel~(b) to the near-zero/negative decomposition of previously active lost latents from panel~(c), and panel~(c) to the same decomposition for never-previously-active lost latents from panel~(d). Main panels show the full-sequence analyses, while insets show the noun-only variants. Overall, the \(k\)-sensitivity values remain moderately small compared to the layerwise trends observed in the figures, indicating that varying \(k\) does not qualitatively change the conclusions.

\subsubsection{Stepwise Evaluation for Lost-Latent Rates}
\label{app:stepwise_lost_rates}

So far, we defined a latent as lost if it was active on the base noun prompt \(t_0\) but inactive on a more specific prompt \(t_k\), using the neutral prompt template from Section~\ref{sec:compositional_consistency}. Here, we extend this analysis to consecutive refinement steps. That is, for each step \(i\), we measure latents that are active on the noun signature of \(t_i\) but inactive on the noun signature of \(t_{i+1}\). The rest of the setup remains unchanged. The comparison is made on the noun token positions under the same prompt template.\\
Formally, we define the stepwise lost-latent rate as
\begin{equation}
\mathcal{R}^{(i \rightarrow i+1)}_l
=
\frac{
\left|\overline{F}_l(t_i)\setminus \overline{F}_l(t_{i+1})\right|
}{
\left|\overline{F}_l(t_i)\right|
},
\label{eq:stepwise_lost_rate}
\end{equation}
thus asking to which extend additional adjectives alter the latent signature.\\
\begin{figure*}[t]
    \centering
    \includegraphics[width=\textwidth]{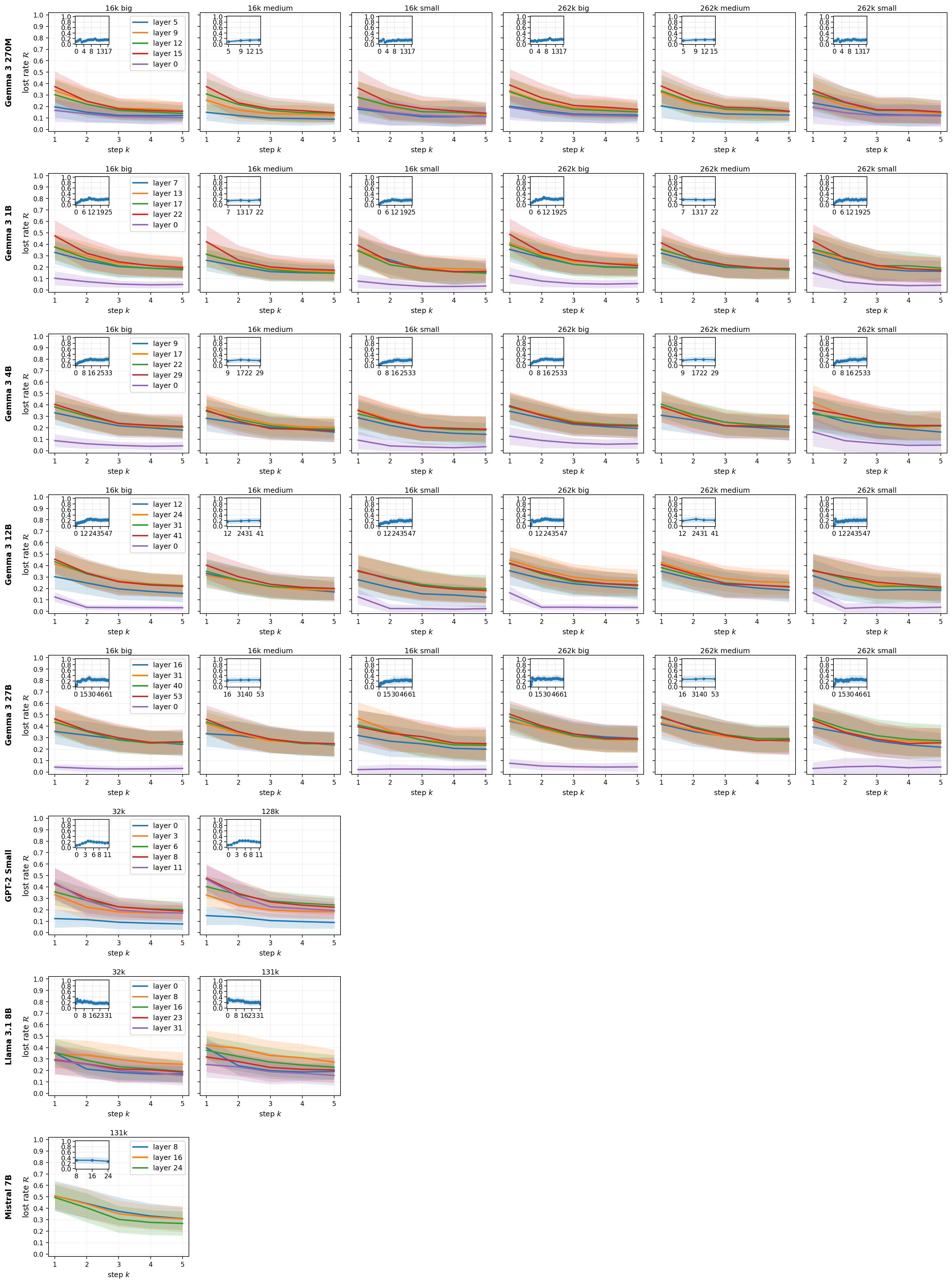}
    \caption{
    Stepwise lost-latent rates for consecutive semantic refinements. Each row corresponds to one model, and panels within a row show the SAE variants evaluated for that model from left to right. For each construction step \(i\), we measure the fraction of latents active on the noun signature of \(t_i\) that are no longer active on the noun signature of \(t_{i+1}\), as defined in Equation~\eqref{eq:stepwise_lost_rate}. The inset reports the layerwise trend for \(i=4\), i.e.\ the transition \(t_4 \rightarrow t_5\), since the analogous layerwise trend for the first transition \(t_0 \rightarrow t_1\) is already shown in Figure~\ref{fig:2} and Figures~\ref{lost_sweepfig_gemma270m}--\ref{fig:lostsweep_llama31_8b}~(a).
    }\label{fig:stepwise_lost_rates}
\end{figure*}
Results are shown in Figure~\ref{fig:stepwise_lost_rates}. The lost-latent rate remains substantial across refinement steps. However, the largest drop typically occurs from \(t_0\) to \(t_1\), i.e.\ when the representation first changes from a bare noun to an adjective noun phrase. Later refinements still induce losses, but these are usually smaller. This possibly indicates, that the first adjective introduces a qualitatively larger representational shift, whereas subsequent adjectives are incorporated into an already modified noun-phrase representation. Thus, once the model has moved from an isolated noun to description prescribed with adjectives, further refinements may be somewhat easier to accommodate without replacing as much of the SAE signature. The inset shows the layerwise trend for the final consecutive refinement, \(t_4 \rightarrow t_5\).
\clearpage
\clearpage
\begin{figure}[!tbp]
    \centering
    \includegraphics[width=\columnwidth]{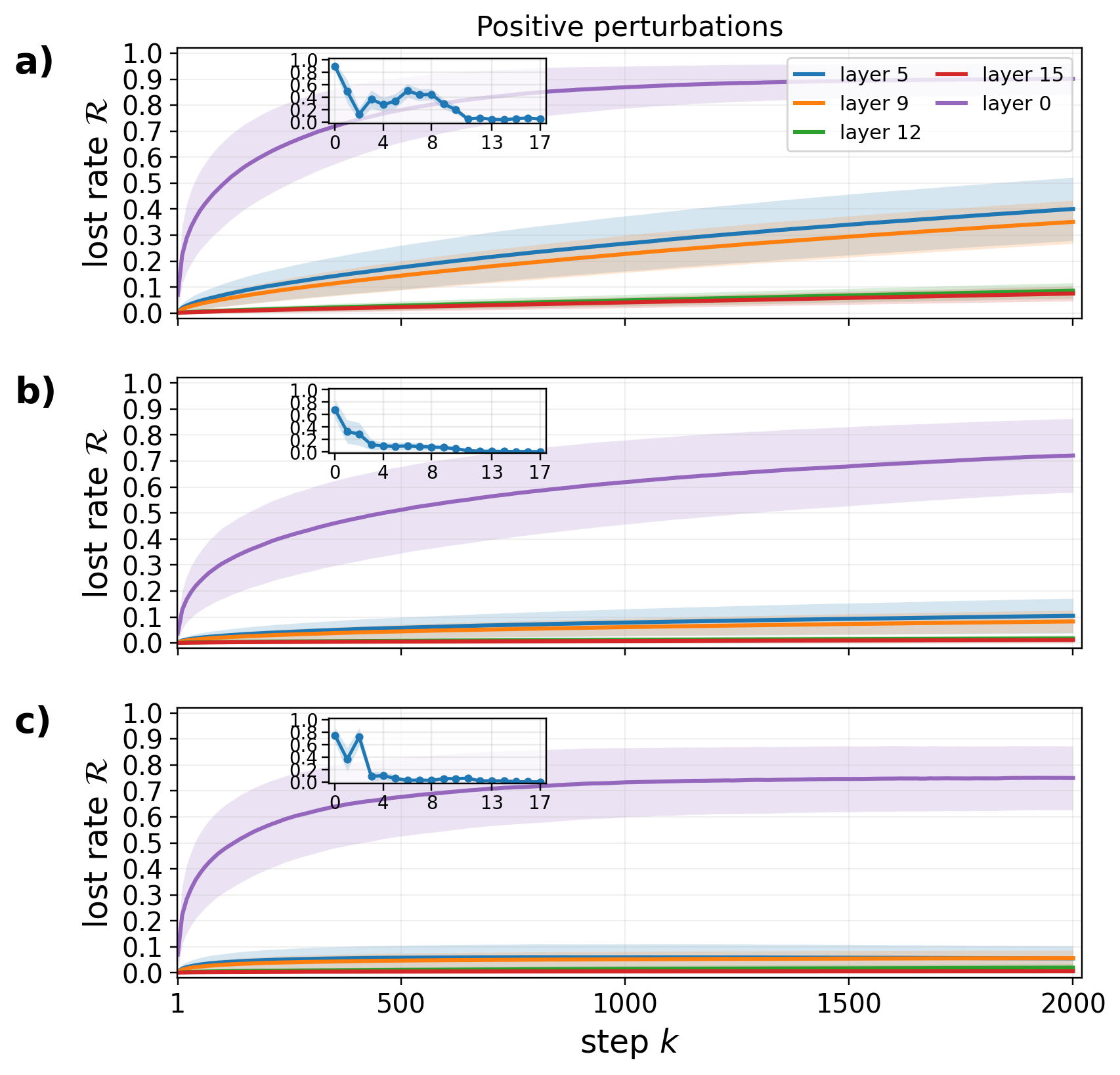}
        \caption{
All panels (a)--(c) show the lost rate from Equation~\eqref{eq:lost_rate} for steering-based constructions from the base case \(t_0\) as the number of steering steps \(k\) increases according to Equation~\eqref{eq:steering}. Four layers across model depth are shown. Panel (a) uses only positive coefficients (\(\mu=\beta/2\)), panel (b) uses equally likely positive and negative coefficients (\(\mu=0\)), and panel (c) uses only negative coefficients (\(\mu=-\beta/2\)). Each construction uses up to 2000 latents. Results are aggregated over 50 random steering runs per dataset instance and across the dataset. The width of the distribution is chosen to be $\beta = 10$. The inset plots show additionally the trend through model depth at $k=1500$ in the construction.
}
    \label{fig:steering}
\end{figure}
\section{Interference}
\label{app:interference_vs_neg_feats}
When assuming that the observed violation of set consistency was actually noise, one possible explanation is that the model combines several positively contributing SAE latents in such a way that they result in a negative projection onto another latent direction previously active, as made plausible by \citet{scaling_templeton_2024}. In that case, some latents would appear absent not because the model explicitly \emph{suppresses} them \cite{mayne2024can, elhage2021mathematical}, but as a byproduct of superposition.\\
\\
To probe this, we simulate controlled perturbations of the residual stream using SAE-resolvable directions. For each base prompt \(t_0\), we first extract the residual-stream activation at layer \(\ell\), denoted \(\mathcal{V}_{\mathrm{res}}^{(l)}\), at the target token position. Specifically, \(t_0\) is inserted into the template also used above "\texttt{This is a \{$t_0$\}.\_}", and \(\mathcal{V}_{\mathrm{res}}^{(l)}\) is taken from the last token position of the inserted noun. We then add synthetic perturbations constructed from randomly sampled SAE latent directions. To obtain progressively more complex deviations, we build these perturbations cumulatively. At step \(k\), we add \(k\) sampled latent directions \(f_i\), each weighted by a random coefficient \(\alpha_i\) drawn uniformly from an interval centered at \(\mu\) with width \(\beta\). The resulting perturbation is rescaled to match the norm of the original residual activation, so as to remain on a comparable scale. Formally,
\begin{equation}
\tilde{t}_k = \sum_{i=1}^k \alpha_i f_i,
\qquad
\mathcal{T}_k =
\mathcal{V}_{\mathrm{res}}^{(l)}
+
\tilde{t}_k
\frac{\lVert \mathcal{V}_{\mathrm{res}}^{(l)} \rVert}
     {\lVert \tilde{t}_k \rVert}.
\label{eq:steering}
\end{equation}
Here, \(f_i\) denotes a SAE latent direction from the SAE decoder matrix trained on layer \(\ell\). Note that this is in accordance with the LRH formulation in Equation~\eqref{eq:decomposition}, as the bias term is implicitly involved in $\mathcal{V}_{\mathrm{res}}^{(l)}$. For each dataset example, we generate 50 random steering trajectories from the base prompt \(t_0\) and compute the lost-latent rate from Equation~\eqref{eq:lost_rate} by comparing \(t_0\) with \(\tilde{t}_k\). Reported values are averaged first across trials and then across dataset instances. Throughout, we use a uniform coefficient distribution with width \(\beta=10\), a pragmatic choice that induces latent interactions on a meaningful scale and is adequate for the qualitative analysis considered here.
This construction serves as a qualitative probe of the origin of latent loss. In particular, it tests whether the observed lost-latent rates from Section~\ref{sec:degradation_from_compositional_consistency} can already be reproduced by interference among positively contributing SAE latents. 
\\
The perturbation results suggest that positive-latent interference, i.e. \(\mu=\beta/2\), may be one contributing mechanism behind the lost-latent rates observed under real prompts. In particular, Figure~\ref{fig:steering}~(a), and more broadly the results across models in Appendix~\ref{app:sweep_steering}, show that purely positive steering can already induce loss rates of an order observed under real input constructions from Section~\ref{sec:compositional_consistency} for some layers. This is qualitatively consistent with the idea that adding concepts, i.e. moving in positive latent directions, can nevertheless produce interference among active latents, so that some previously active latents are no longer recovered by the SAE. This remains compatible with the set-consistency intuition from Section~\ref{sec:motivation_set_consistency}. The more specific prompt may indeed recruit additional latents while preserving the content of the less specific prompt, yet the resulting superposition can still induce latent loss as a side effect. At the same time, this mechanism does not explain the empirical loss patterns in a sufficiently consistent way across layers and models when compared to the real-input results from Section~\ref{sec:degradation_from_compositional_consistency} and Appendix~\ref{app:sweep_lost_feats}. This can be red of the inset plots, which show the trend across model depth at step $k=1500$ in the construction, which do not align with the trend that the observed lost rate tends to increase with model depth.\\
We also test perturbations with mixed-sign coefficients, i.e. \(\mu=0\), and purely negative coefficients, i.e. \(\mu=-\beta/2\), to assess whether adding negative contributions increases the explanatory power of the construction. Empirically, however, these settings do not yield a clearer match to the observed loss patterns and often induce weaker effects instead. See Figure~\ref{fig:2}~(b)--(c) and Appendix~\ref{app:sweep_steering}. As a qualitative side observation, this is in line with the linear representation hypothesis introduced in Section~\ref{sec:terms_and_defs}, according to which semantic content is primarily encoded by positive latent directions \citep{10.5555/3692070.3693675, elhage2022toy, engels2025not}. Overall, the perturbation experiments suggest that positive-latent interference is a plausible contributing factor, but does not offer a sufficient explanation of the lost-latent rates observed under real inputs.
\subsection{Additional Results}
\label{app:sweep_steering}
Similar to Figure~\ref{fig:steering}, the following figures report the results of the latent steering experiment described above for all remaining models and SAE types. Each figure corresponds to one of the investigated models, and each row of panels within a figure corresponds to an SAE type trained on that model. The three columns represent different variants of the experiment. Column~(a) shows the steering setup with \(\mu=\beta/2\), that is, using only positive coefficients in the linear construction. Column~(b) shows the balanced setting with \(\mu=0\), corresponding to a mixture of positive and negative latents. Column~(c) shows the setting \(\mu=-\beta/2\), in which only negative coefficients are used. The inset displays the full layer-wise trend at step \(k=1500\).\\
\\
From Figure~\ref{fig:constructsweep_gemma_270m} to Figure~\ref{fig:constructsweep_gemma_27b}, we observe for the Gemma SAEs that relevant loss rates arise primarily for positive coefficients (column~(a), and mainly in the early and middle layers, as shown by the inset plots. By contrast, for mixed (column~(b) and purely negative coefficients (column~(c), the loss rates remain generally low. As discussed above, this is consistent with the view that semantic content is encoded primarily in the positive latent direction. Under this interpretation, interference from positive latent projections may contribute to the observed loss in early-to-mid layers and could therefore partially explain the effect reported in Section~\ref{sec:motivation_set_consistency}, although this remains speculative. At the same time, this account does not explain the layerwise trend itself.\\
For the other models, this pattern does not hold. For GPT-2 (Figure~\ref{fig:constructsweep_gpt2_small}), Llama-3.1-8B (Figure~\ref{fig:constructsweep_llama31_8b}), and Mistral-7B (Figure~\ref{fig:constructsweep_mistral_7b}), all three coefficient settings yield high loss rates, with steep increases as \(k\) grows. In these cases, interference of various kinds could in principle explain the loss rates observed in Appendix~\ref{app:sweep_lost_feats}. However, results there show that the loss rates on real input data are of similar magnitude across models. Since the steering results differ qualitatively across models, it seems unlikely that this proposed interference effect is the dominant underlying cause. Still, the Gemma results may be seen as more compatible with LRH, insofar as negative latents do not substantially alter the pattern of previously active latents, suggesting an interpretation that meaning is encoded mainly along the positive latent direction, more aligned with interpretations of the strong LRH \citep{10.5555/3692070.3693675, elhage2022toy}. Why SAEs for the other models behave differently in this respect, and what this implies for LRH more generally, remains an open question for future work.
\clearpage
\sweepfig{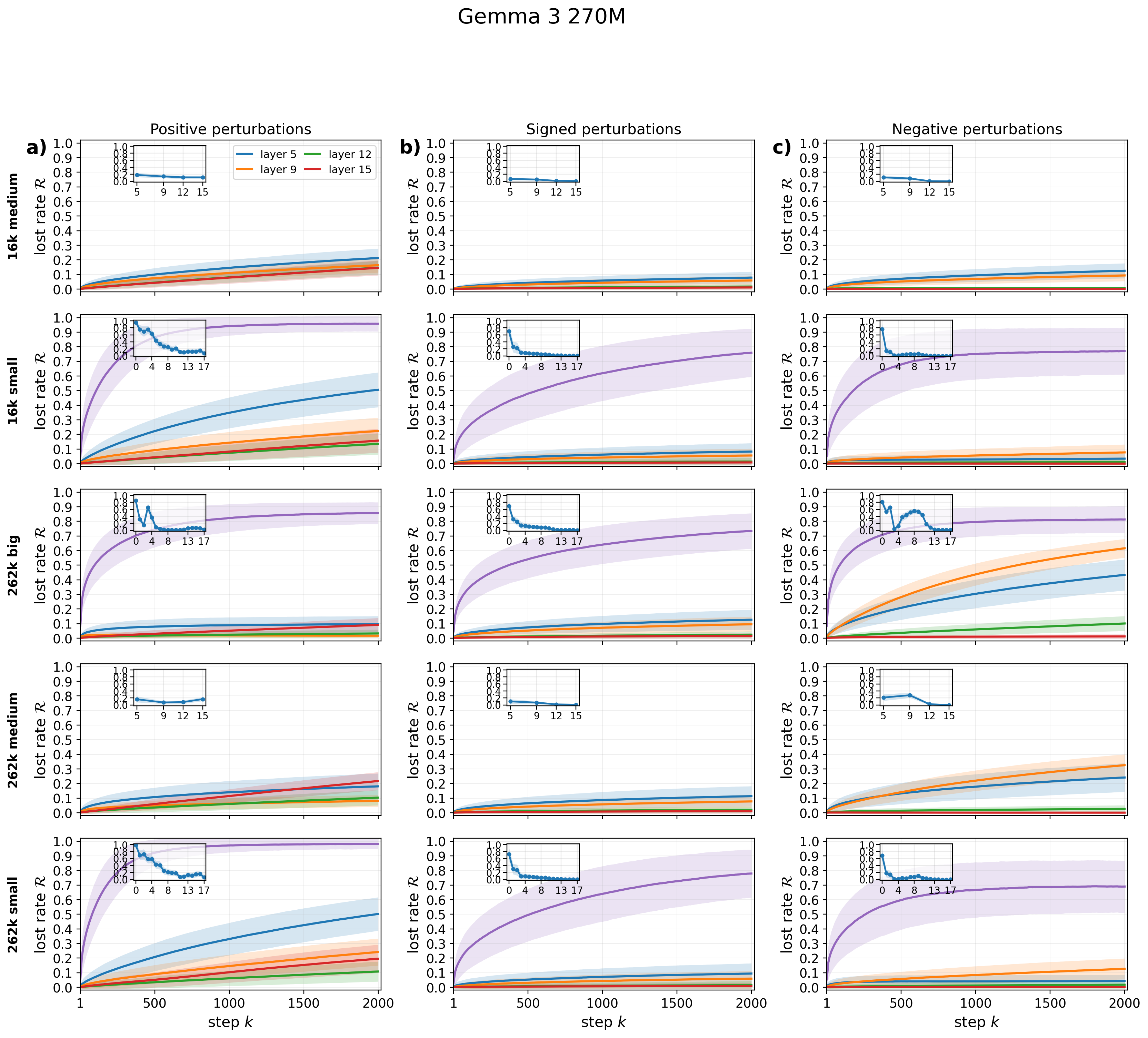}{Feature steering results for Gemma 3 270M.}{fig:constructsweep_gemma_270m}
\sweepfig{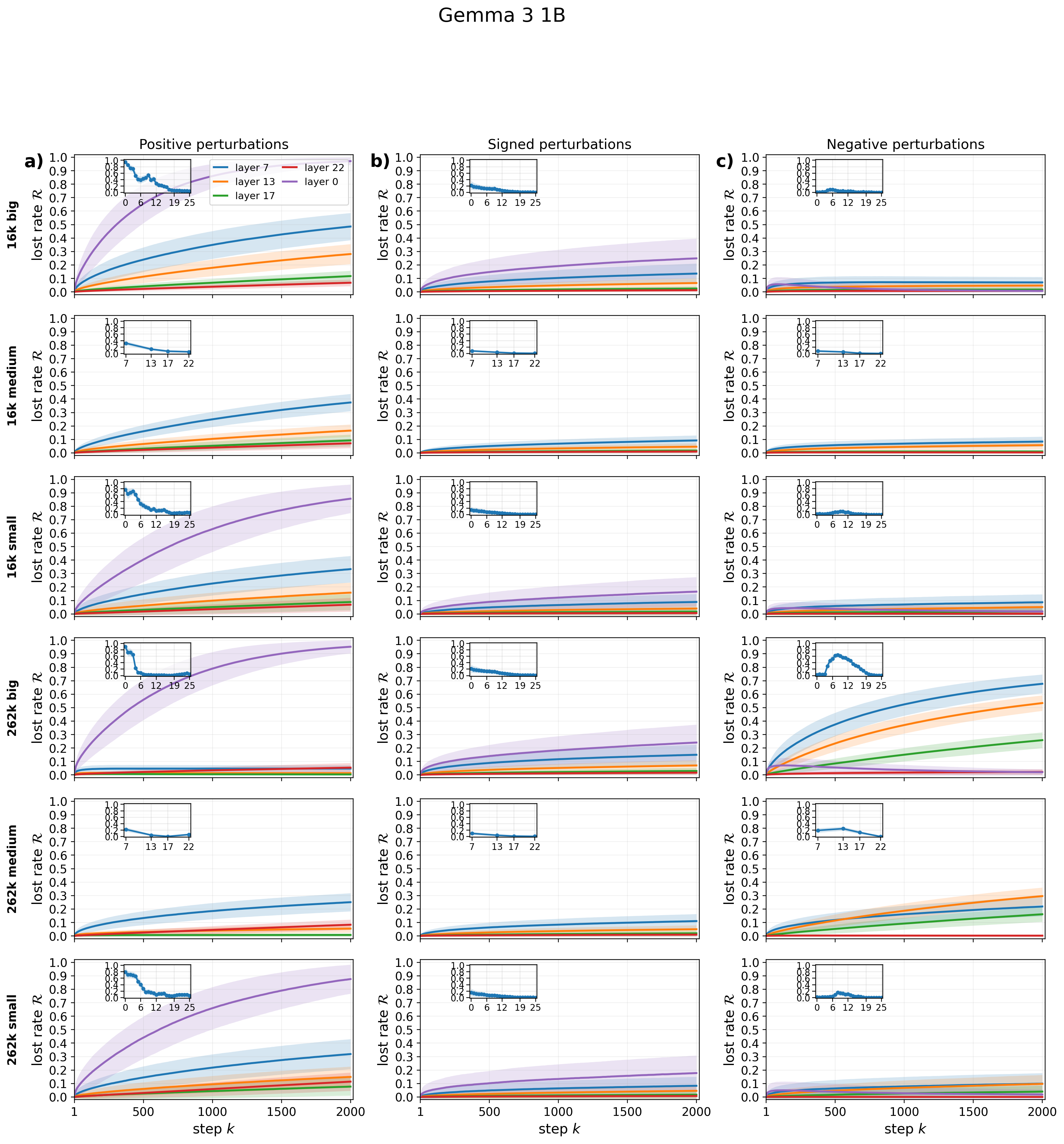}{Feature steering results for Gemma 3 1B.}{fig:constructsweep_gemma_1b}
\sweepfig{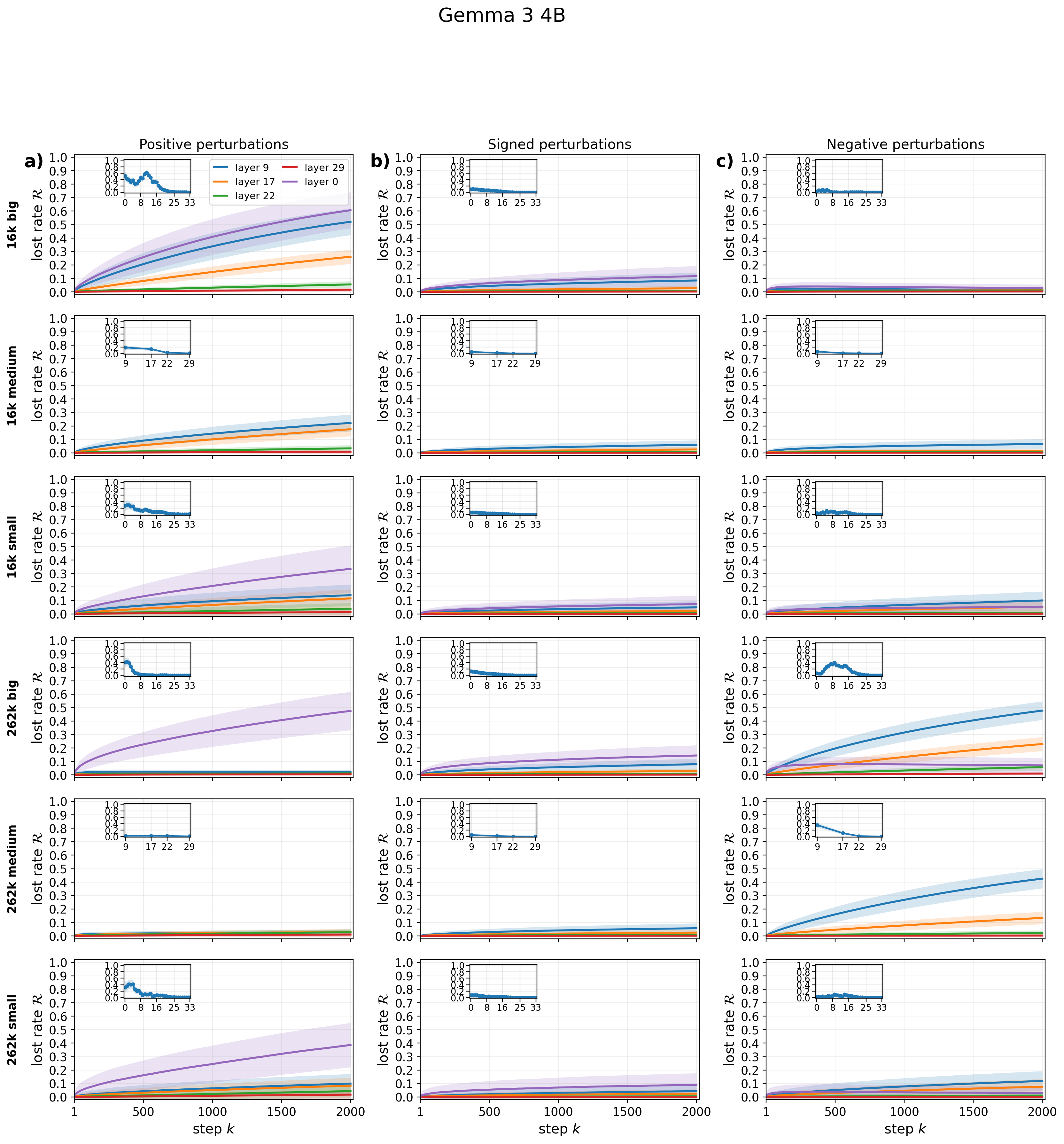}{Feature steering results for Gemma 3 4B.}{fig:constructsweep_gemma_4b}
\sweepfig{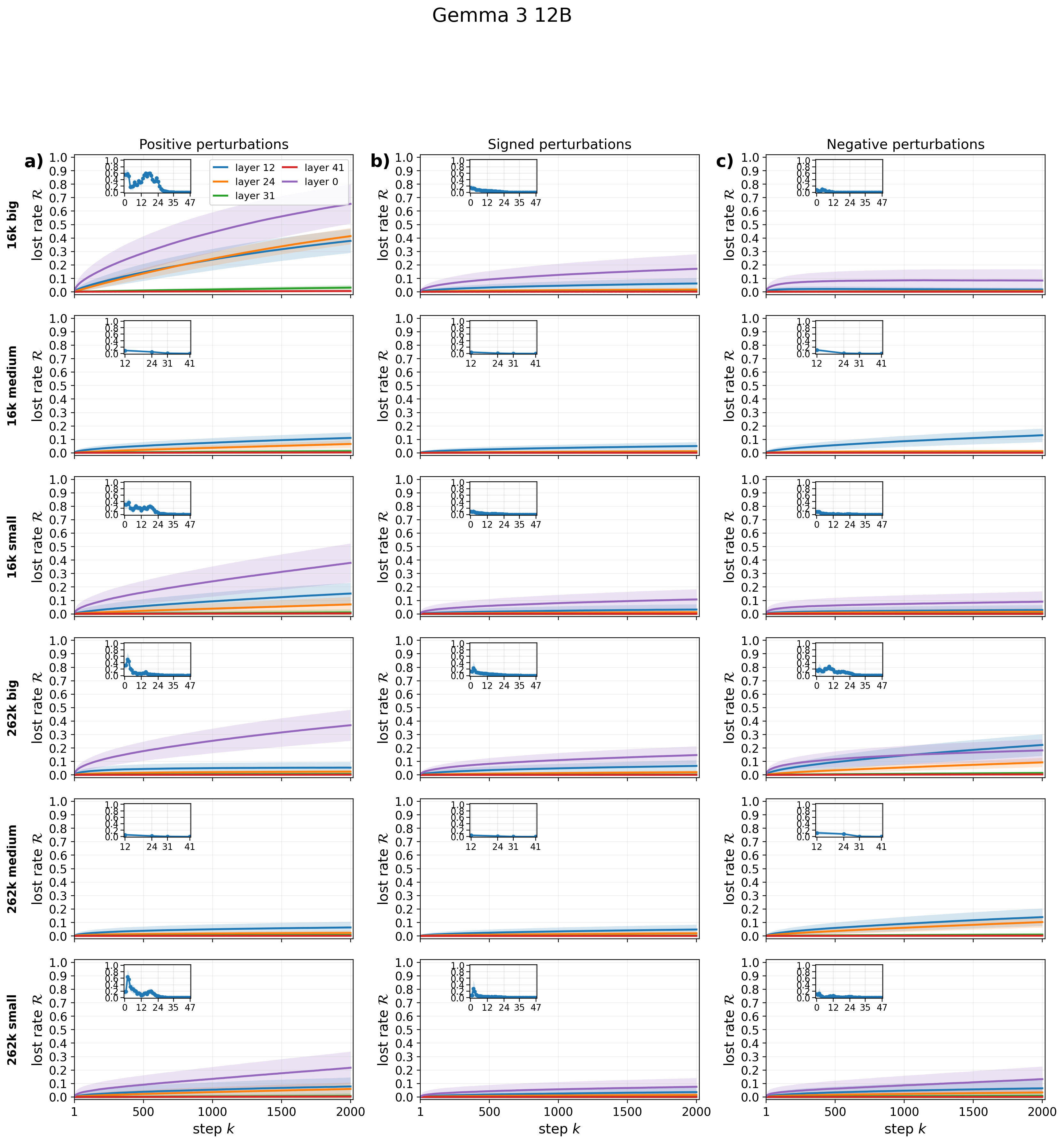}{Feature steering results for Gemma 3 12B.}{fig:construtcsweep_gemma_12b}
\sweepfig{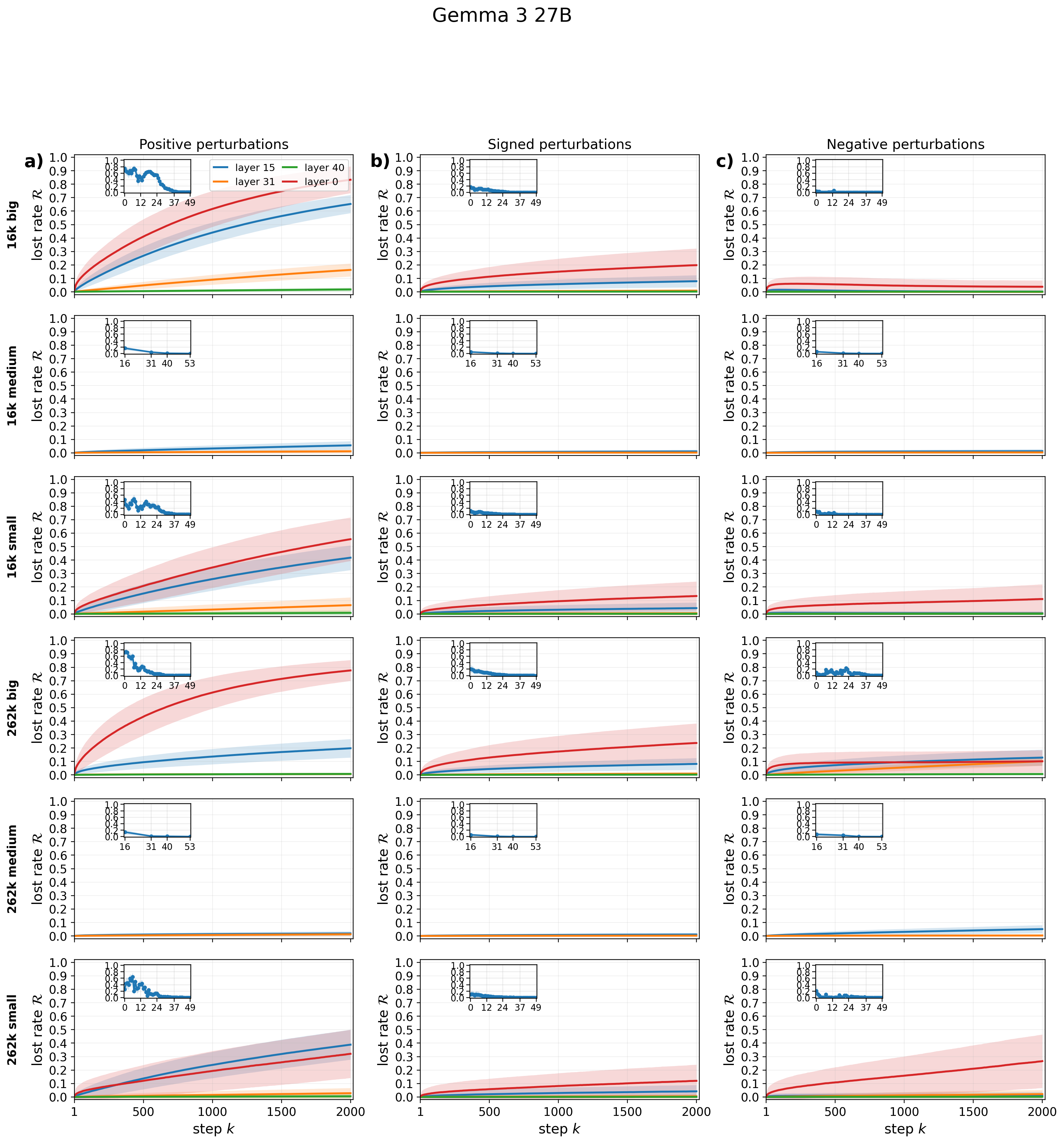}{Feature steering results for Gemma 3 27B.}{fig:constructsweep_gemma_27b}
\sweepfig{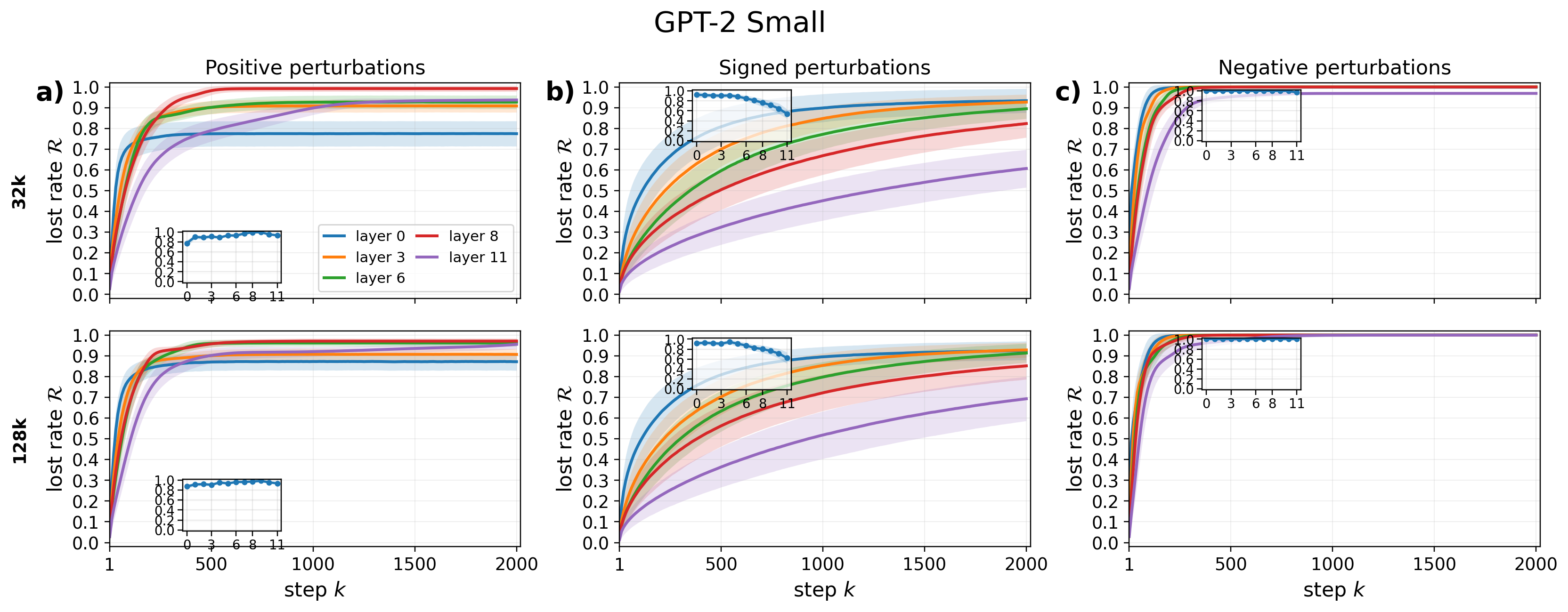}{Feature steering results for GPT-2 Small.}{fig:constructsweep_gpt2_small}
\sweepfig{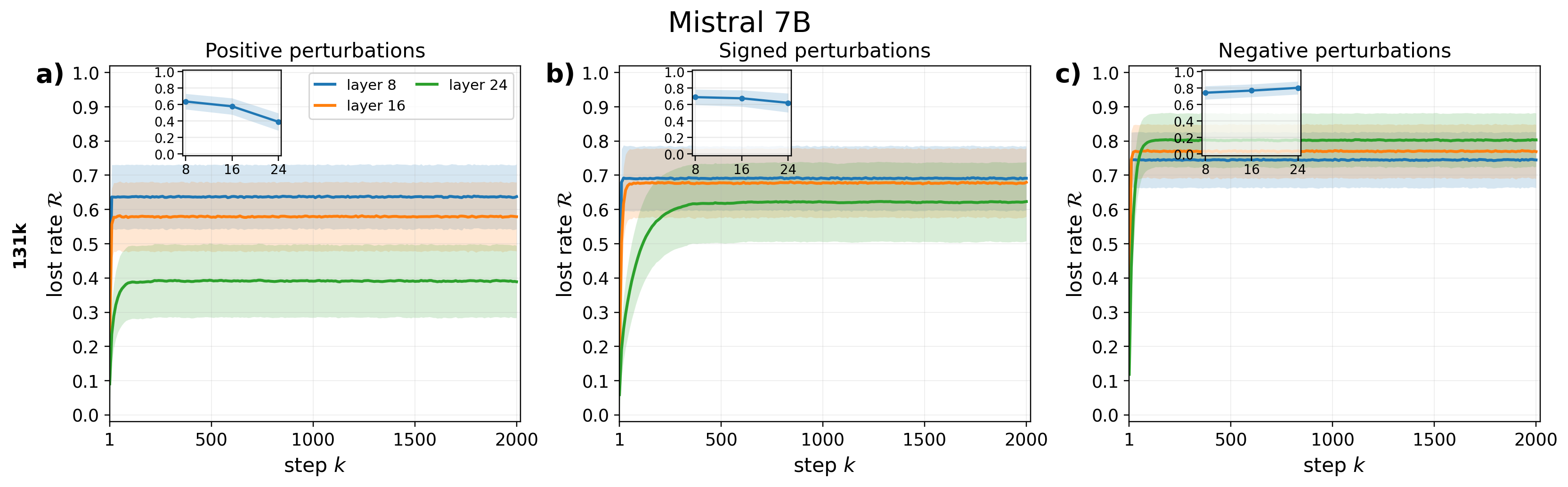}{Feature steering results for Mistral 7B.}{fig:constructsweep_mistral_7b}
\sweepfig{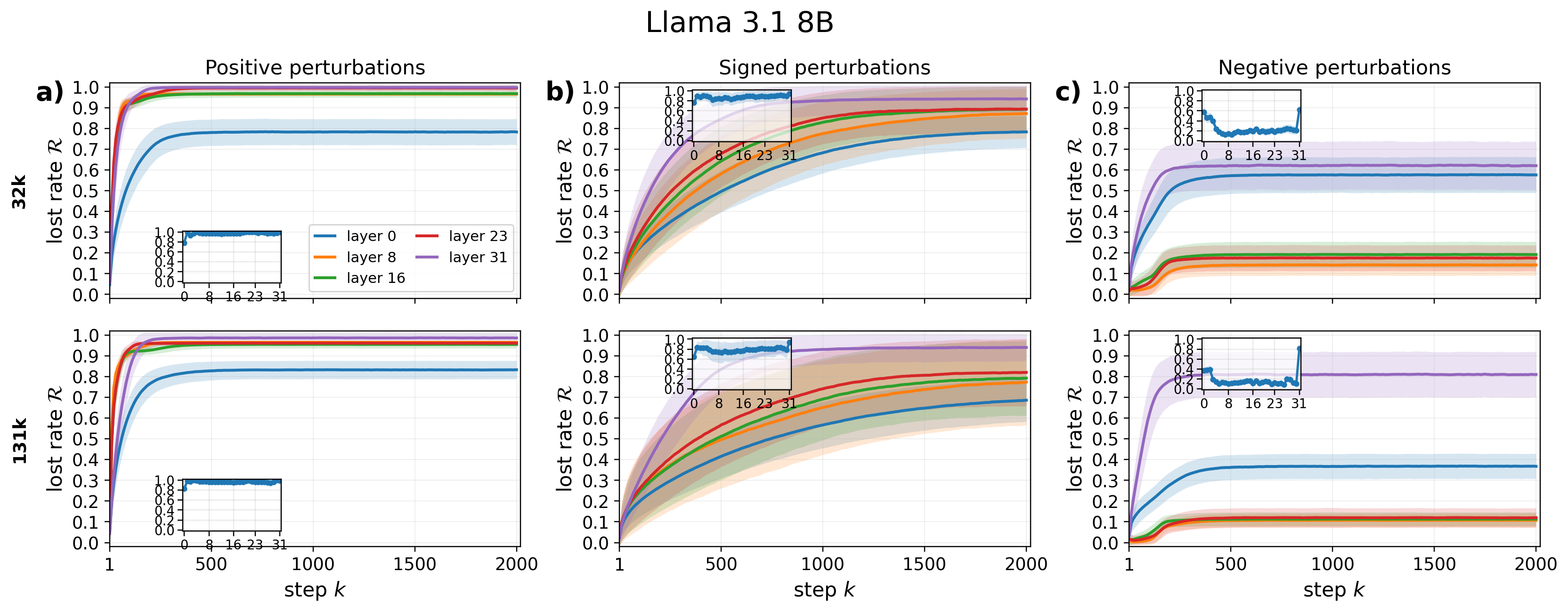}{Feature steering results for Llama 3.1 8B.}{fig:constructsweep_llama31_8b}

\FloatBarrier
\begin{strip}
\centering
\includegraphics[width=\textwidth,keepaspectratio]{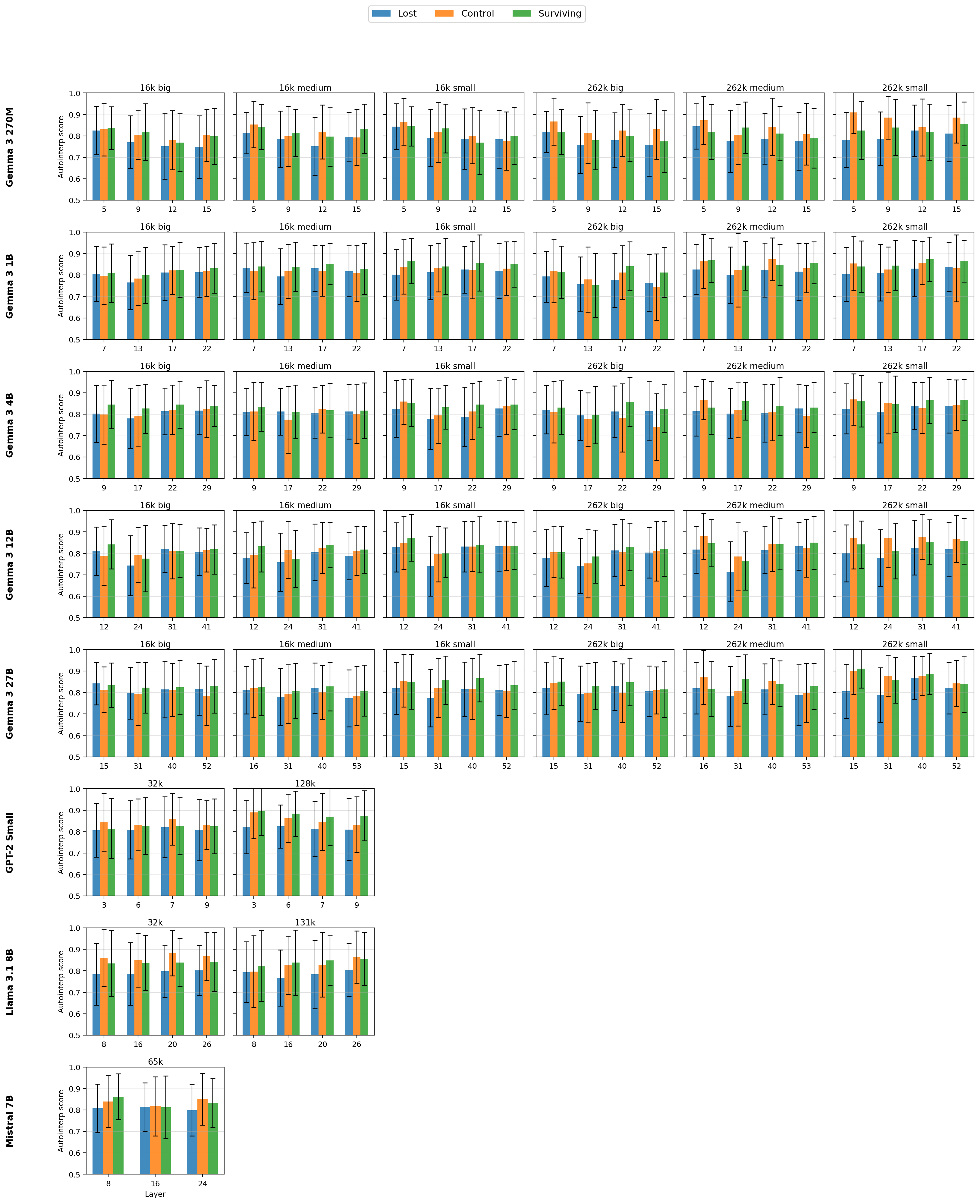}

\vspace{-0.6em}
\captionof{figure}{
Auto-interpretability scores for lost, surviving, and control latents across models and SAE variants. Each row corresponds to one model, and panels within a row show the SAE variants evaluated for that model from left to right. Within each panel, bars report mean auto-interpretability scores for the three latent groups at representative stages of model depth, approximately \(25\%\), \(50\%\), \(65\%\), and \(85\%\) of layers. Error bars indicate one standard deviation across evaluated latents. Lost latents are those active for the base prompt \(t_0\) but inactive for the composed prompt \(t_1\); surviving latents remain active in both prompts; control latents are randomly sampled from the SAE dictionary.
}
\label{fig:autointerp_lost_surviving_control}
\vspace{-0.5em}
\end{strip}
\subsection{Auto-interpretability of Lost Latents}
\label{app:autointerp_lost_latents}

We additionally ask whether latents lost under composition are less interpretable than latents that remain active. For each noun--adjective pair, we compare the base prompt \(t_0\) with the composed prompt \(t_1\) and classify latents into three groups: \emph{Lost} latents, active on \(t_0\) but inactive on \(t_1\); \emph{surviving} latents, active in both prompts; and randomly sampled \emph{control} latents. For each group and layer, we evaluate up to 100 latents.\\
We follow the SAEBench auto-interpretability protocol \citep{karvonen2025saebench}. For each latent, cached SAE activations are used to collect activating examples. During explanation generation, GPT-4o-mini receives the surrounding context with the activating token marked by API-call and produces a natural-language description of the latent. During scoring, the model receives the explanation and held-out sequences, consisting of true activating examples and random non-activating examples, and predicts which sequences contain an activation. The score is the resulting sequence-level classification accuracy. We compute one score per latent and average these scores within the \emph{lost}, \emph{surviving}, and \emph{control} groups for each layer. Latents are discarded if, after masking all top-\(k\) activation positions, no remaining activation exceeds \(10^{-6}\); such latents effectively fire only in the top-\(k\) region and cannot provide disjoint importance-weighted samples. We therefore cache more than 100 candidate latents per group to retain 100 evaluated latents after filtering.\\
\\
Results across investigated models and SAE types are shown in Figure~\ref{fig:autointerp_lost_surviving_control}. We report auto-interpretability scores for lost, surviving, and control latents at four representative depths, approximately \(25\%\), \(50\%\), \(65\%\), and \(85\%\) through the model. The groups occupy a similar score regime, with means often within each other's error bars, so the results should not be read as a strong separation. Still, lost latents show a weak recurring tendency toward the lowest mean scores.\\
Note, this does not suggest these latents are therefore less relevant and thus discarded by the model under the more specific prompt. It suggests that latents lost under semantic refinement may be somewhat less well captured by current automatic explanations, possibly because they participate in more distributed or context-dependent computations. This provides weak supporting evidence for the more nuanced view on these latents, also suggested by the pre-activation analyses above.

\section{Resources and AI Use}
\label{sec:resources_ai_use}

\paragraph{Compute resources.}
All experiments were performed on a node with four NVIDIA RTX~3090 GPUs. Training the SAEs for the toy-model experiments in Section~\ref{app:toy_union_setup} required approximately one day of compute. The latent-set clustering experiment in Section~\ref{app:interactive_sae_clusters} required approximately three hours. The human-concepts experiments following \citet{shani2025tokensthoughtsllmshumans} required approximately one day of compute. The lost-latent experiments in Section~\ref{sec:motivation_set_consistency} and Appendix~\ref{app:sweep_lost_feats}--\ref{app:sweep_steering} required approximately six days in total. The auto-interpretability analysis in Appendix~\ref{app:autointerp_lost_latents} required approximately two additional days. Preliminary experiments required approximately one further week of compute.\color{black}

\paragraph{Datasets and software resources.}
For the toy-model experiments, we build on the repository and SAE-training infrastructure of \citet{feature_chanin_2025}, adapting their training setup to our LRH-aligned synthetic activation data. For the natural-text clustering experiment, we sample snippets from \texttt{The Pile} \citep{gao2020pile800gbdatasetdiverse}. For the conceptual-similarity experiments, we use the human-concepts data from \citet{rosch1973internal, rosch1975cognitive, mccloskey1978natural} and orient our evaluation procedure around the experimental setup from \citet{shani2025tokensthoughtsllmshumans}. The semantic construction experiments were implemented within our own experimental pipeline, no additional external datasets or task-specific code resources are used. For the auto-interpretability analysis, we use the SAEBench auto-interpretability pipeline \citep{karvonen2025saebench}. All experiments involving pretrained models and SAEs were implemented using SAELens \citep{SAELens} and TransformerLens \citep{TransformerLens}. The pretrained language models and SAE suites used in this work are listed and credited in Appendix~\ref{app:saes_used}. All external artifacts are used under their publicly available licenses or access terms and only for research purposes.

\paragraph{AI use.}
We used generative AI tools for language refinement, plotting-code assistance, and coding support during implementation checks and debugging. AI-generated suggestions were used only as auxiliary support. The authors remained responsible for the experimental design, implementation decisions, result interpretation, scientific claims, and final manuscript, and reviewed and validated all submitted content. No generative AI system is listed as an author. 
\end{document}